\documentclass[final,5p,times,twocolumn]{elsarticle}
\usepackage{mathtools}
\usepackage{amssymb}
\usepackage{amsmath}
\usepackage{setspace}
\usepackage{geometry}
\usepackage{nomencl}
\usepackage[american]{babel}
\usepackage{url}
\usepackage{subcaption}
\usepackage[colorlinks=true,linkcolor=blue,anchorcolor=blue,citecolor=blue,filecolor=blue,menucolor=blue,runcolor=blue,urlcolor=blue,breaklinks]{hyperref}
\usepackage{booktabs} 
\hypersetup{pdfauthor={Name}}
\usepackage{siunitx}
\usepackage[switch]{lineno}

\journal{arxiv}

\begin{document}
\begin{frontmatter}
\title{Artificial intelligence-driven digital twin of a modern house demonstrated in virtual reality}

\author[adilsaddress]{Elias Mohammed Elfarri}
\author[adilsaddress]{Adil Rasheed\corref{mycorrespondingauthor}}
\cortext[mycorrespondingauthor]{Adil Rasheed}
\ead{adil.rasheed@ntnu.no}
\author[omersaddress]{Omer San}
\address[adilsaddress]{Department of Engineering Cybernetics, Norwegian University of Science and Technology}
\address[omersaddress]{School of Mechanical and Aerospace Engineering, Oklahoma State University}

\begin{abstract}
A digital twin is a powerful tool that can help monitor and optimize physical assets in real-time. Simply put, it is a virtual representation of a physical asset, enabled through data and simulators, that can be used for a variety of purposes such as prediction, monitoring, and decision-making. However, the concept of digital twin can be vague and difficult to understand, which is why a new concept called "capability level" has been introduced. This concept categorizes digital twins based on their capability and defines a scale from zero to five, with each level indicating an increasing level of functionality. These levels are standalone, descriptive, diagnostic, predictive, prescriptive, and autonomous. By understanding the capability level of a digital twin, we can better understand its potential and limitations. To demonstrate the concepts, we use a modern house as an example. The house is equipped with a range of sensors that collect data about its internal state, which can then be used to create digital twins of different capability levels. These digital twins can be visualized in virtual reality, allowing users to interact with and manipulate the virtual environment. The current work not only presents a blueprint for developing digital twins but also suggests future research directions to enhance this technology. Digital twins have the potential to transform the way we monitor and optimize physical assets, and by understanding their capabilities, we can unlock their full potential.     
\end{abstract}

\end{frontmatter}
\section{Introduction}
\label{sec:introduction}
\begin{figure*}[!ht]
\centering
\includegraphics[width=\linewidth]{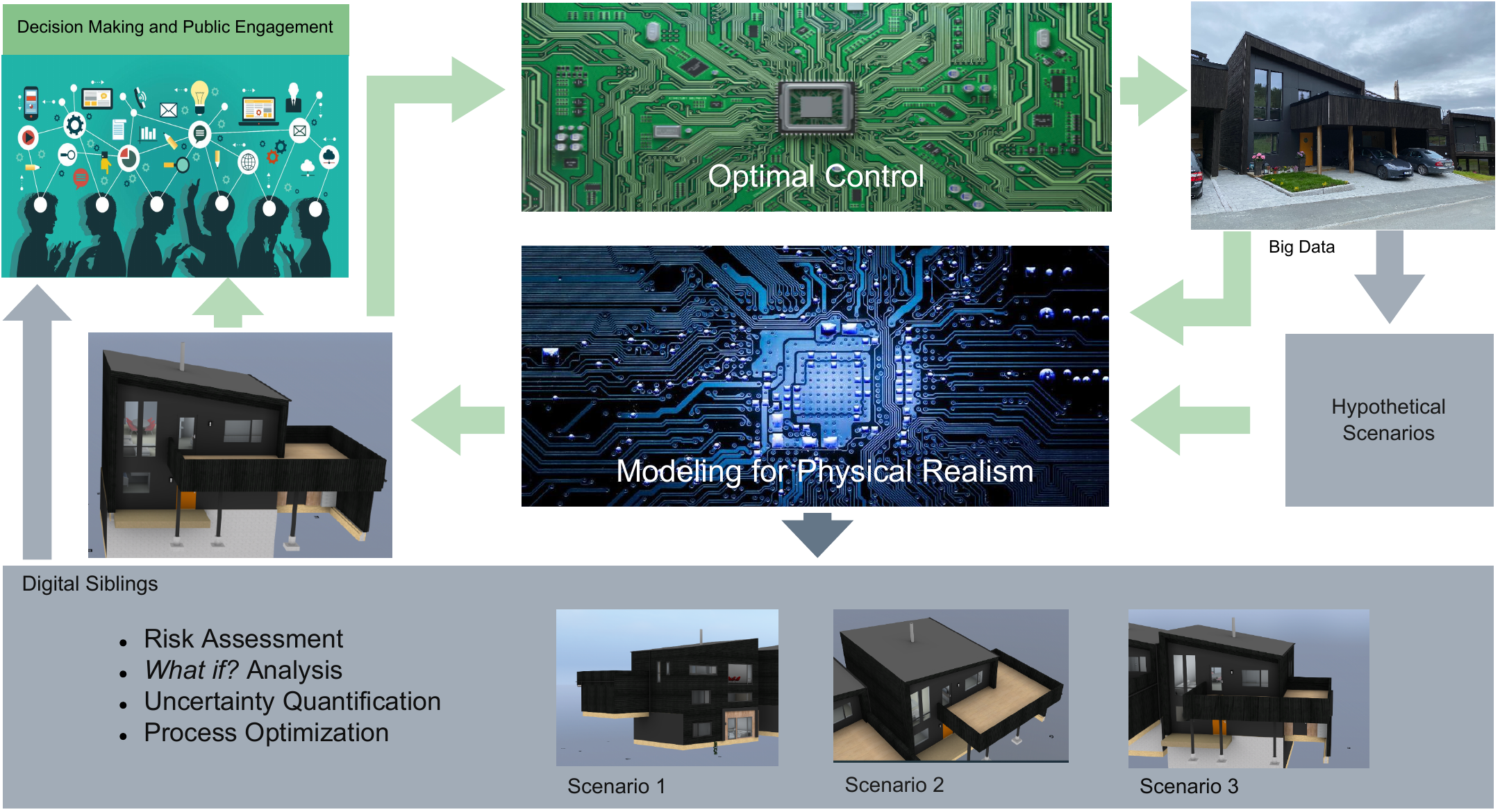}
\caption{Schematic of a digital twin, and digit sibling concept. The top-right most box with a house represents any asset equipped with sensors acquiring big data. The data is processed using models to improve the spatio-temporal resolution for instilling physical realism in the digital twin. Information from the digital twin can be used for informed decision making and public engagement. Additionally it can be used for optimally controlling the asset. The green arrows signify real-time data / information transfer. The same architecture can also be used for conducting offline hypothetical scenario analysis, in which case can be called digital siblings.}
\label{fig:dt1}
\end{figure*}

A digital twin (DT) is a virtual replica of a physical asset, enabled through data and simulations, that can be used for real-time monitoring, optimization, and decision-making \cite{rasheed2020dtv}. The motivation to study DTs is rooted in the potential cost savings and efficiency gains they offer. In Fig. \ref{fig:dt1}, we can see the concept of a DT. The physical asset is located in the top right side of the figure, equipped with various sensors that provide real-time big data. However, this data has limited spatio-temporal resolution and does not tell about the future state of the asset. To complement the measurement data, models are used to create a digital representation of the asset. If the DT can provide the same information as the physical asset, it can be utilized for informed decision-making and optimal control. The green arrows in the figure show real-time data exchange and analysis. To perform risk assessment, what-if analysis, uncertainty quantification, and process optimization, the DT can be run in an offline setting for scenario analysis. It is then called digital siblings. The grey box and arrows represent the digital sibling. Additionally, the DT predictions can be archived during the lifetime of the asset and can be used for designing a next generation of assets, in which the concept is referred to as digital threads.
\begin{figure*}
\centering
\includegraphics[width=\linewidth]{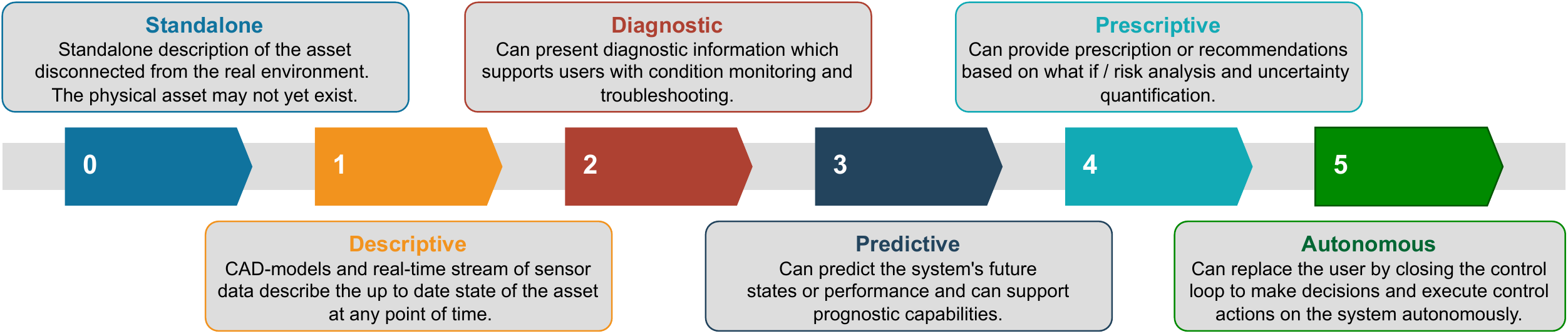}
\caption{Description of capability levels of a digital twin.}
\label{fig:cap_level_framework}
\end{figure*}
The authors in \cite{San2021haa} present a DT capability level scale adapted from a DNV GL report \cite{dnvgl-cap-levels} that divides a DT into six distinct levels. These levels are 0-Standalone, 1-Descriptive, 2-Diagnostic, 3-Predictive, 4-Prescriptive and 5-Autonomous (Fig. \ref{fig:cap_level_framework}). Standalone DT can exist even before the asset is built and can consist of solid models. When the asset is in place and is equipped with sensors, data can be streamed in real-time to create a descriptive DT, giving more insight into the state of the asset. When analytics tools are applied to the incoming data stream to diagnose anomalies, the DT advances to a diagnostic level. At the first three levels, the DT can provide information/insight only about the past and present. However, a predictive DT can describe the future state of the asset. Using the predictive DT, one can do scenario analysis to provide recommendations to push the asset to the desired state. This is then referred to as prescriptive level. Lastly, the asset updates the DT at the autonomous level, and the DT controls the asset autonomously. We will refer to this setting as the capability level framework for DTs from now onwards.

\begin{table*}[!htb]
\begin{center}
\caption{Comparing current work with industry DT services.}
\begin{tabular}{ |p{2.25cm}|p{2cm}|p{2cm}|p{2cm}|p{1.75cm}|p{2cm}|    }
\hline
\multicolumn{6}{|c|}{Comparing Digital Twin Solutions} \\
\hline
\textbf{Capability} & Standalone & Descriptive & Diagnostic & Predictive & Prescriptive  \\
\hline
Current work & X & X & X & X & X  \\
\hline
Matterport  & X & - & - & - & -  \\
\hline
Openspace  & X & - & - & - & -  \\
\hline
Revit  & X & - & - & - & -  \\
\hline
Invicara  & X & X & - & - & -  \\
\hline
TwinView     & X & X & - & X & -  \\
\hline
\end{tabular} 
\label{table:cap_levels_comparison}
\end{center}
\end{table*} 
Within the field of DTs there is no consensus as to what qualifies as a DT application \cite{Shahzad2022dti}.  Organizations and sectors operate with different definitions of DTs that are simply too vague and generic to provide any indication of the current capabilities of the DT \cite{Batty2018dt}. Due to the ambiguity of the definitions of DTs, researchers and practitioners may dismiss them as mere marketing hype. As a result, once the excitement and the inevitable backlash have passed, interest in and use of this promising technology may not reach its full potential \cite{wright2020tell}. Furthermore within the built environment, a BIM of a building relates somehow to the DT of the asset, but many still struggle to see the liaison between the two \cite{douglas2021bdtc}. More recently some encouraging advancement has started to happen in the field for e.g. \cite{Wang2022bii} improves the construction efficiency to ensure the infrastructure needs of urban development using the BIM model. In \cite{Lydon2019cso} the focus is on the modelling methodology used by the energy domain to support the development of a DT for a multi-functional building element. The authors in \cite{Wang2022dlf} argue that a comprehensive perception of physical systems is the preconditions for DTs implementation while in \cite{Zhao2022dac} it is demonstrated that DT technologies can enable efficient and responsive planning and control of facility management activities by providing real-time status of the building assets. Researchers in \cite{Angjeliu2020dot} studied the structural system integrity using finite element method in historical masonry buildings using the concept of the DT. It can be easily realized that the capability level scale concept has the potential to make the DT related communication more standardized and can help compare different existing DT solutions available in the market. A brief overview of the capability of existing solutions in the market is presented in Table \ref{table:cap_levels_comparison}. It shows a clear lack of high capability level DTs in the industry. The connection between the DT capability level and the service that each company provides depends on what the focus of the provider is. For instance, Matterport \cite{e1m} and Openspace are virtual tour services that focus on providing 360 degrees photogrammetry for building management \cite{Matterport, openspace_ai}, and therefore only qualify as a standalone DT service. Note that there are many other similar standalone services in the market \cite{ocean_engineering_vr,  onsiteviewer}. Revit is a BIM modeling software and is therefore only limited to constructing 3D models of a building \cite{AutodeskRevit}. Invicara and TwinView have an ambition of allowing customers to combine BIM models with IoT sensor data integration qualifying both companies for descriptive DT, with the addition of Twinview providing predictive DT capabilities \cite{TwinView, Invicara}. None of the above-mentioned applications qualify as diagnostic or prescriptive DTs. To this end, the current work 
\begin{itemize}
    \item introduces the concept of DT and its capability level in the context of built environment. 
    \item presents the basic ingredients to get started with building a DT.
    \item combines the power of AI, advanced sensor technologies, and virtual reality (VR) to develop the DT which is used as a way to communicate the concept and its values.
    \item proposes future research directions to enhance the capability of DTs.
\end{itemize}

It is important to stress that the objective of this paper is not to present a detailed analysis of the diverse class of data used in this project. Instead, we are using data, and their analysis to demonstrate the potential value of DTtechnology. To the best of our knowledge, this has never been attempted before at the fidelity addressed in the current article.

This paper is structured as follows. Section~\ref{sec:theory} presents the relevant theory that has been used to develop the DT. We then outline the methodology of the work in Section~\ref{sec:methodandsetup}, namely how the data was generated, how the models were trained, and how they were evaluated. In Section~\ref{sec:resultsanddiscussions} we present the results and discuss them. Finally, we summarise our findings and outline future work in Section~\ref{sec:conclusionsandfuturework}.

\section{Theory}
\label{sec:theory}
This section gives a brief description of the concepts, algorithms, and tools utilized to develop and demonstrate the DT concept using the capability level framework. Since one of the goals of this work is to provide a blueprint for developing DT of any asset from scratch, for the sake of completion, we have also included some theories (eg. collaborative filtering and big data cybernetics) which we have not yet used in the current work. 
    \subsection{3D Model Representation}
    As explained earlier, a standalone DT is a virtual description of an asset, disconnected from its physical counterpart. In the current context the house might not even exist at the inception of a standalone DT. However, the standalone DT can give the stakeholders a feel of the building and its environment, enabling them to make informed decisions. In creating the standalone DT we involved three basic steps, 3D modeling, rendering/texturing, and creating a virtual representation of the building. These are explained in the following section.
    \subsubsection{3D Modeling}
    The term 3D modeling is the process of using a software tool to construct a 3D representation of an object (house in the current context). The 3D model also called the Computer Aided Design (CAD) model consisting of point clouds, edges, and surfaces giving the illusion of physical objects. The 3D models can be utilized for engineering analysis using Computational Fluid Dynamics (CFD) and Finite Element Methods (FEM). 
    \subsubsection{3D Rendering}
    3D rendering is a computer graphics process in which surfaces of a 3D model are overlayed with textures such that the 3D object achieves a photorealistic appearance \cite{Yoon2008rtm}. 3D rendering can be broken down into three steps; visual texturing, lighting, and detailing. Visual textures refer to the visual perception of a spatial surface with a variety of details such as color, orientation, intensity, size, shape, and density \cite{Gimelfarb2008tab, Haralick1979sas}. In addition, lighting, reflection, and shadows can all be generated in 3D rendering software, and the light-absorbing characteristics of the materials are integrated into the texture. Detailing is the last step and requires a designer to carefully sculpt wear and tear, imperfections, dents, and other details into the surfaces, giving the model a more lifelike impression. For texturing to be interpreted across different platforms, it is usually represented in the form of a uv-map. uv-mapping is projecting a 2D image onto a 3D surface, where the "uv" part refers to the axes of the image projection. 
    \subsubsection{Virtualization using 3D Game Engine}
    A 3D game engine is a software development environment that effectively allows for the rapid development of interactive 3D experiences and games. A game engine is known to support a programming environment, 2D and/or 3D rendering, accurate physics engines, and many more well-optimized features that would take much time for a single developer to create on their own. Thanks to commercially available game engines, it is easy for an individual to only focus on the specifics of their own game, simulation, or experience \cite{Indraprastha2009tio}. Creating a standalone DT of an asset with 3D modeling and 3D rendering allows only for a static 3D model while using a 3D game engine through scripting language allows for state management of the 3D asset. This enables real-time evolution of the DT with respect to the asset it represents.
    \subsection{Virtual Reality}
    The concept of VR is not recent and can be defined as a model of the real world that is maintained in real-time, sounds and feels real with the possibility to directly and realistically manipulate the environment. Today there exist many affordable VR solutions generally consisting of a headset and complimentary controllers, that either utilize their own internal hardware or external processing power to render the virtual environment. Such a representation of a virtual environment through the usage of VR hardware compliments very well the visualization of a DT asset, which allows for a more realistic representation and feel of the asset.

    \subsection{Time Series Prediction and Forecasting Model}
    A time series can be defined as sequenced data consisting of real-valued continuous numerical observations that are a function of time \cite{Lin2003asr}. The data collected in the current work comes from sensors sampled at regularly spaced intervals; thus, the data can be viewed as continuous-valued but discrete in time. Time series predictions and time series forecasting, while being slight variations of the same thing, can often be confused to mean the same thing \cite{prediction_vs_forecasting}. In the context of machine learning (ML) and this work, a time series predictions model will refer to a regression model capable of predicting unknown or unseen values based on present information. On the other hand, a time series forecasting model is a regression model capable of making future predictions based on learned trends and seasonality, amongst other things.

        \subsubsection{AutoRegressive Integrated Moving Average}
        ARIMA, a time series forecasting model, is a versatile model that can capture both the linear and non-linear patterns in the data, as well as handle different types of seasonality and trend. The ARIMA model consists of three components: Autoregression (AR), Integration (I), and Moving Average (MA). AR models the relationship between an observation and a number of lagged observations. The idea behind this is that the current value of a time series is a function of its past values. AR models can be used to capture linear patterns in the data. Moving Average (MA) models the relationship between the errors of the time series. The idea behind this is that the errors in a time series are correlated with the errors of the previous time points. MA models can be used to capture non-linear patterns in the data. Integration (I) is a technique used to transform a non-stationary time series into a stationary one. A stationary time series has a constant mean and variance over time, which makes it easier to model. Integration is achieved by taking the difference between consecutive observations. ARIMA models are typically denoted as ARIMA($p,d,q$), where $p$, $d$, and $q$ are integers that represent the order of the AR, I, and MA components, respectively. 

        \subsubsection{Prophet}
        Prophet \cite{prophet} is a popular time series forecasting model designed to handle seasonality, holiday effects, and other time series features that are commonly encountered in real-world data. It is based on a decomposable time series model that can capture trend, seasonality, and holiday effects using piecewise linear models. Prophet is also robust to missing data and can handle outliers and changes in trend. Additionally, it offers a wide range of customizable options and hyperparameters to fine-tune the model performance. Overall, Prophet has gained popularity due to its ease of use, flexibility, and ability to provide accurate and reliable forecasts for a variety of time series applications.
        
         \subsubsection{Long Short Term Memory Networks}
         LSTM \cite{Hochreiter1997lst} is a type of RNN architecture that is specifically designed to handle time series data. LSTM networks are capable of learning and remembering long-term dependencies in time series data, making them well-suited for a wide range of applications such as speech recognition, natural language processing, and temperature prediction. Unlike traditional RNNs, which have a simple structure and can suffer from the vanishing gradient problem, LSTMs use a memory cell and a set of gates to selectively store, retrieve, and forget information. The memory cell can retain information over long time periods, while the gates control the flow of information into and out of the cell. This allows LSTMs to capture complex patterns and dependencies in time series data, making them a powerful tool for time series forecasting and prediction.
        
        \subsubsection{Gradient Boosting Machines}\label{theory:ensemble_learning}
        GBM is a powerful ML algorithm used for both regression and classification tasks. It works by iteratively building an ensemble of weak decision trees, where each new tree is trained to correct the errors made by the previous trees in the ensemble. GBM is a form of boosting, which means it improves the accuracy of the model by focusing on the misclassified samples in each iteration. The algorithm works by minimizing a loss function, such as mean squared error or log loss, using gradient descent. GBM has several advantages over other ML algorithms, such as the ability to handle missing data and outlier detection, and it is less prone to overfitting. GBM has become a popular choice for solving complex ML problems. GBM were one of the first sequential ensemble methods of its kind created by Friedman and have evolved to many of the state-of-the-art tree-based algorithms such as XGBoost, CatBoost and LightGBM  \cite{Chen2016xas, Prokhorenkova2018cub, Ke2017lah}.
        
        \subsubsection{Stacking}
        Stacking uses output predictions of base models as input to a second-level model, usually called the meta-learner. However, one cannot simply train the base models on the full training data, generate predictions on the full test set and then output these for the second-level training. This would not lead to the benefits that stacking provides. Instead, $K$-folds of the dataset is created. Then each model is fitted on $K-1$ of the training set and predicts only on $\frac{1}{K}$ of the data set, this is done for $K$ iterations until all data appears on the test set. All $K$ predictions of a single model are concatenated into the size of the original test set vector, this is done for each of the models. Finally, all these vectors are fed together as features to the meta-regressor, which produces the final predictions \cite{ibm_stacking, kaggle_stacking}. Empirical evidence shows that model stacking makes the model more robust to changes in the data set, allowing for better generalization \cite{Clarke2003cbm}. This is because the stacking deduces the bias in a model on a particular data set, and then corrects said bias in the meta-learner \cite{Wolpert1992sg}. 
        
        \subsubsection{Weight Averaging}    
        A simple but powerful way to create a strong predictor is by using parallel ensembling. One way to think of parallel ensemble learning is weighting the predictions of multiple different models. Another way of parallel ensemble predictions is to have multiple models of the same type e.g. LSTM, but each LSTM model has different hyperparameters or is fed with different features, and then their predictions are weighted to get the final prediction. One can combine and experiment with endless types of parallel ensembling, as each dataset might work very well with a specific kind. In Equation \ref{eq:weight_averaging} assuming predictions from $N$ different models, then the final prediction $\mathbf{\hat{y}}_f$ is the weighted average of all the individual models represented by $\mathbf{\hat{y}}_i$. 
         \begin{align}\label{eq:weight_averaging}
            \mathbf{\hat{y}}_f = \frac{\sum^N_{i=1} \mathbf{\hat{y}}_i w_i}{\sum^N_{i=1} w_i}   
        \end{align}
        where the weights $w_{i}$ given by 
        \begin{equation}
        w_i = \frac{1}{(\sqrt{\frac{1}{N} \sum^N_{k=1}(y^{val}_k - \hat{y}^{val}_k)^2})^p}
        \end{equation}
        are calculated using the predictions on the validation set (represented with the subscript $^{val}$) 
        Note that the tunable hyperparameter $p$ can be chosen to be any value greater than zero.
     
\subsection{Sun Position Prediction Model}\label{predictive:sun_model}
The approximate algorithm that is used in the current implementation of the predictive DT is inspired by \cite{SunPositionCSharp} which is taken from Montenbruck's book on algorithms about astronomical phenomena. The precision of these calculations is in the range of 01.03.1900 till 28.02.2100 as stated by the author \cite{dunlop2013astronomy}. The resulting sun position algorithm in Unity is accurate enough that it can be used for external lighting simulations on the house. In order to calculate the azimuth and altitude of the sun accurately, several parameters are required to convert the input date into the correct format. These parameters include $JD$, $JC$, $t_{UT}$, $t_{SR}$, $t_{SRUT}$, and $t_{LSR}$. Additionally, the ecliptic coordinates of the sun, $\lambda$ and $\beta$, are calculated using the values of $L_0$ and $M_0$. The ecliptic coordinate system is used to represent the apparent positions of any solar system object. These coordinates, together with $\Omega$, are then used to calculate the equatorial coordinates, $\alpha$ and $\delta$. This enables us to determine the position of the sun relative to the earth, with the earth located at the origin of the equatorial coordinate system. Finally, the coordinates are converted into azimuth and altitude angles, $\phi$ and $\theta$, which provide the output sun coordinates relative to a stationary point on the earth's surface. In our case, the stationary point is the longitude and latitude positions of the house.

First, in order to calculate the number of days relative to the reference date and time of January 1, 2000, 12h Universal Time (J2000), we need to calculate the Julian Date. The Julian Date is represented by $JD$ and can be calculated using the following equations: 
\begin{equation*}
    JD = 367Y - \frac{7}{4}(Y + \frac{9}{12}M) +  \frac{275}{9}M + D - 730531.5
\end{equation*}
This relates to the Julian Century as
\begin{equation*}
    JC = \frac{JD}{36525}
\end{equation*}
Furthermore using Julian Century, Sidereal time (given in hours) can be defined, which is the meridian of Greenwich at midnight (00h) for a given date as well as the conversion from sidereal time of the Greenwich meridian for Universal Time can be calculated as follows.
\begin{align*}
    t_{SR} = 6.6974 + 2400.0513JC   \\
    t_{SRUT} = t_{SR} + \frac{366.2422}{365.2422}t_{UT}
\end{align*}
Where 365.2422 is the length of a tropical year given in days. Finally, the local sidereal time for a geographical longitude $L$ can be found as such.
\begin{equation*}
    t_{LSR} = t_{SRUT} + L
\end{equation*}
Local sidereal time will come in handy later when calculating the altitude and azimuth of the sun. Firstly number of days $d$ from J2000 for the input date is given as. 
\begin{equation*}
    JD_d = JD + \frac{t_{UT}}{24}
\end{equation*}
Which is needed to calculate the relative centuries from the reference time. 
\begin{equation*}
    JC_d = \frac{JD_d}{36525}
\end{equation*}
Using the Julian century of the input date the sun's mean longitude and its mean anomaly can be calculated.
\begin{equation*}
    L_0 = 280.466 + 36000.770  JC_d
\end{equation*}
\begin{equation*}
    M_0 = 357.529 + 35999.050 JC_d
\end{equation*}
The sun's equation of center C is given as
\begin{equation*}
    C = (1.915 - 0.005JC_d)\sin(M_0) + 0.020\sin(2M_0)
\end{equation*}
Using the Equation of center of the sun the ecliptic longitude can be found. Note that the ecliptic latitude is approximately zero ($\beta \approx 0$).
\begin{equation*}
    \lambda = L_0 + C
\end{equation*}
There are many intermediate calculations that need to be done in order to get the correct position of the sun relative to where a person's relative geographical longitude and latitude. Consequently in order to find the azimuth and altitude it is important to find the Sun's equatorial coordinates, right ascension $\alpha$ and declination $\delta$ which are both relying on the obliquity of the ecliptic $\Omega$.
\begin{equation*}
    \Omega = 23.439 - 0.013JC_d
\end{equation*}
\begin{equation*}
    \alpha = \arctan(\tan(\lambda) \cos(\Omega))
\end{equation*}
\begin{equation*}
    \delta = \arctan(\sin(\alpha) \sin(\Omega))
\end{equation*}
Now it is possible to proceed to find the horizontal coordinates for the sun for a given input of geographical longitude $L$ and latitude $B$. First the hour angle of the object is given as.
\begin{equation*}
    HA = t_{LSR} - \alpha
\end{equation*}
Resulting in the final Equations for the altitude of the sun $\theta$ and the azimuth $\phi$ respectively, where the algorithmic version of the azimuth $\phi$ is calculated using Arctan2 as suggested by \cite{Zhang2021asa}. 
 
 \begin{equation*}
     \theta = \arcsin(\sin(B)\sin(\delta) + \cos(B)\cos(HA))
 \end{equation*}
\begin{equation*}
    \phi = \arctan\Big(\frac{-\sin(HA)}{\tan(\delta)\cos(B)-\sin(B)\cos(HA)}\Big)
\end{equation*}
 These angles are required is utilized to correctly render the objects in the DT.
 
    \subsection{Recommender Systems}
    Recommender systems are widely used for information filtering, providing users with valuable insights from relevant data sources. These insights can be inferred from the data or concatenated from a collection of data features.

    \subsubsection{User-Based Collaborative Filtering}
    Collaborative filtering is an information filtering technique that predicts a user's interests or behavior by collecting data from many users. The underlying assumption is that if person A behaves similarly to person B in a specific context, then person A might behave similarly to person B in another context than a randomly chosen person from the population. Companies such as Amazon, Netflix, YouTube, and other services extensively use recommender systems to learn about user preferences and provide personalized recommendations based on their behavior and similarities with other users \cite{Steck2021dlf}. However, the main challenge with this approach is that it requires a lot of data about user behavior, not necessarily from the user in question, but from the entire user base from which the data is collected. Moreover, the cold start problem arises when a new user registers and has not provided any interaction yet, making it impossible to provide personalized recommendations \cite{Bobadilla2012acf}. At the beginning, the algorithm may not be very accurate but becomes more precise as more data is collected while the user is active.

    \subsection{Control Systems}
    \label{cybernetics}      
        \begin{figure*}[!htb]
            \centering
            \includegraphics[width=\linewidth]{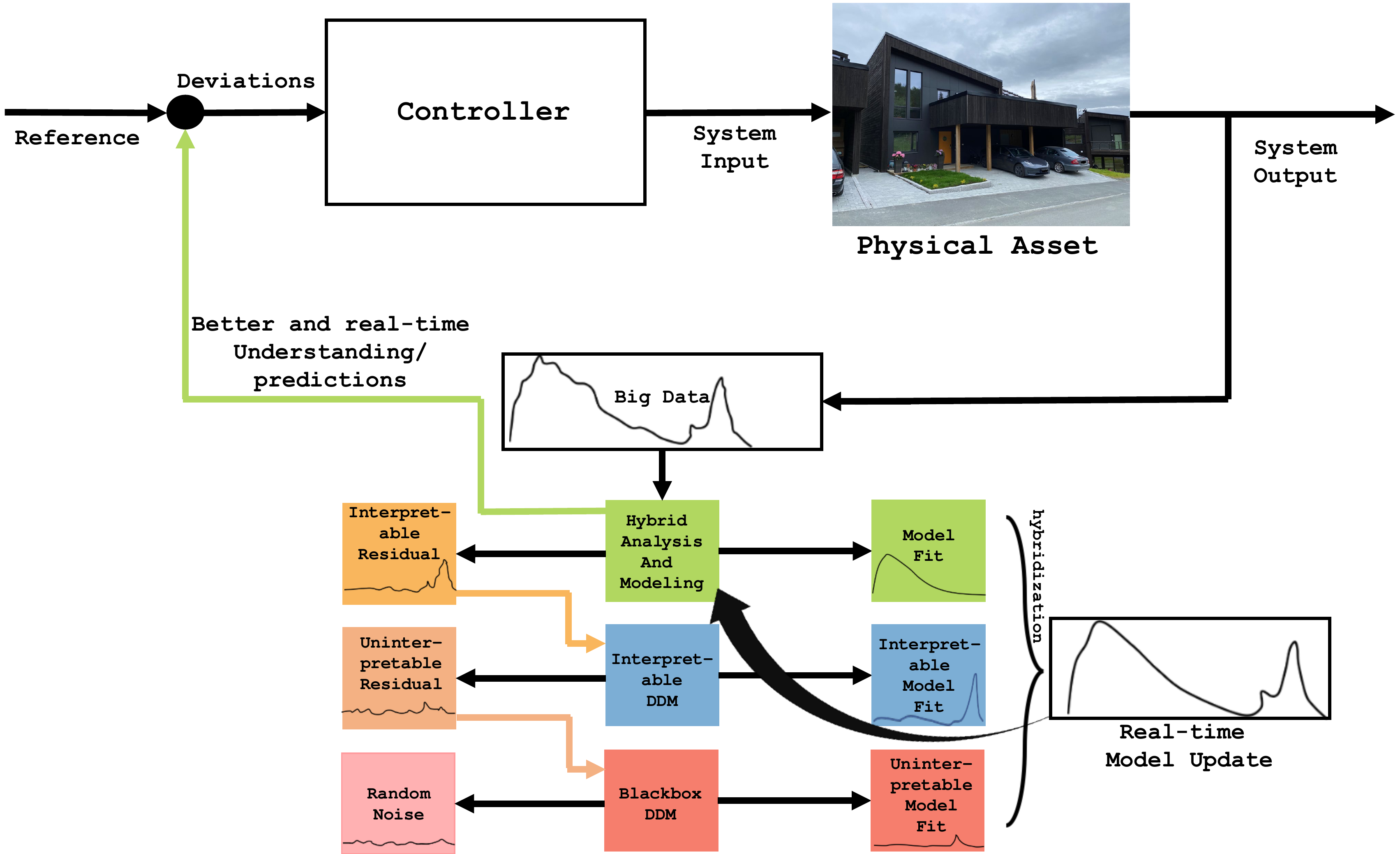}
            \caption{Big Data Cybernetics control loop.}
            \label{fig:bigcyb}
        \end{figure*}
        
        The aim of cybernetics in autonomous DTs is to guide the house towards an optimal set point. To achieve this, the system's output is continuously monitored and compared against a reference. The difference between the two, referred to as the error signal, is fed back to the controller, which generates a system input to bring the output set-point closer to the reference. As more sensors and communication technologies become available, larger volumes of real-time data, i.e., big data, are being generated. However, the quantity of interest may not be directly measurable, and it becomes a challenge to extract and understand the relevant information to be used for control purposes. Big Data Cybernetics is a new field of research that aims to address this challenge in a real-time control context. The first step involves interpreting the big data using well-understood physics-based models. The difference between the observation and the physics-based model is called the interpretable residual. In the second step, interpretable data-driven modeling approaches are used to model and analyze this residual. The remaining uninterpretable residual is then modeled using more complex black box models like Deep Neural Networks, which generally represent noise that can be discarded. This approach is known as Hybrid Analysis and Modeling (HAM), which continuously loops with the availability of new data resulting in ever-improving models. HAM is a promising new approach that aims to combine existing physics-based models with interpretable and non-interpretable big data-driven modeling techniques. Fig.~\ref{fig:bigcyb} illustrates the looped steps of the HAM process and the overall Big Data Cybernetics philosophy. HAM models can be utilized not only in the context of the interpretability of big data, but also in the Controller block. They can be incorporated alongside model-free control algorithms such as reinforcement learning (RL), or with model predictive control (MPC) algorithms, to improve the overall control performance of the system.

\section{Method and Set-up}
\label{sec:methodandsetup}
This section gives the details of setting up DTs at different capability levels. 

\subsection{Set-up for Standalone DT}
The asset in this case consists of a house, its surroundings terrain, and objects inside the house. Creating a standalone DT involved creating CAD models as a starting point. Here we give a detail of the steps involved in building the standalone DT.  

\subsubsection{3D CAD Model of the House}
\begin{figure*}[!htb]
    \centering
    \includegraphics[width=\linewidth]{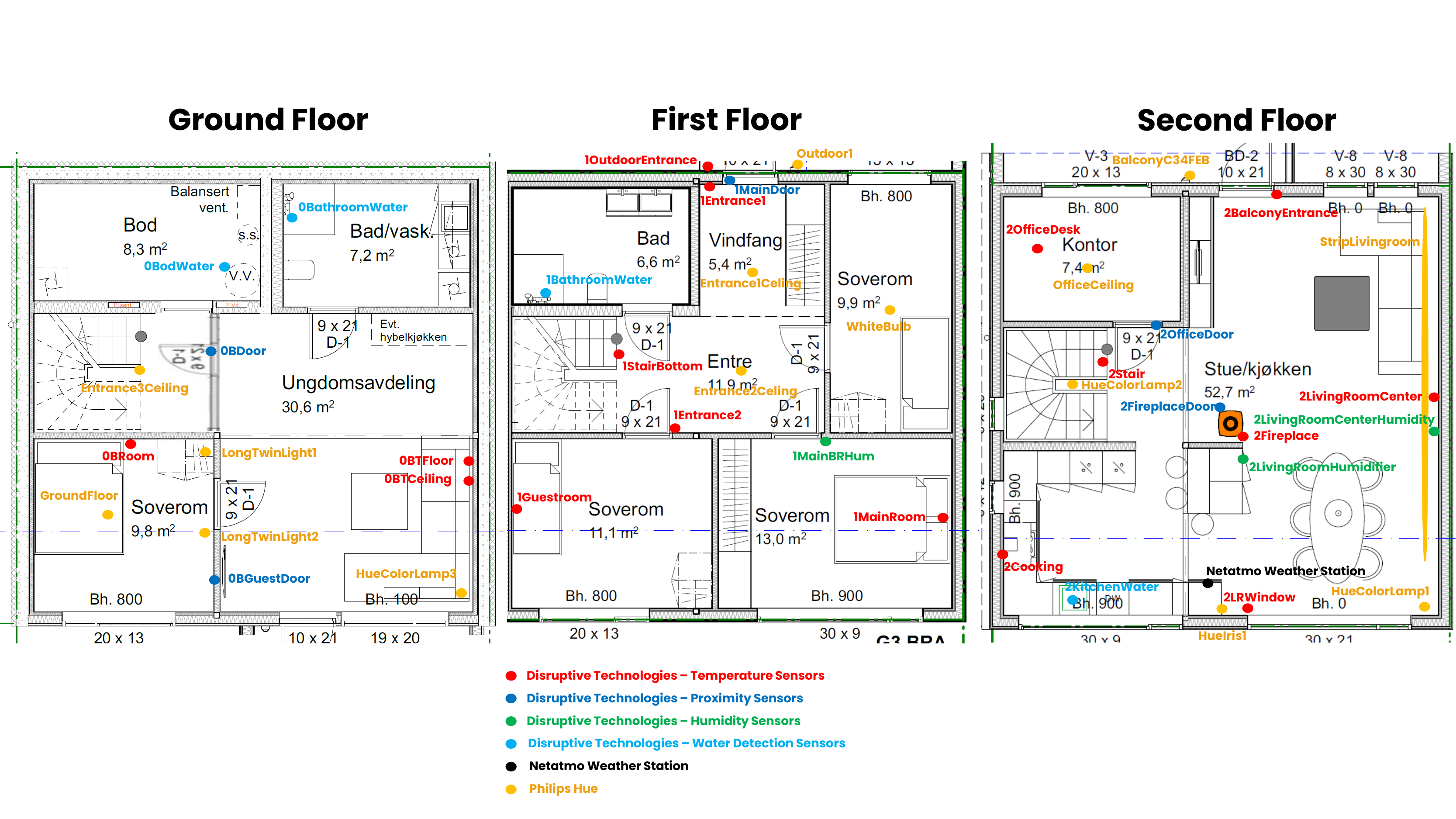}
    \caption{2D floor plans of the target house with all the sensors and devices setup for the project. All sensors are identified by their label and color as there are temperature, proximity, humidity, water detection and light sensors. All sensors are integrated into the Unity application environment where the BIM model of the house lives.}
    \label{fig:floorplan-sensors}
\end{figure*}
    The house's 3D CAD model was created using Autodesk Revit software based on the 2D floor plans, as shown in Fig.~\ref{fig:floorplan-sensors}. The three floors were stacked on top of each other, and details such as doors and windows were accurately placed. To distinguish different components of the house, each surface was tagged and textured appropriately. This tagging ensured that components could be identified accurately at a later stage if necessary. The scene was rendered using Autodesk 3DS Max and converted into a format compatible with the Unity Game Engine. High-quality textures were obtained from Architextures \cite{architextures}, a library designed for architectural models, and imported into 3DS Max. The textures were then applied to the uv-maps of walls, floors, and other surfaces to achieve the final look of the building. 
    
\subsubsection{Unity Game Engine for Interaction with the 3D Objects}
The Unity Game Engine offers a C\# .NET programming environment, 3D rendering capabilities, accurate physics engines, and optimized features that help developers create games more efficiently. Importing CAD models into Unity allows for real-time interaction with objects and instills physical realism through precise physics engines. If the exact physics engine is not available, it can be implemented within the Unity framework using programming languages such as C\# or Python. While Revit and 3DS Max are integrated by default, setting up the connection between 3DS Max and Unity requires additional steps. Unity can be linked with 3DS Max through a middleware called FBX Exporter, which was developed collaboratively by Unity and Autodesk to facilitate the workflow between their programs \cite{unity-3dsmax}. To use the FBX Exporter, it must be added as a plugin to the Unity project after installing all necessary software. Once the 3DS Max model is exported into Unity, some final setup is required. This includes selecting the 3D model in Unity, going under Materials" in the Unity inspector, and checking off Use External Materials" under Location" to import the textures from 3DS Max. Finally, clicking on Generate Colliders" under the model enables Unity to treat the house as a physical object and interact with other physical objects in compliance with Newtonian mechanics.

\begin{figure}[!htb]
       \begin{subfigure}{0.5\linewidth}
         \centering
         \includegraphics[width=\linewidth]{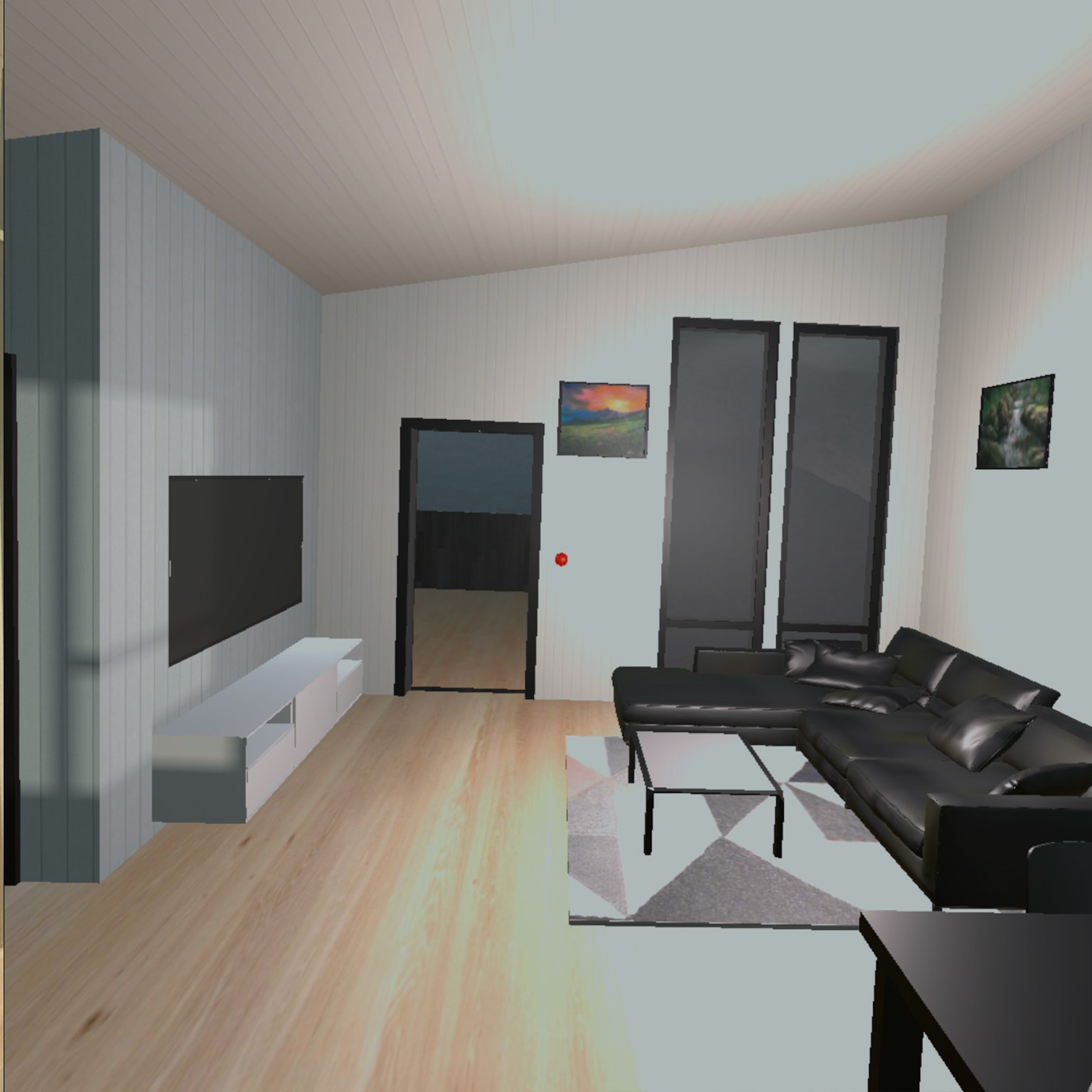}
         \caption{Virtual living room.}
       \end{subfigure}\hfill
       \begin{subfigure}{0.5\linewidth}
         \centering
         \includegraphics[width=\linewidth]{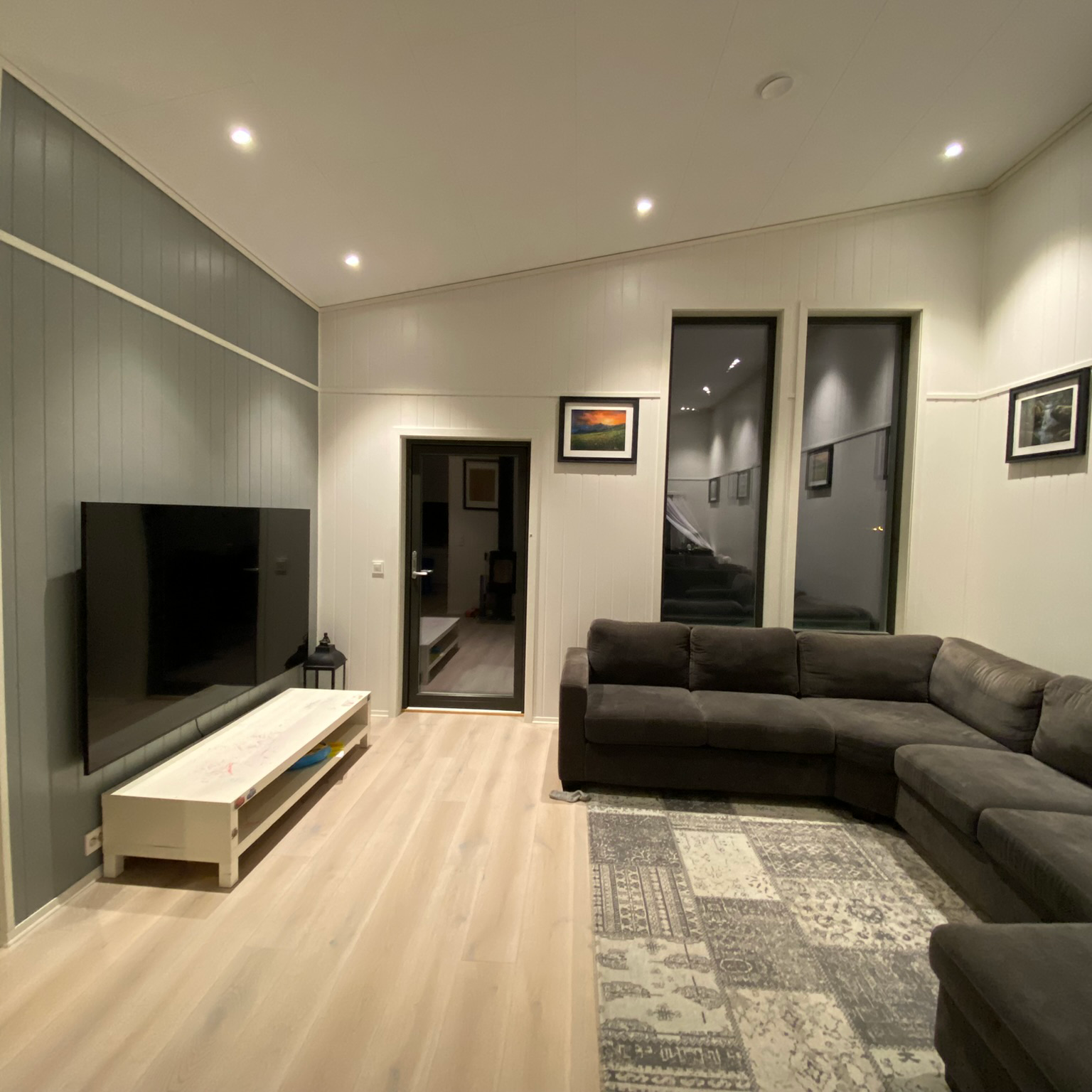}
         \caption{Real living room.}
       \end{subfigure}
        \caption{Comparison of furnished living room.}
        \label{fig:real_virtual_stue}
    \end{figure}

This work uses Unity Version 2020.3.16f1 with the Universal Render Pipeline version 10.5.1 (URP). Unity offers three different graphics rendering pipelines: the standard pipeline is a general-purpose tool with limited features and options; the High Definition Render Pipeline (HDRP) focuses on creating high-end graphics that are supported only by cutting-edge graphics cards and machinery; and URP is customizable and provides optimized graphics across a broad range of platforms \cite{unity_rp}. URP is preferred for this project because the DT is intended to run on low-end android VR platforms, such as the Oculus Quest 2. Additionally, the setup requires the following add-on packages from the Unity Asset Store:
    \begin{itemize}
    \setlength{\itemsep}{0mm}
        \item FBX Exporter version 4.1.1
        \item Oculus XR Plugin 1.9.1
        \item Post Processing 3.1.1
        \item Test Framework 1.1.27
        \item TextMeshPro 3.0.6
        \item Timeline 1.5.2
        \item Toolchain Win Linux x64 1.0.0
        \item Unity UI 1.0.0
        \item XR Legacy Input Helpers 2.1.8
        \item XR Plugin Management 4.1.0
        \item NuGet 3.0.2
    \end{itemize}
    \noindent Each of which supports some particular module of the setup, for example, the Oculus XR Plugin makes it easier to integrate the application in Unity with the Oculus Quest 2 with minimal setup. While NuGet is used to manage .NET-specific libraries. Unity also allows the possibility of bringing different 3D models together to live in the same environment. This could be a car, or furniture, as seen in Fig. \ref{fig:real_virtual_stue}. Note that when importing models in Unity, it is crucial to be aware of the relative scale/unit associated with the model. 
    
    \subsubsection{Other 3D objects}
    To create the paintings on the walls, we used 3DS Max to design the image frames and add images of the actual paintings. For other furniture pieces like the carpet, TV table, and sofa, we downloaded models from open-source websites such as Polantis \cite{polantis}, which provides free CAD and BIM 3D objects that resemble furniture from brands like IKEA. To ensure optimal performance, it's important to use 3D models with a low polygon count. High-polygon models can create a performance bottleneck, but it's possible to use 3DS Max to reduce the polygon count of downloaded models without compromising their appearance. It's worth noting that Unity 3D objects work on a meter scale by default, while external 3D models may be generated using other units of measurement like centimeters or inches. Therefore, it's important to be aware of the units of measurement used in external models and scale them appropriately before importing them into Unity. 
    
    \subsubsection{Terrain}
    Creating terrain in Unity based on a height map of Trondheim is not a straightforward task. However, the Norwegian government provides accurate height maps of Norwegian terrain with up to $1m\times1m$ resolution. One can apply for this data at Kartverket website, by selecting the specific region one wants to extract from the map of Norway and sending a request \cite{kartverket}. The data retrieved from Kartverket needs to be preprocessed from its original file format GeoTIFF to RAW, which is the input format that Unity supports for height maps. Here is an enumerated instruction list to convert heightmap data into a format usable in Unity: 
    \begin{enumerate}
        \item Apply for heightmap data from Kartverket.
        \item Import all the GeoTIFF files to the software QGis.
        \item Merge the GeoTIFF files into one common one using QGis merge option  (Raster $\rightarrow$ Miscellaneous $\rightarrow$ Merge) use output data type as UInt16 and click run.
        \item Export file by converting it to a BIL file, which is a variant of the RAW file type. This is done by navigating to (Raster $\rightarrow$ Convert $\rightarrow$ Translate).
        \item Click on Advanced Parameters, and add the following commands "-ot UInt16 -outsize 2049 2049" into "Additonal command-line parameters [optional]", 
    note outsize needs to be scaled to be in bit format +1 i.e. 1025,2049,4097 etc.
        \item Click on "Converted $\rightarrow$ Save to File" and change "save as type" to BIL files (*.bil).
        \item Click run, go to the folder where you saved the BIL file, create a copy and change filetype to ".raw".
        \item Finally upload the extracted RAW file into a Unity Terrain object using the height map property of the object.
    \end{enumerate}
    The 3D model of the house is placed in the location based on its real location in terms of longitude and latitude, where the top left corner of the terrain height map functions as the origin in a Unity grid, such that the house can be precisely positioned.
    \subsubsection{Orientation of the House} The Unity environment also simulates a day-night-cycle, including a highly accurate algorithm for the sun. Here orientation and altitude of the house are essential to synchronize the house's position relative to the sun simulation. For the orientation, the house needed to be rotated $203^{\circ}$ from the cardinal north direction of a compass. This was to align the front of the house correctly with the sun's movement. Since the setup of the standalone DT also includes a correct height map terrain of the outside environment in Trondheim, the house was elevated to the correct altitude, which is $211m$.
    
\subsection{Set-up for Descriptive DT}
At this stage, the physical house has been constructed, and a standalone DT has been developed. The next step is to refine the DT to match the physical house and equip it with a diverse class of sensors. A data stream is then established to convert the standalone DT into a descriptive DT. More details regarding the placement of the sensors can be found in Fig. \ref{fig:floorplan-sensors}, and are explained in detail in the following sections.

\subsubsection{Netatmo Weather Station}\label{Netatmo}
The Netatmo Weather Station is located in the living room, as shown in Fig. \ref{fig:floorplan-sensors}, and records data once every five minutes \cite{Netatmo}. The station measures temperature, humidity, CO2 concentration, noise levels, and pressure. While the weather station can be customized and equipped with many more sensors, it is only equipped to monitor the aforementioned data. All Netatmo endpoints are located at "https://api.netatmo.com". Before accessing the data, it is important to first make a POST request for an access token by adding the "/oauth2/token" endpoint to the Netatmo API link. The request must include the grant\_type, client\_id, client\_secret, username, password, and scope \cite{Netatmo_auth}. Once the request is successful, Netatmo grants access to the current state of the sensor via an access token.

\subsubsection{Philips Hue}
The Philips Hue sensors are located throughout the house, as depicted by the yellow dots in Fig.~\ref{fig:floorplan-sensors}. A total of 16 Philips Hue lights were used in the project. Information about the brightness, state of the light (on/off), colors, and any other relevant information about the lights can be retrieved in real-time using the Philips Hue API. The API that hosts the endpoints is "https://api.meethue.com". To refresh an access token, one must provide a valid refresh token as part of the request and perform a POST request to the "/v2/oauth2/token" endpoint. Assuming a valid access token, one can access the light data by making a GET request to the "/bridge/$<$USER\_ID$>$/lights" endpoint, where USER\_ID is an id granted by Philips Hue when an account is registered. Like with Netatmo, the access token must be provided as part of the GET request to obtain the data.

\subsubsection{Disruptive Technologies Sensors}\label{disruptive}
A total of twenty-eight sensors for disruptive technologies are positioned on all floors, as shown in Fig.~\ref{fig:floorplan-sensors}. Among them are 16 temperature sensors, with one placed outside the house to track external temperatures. There are also five proximity sensors, with one used to detect the opening and closing of the fireplace doors, while the remaining four monitor the open-close status of other doors. Four water sensors detect any water leakage in different locations, while three humidity sensors also record the temperature. Moreover, all sensors can function as touch sensors, meaning that if one is pressed, it could trigger an event. This provides significant freedom for experimentation, such as triggering an email or an action inside the disruptive technologies platform.

The authentication process for the sensors is similar to that of Netatmo but with an additional security layer. To obtain an access token, the email and key\_id provided at the time of account registration must be encrypted using the HS256 hash algorithm. The encrypted token must then be sent using a POST request to the "https://identity.disruptive-technologies.com/oauth2/token" endpoint, along with an assertion and grant\_type \cite{dt_auth}. With a valid access token in a GET request, various data can be obtained using the endpoint "https://api.disruptive-technologies.com/v2", with parameter variations provided in the documentation \cite{dt_docs}. Note that data is updated at a frequency of every fifteen or five minutes or when a significant change occurs.

\subsubsection{Weather Conditions}\label{weather_conditions}
The house's local weather data, including wind speed and direction, is obtained by making requests to the "api.met.no" weather forecast API, which is operated by the Norwegian Meteorological Institute. While there are several endpoints available, we use the "Nowcast 2.0" endpoint for our specific use case. To make a successful request to this endpoint, a User-Agent identity is required to identify the purpose of the request \cite{metAPI}. In order to visualize the wind direction and speed, a Unity vignette is created, which appears around the house. This visual effect is designed to show a wavy white line moving across the sky, indicating the wind's direction and velocity. The vignette is a useful tool for conveying the wind's speed and direction at any given moment, making it easier to understand the weather conditions outside.

\subsection{Set-up for Diagnostic DT}\label{chapter:diagnostic}
Once the data stream is established in the descriptive DT, data analysis should give additional insight into the asset and this should be presented to the end user in an easy-to-understand format. The set-up presented here for the diagnostic DT enables that. 
\subsubsection{Virtual Reality User Interface}\label{VRUI}
\begin{figure}[!htb]
    \centering
    \includegraphics[width=\linewidth]{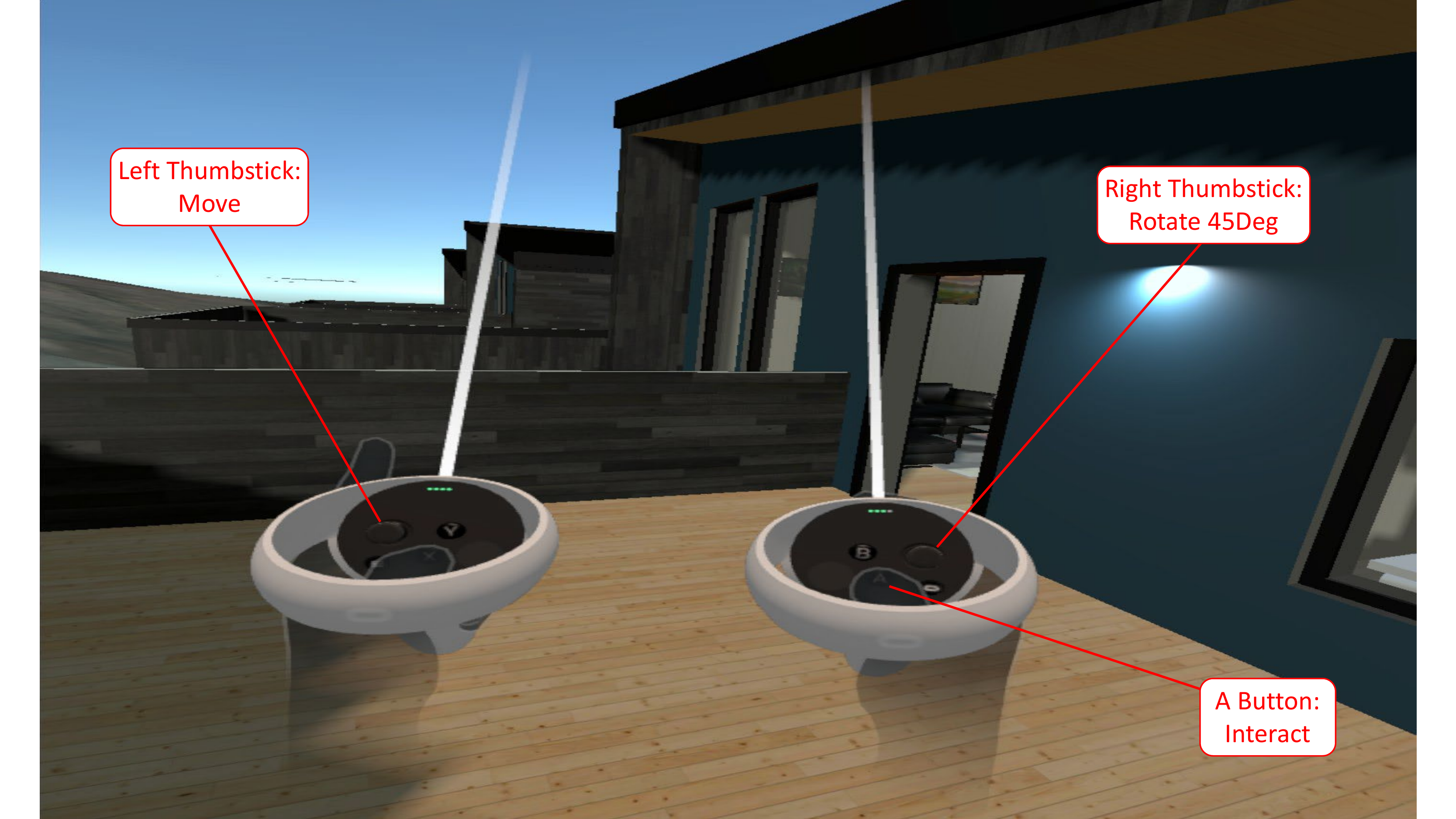}
    \caption{How to use the Oculus controllers in VR. If the play area is big enough, one is also able to physically move around or rotate without using the thumbsticks. Pointing at an interactable object with the right hand controller and clicking on the "A" button triggers events.}
    \label{fig:controllers}
\end{figure}

The VR setup uses an Oculus Quest 2, where a UI "tablet" is implemented on the user's left hand. This way, the user can move the menu in and out of sight. The main focus of the UI is to give the user a sense of empowerment, curiosity, and, most importantly, feedback, which are all ideas derived from the Octalysis Framework for Gamification \cite{octalysisGamification}. The empowerment and feedback come directly from the real-time control that a user feels when they can, for example, slide the time of the day and see the weather change to reflect reality. Furthermore, this responsive system triggers an exploratory curiosity in the user, making them want to seek out the remaining contents of the UI system. Recall that the diagnostic DT is mainly about monitoring and troubleshooting, meaning that sparking the user's interest in seeking diagnostic information is as important as presenting the information itself. 

\begin{figure*}[!htb]
   \begin{subfigure}{0.495\linewidth}
     \centering
     \includegraphics[width=\linewidth]{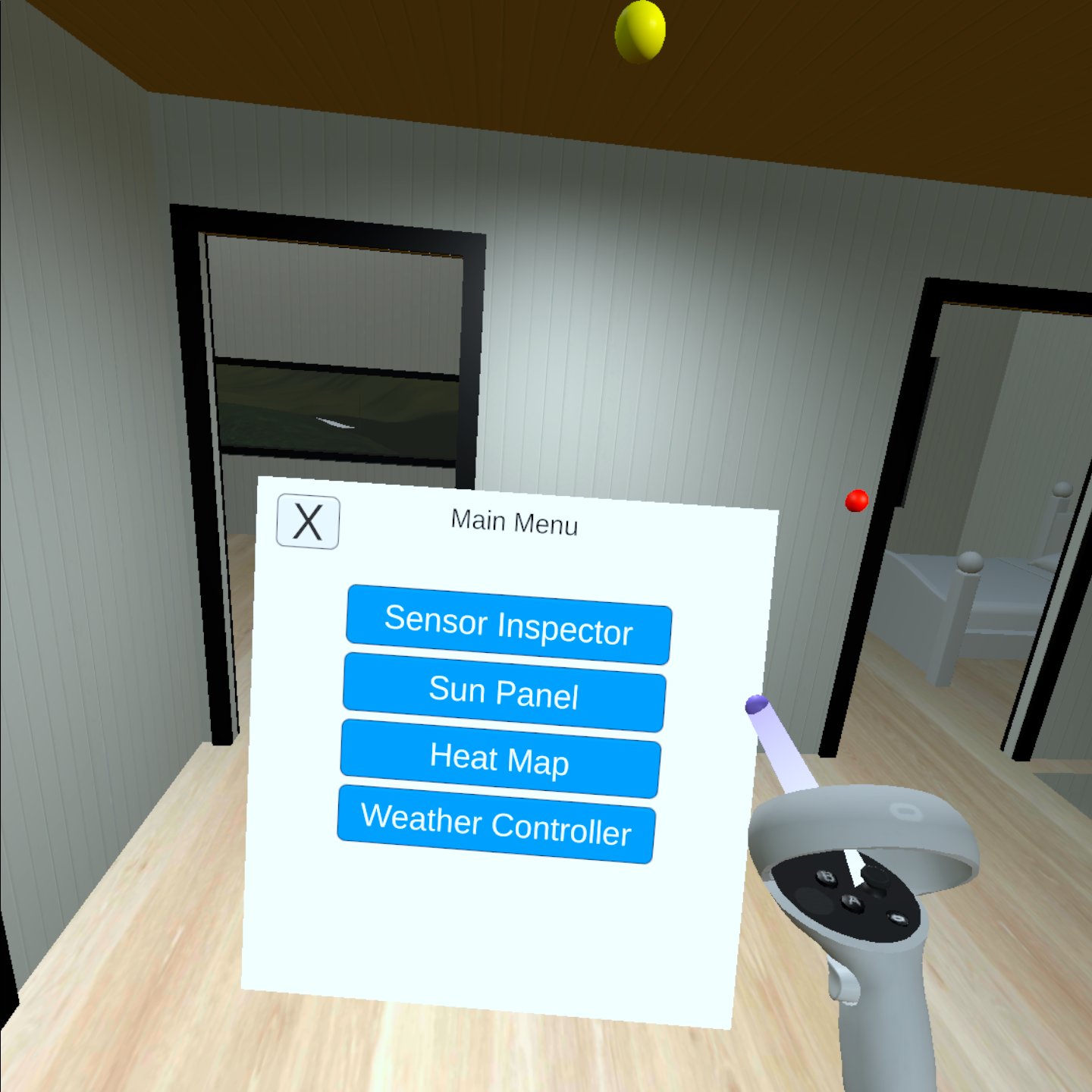}
     \caption{Main Menu.}
      \label{fig:ui_usagea}
   \end{subfigure}
   \begin{subfigure}{0.495\linewidth}
     \centering
     \includegraphics[width=\linewidth]{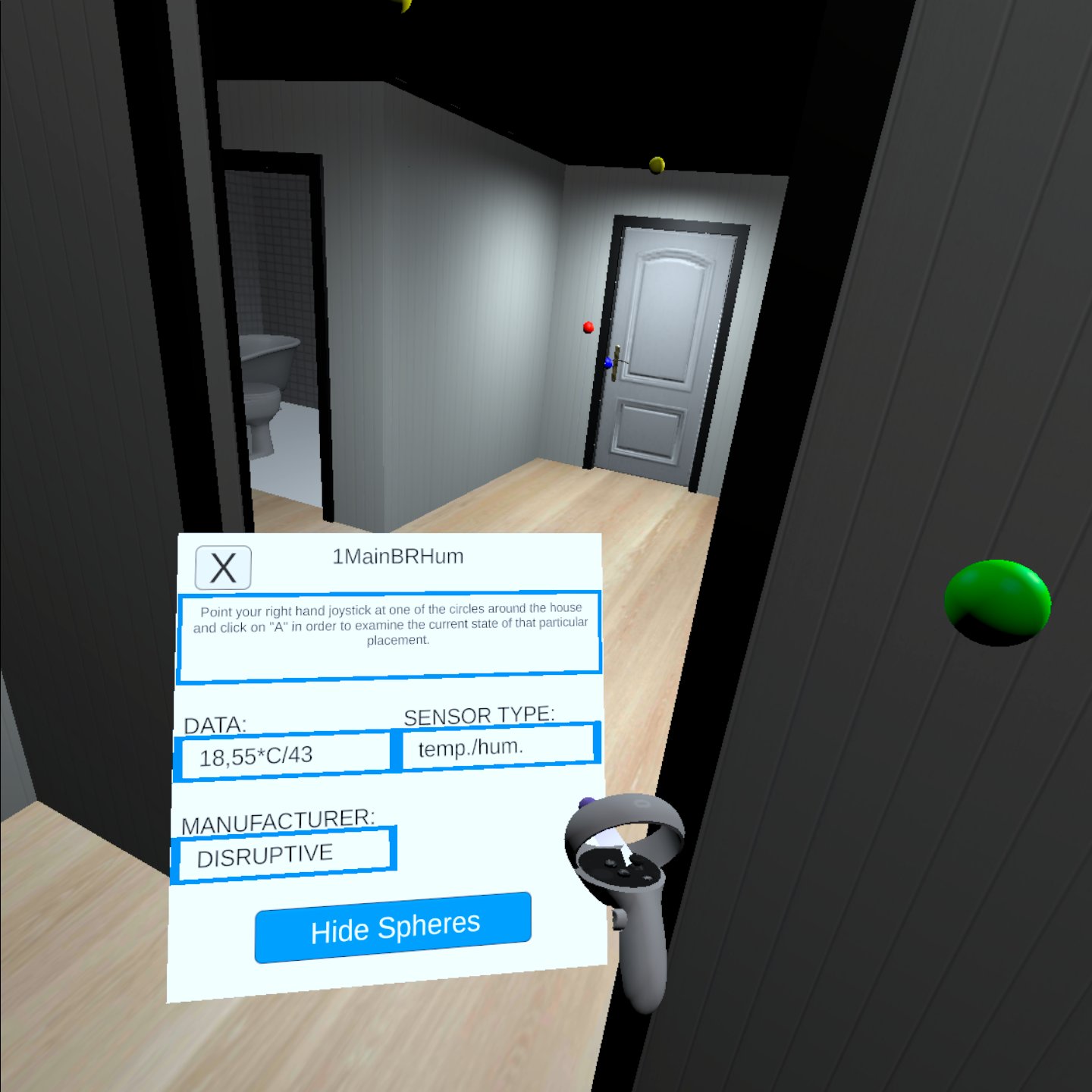}
     \caption{Sensor Inspector.}
      \label{fig:ui_usageb}
   \end{subfigure}\\
      \begin{subfigure}{0.495\linewidth}
     \centering
     \includegraphics[width=\linewidth]{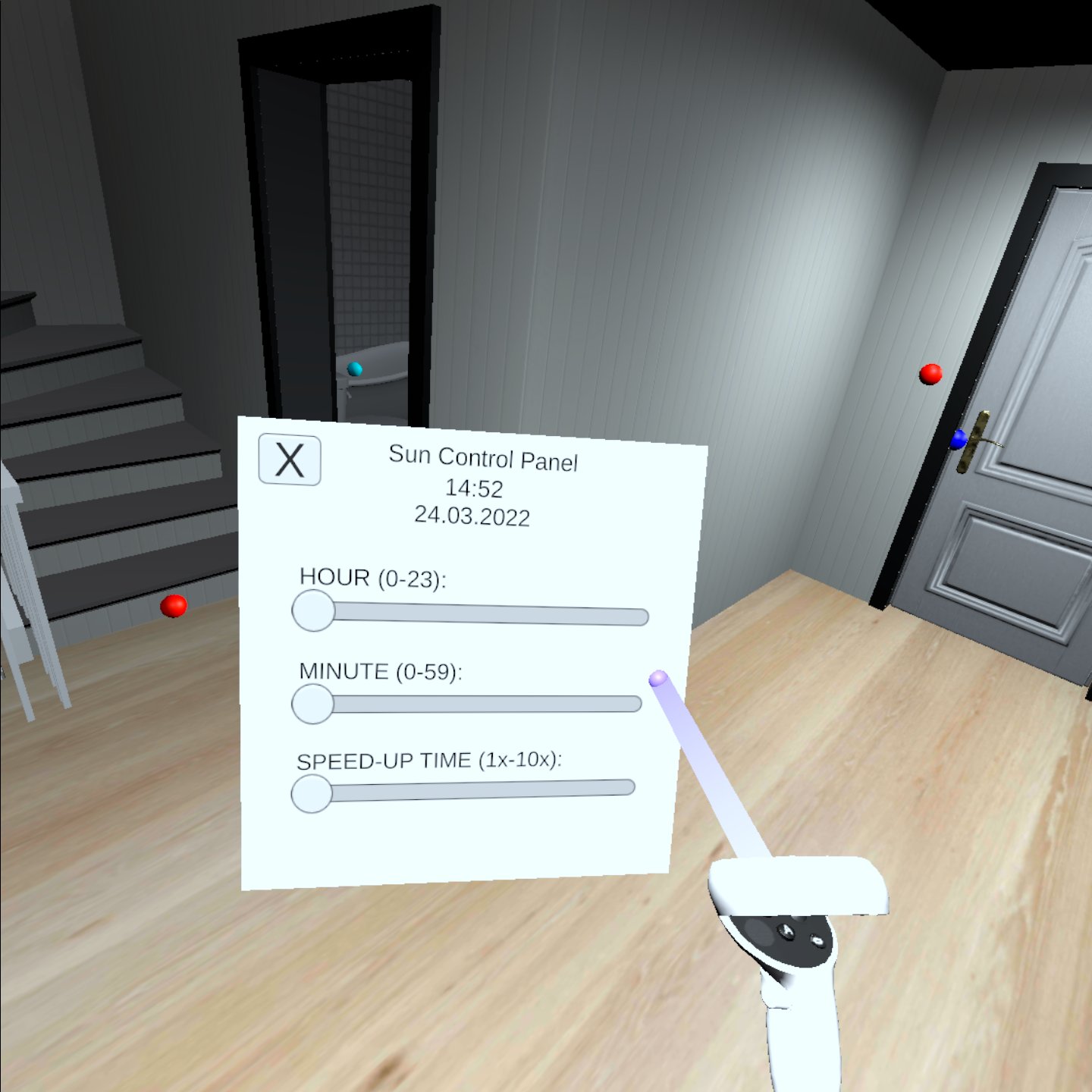}
     \caption{Sun Panel.}
      \label{fig:ui_usaged}
   \end{subfigure}
      \begin{subfigure}{0.495\linewidth}
     \centering
     \includegraphics[width=\linewidth]{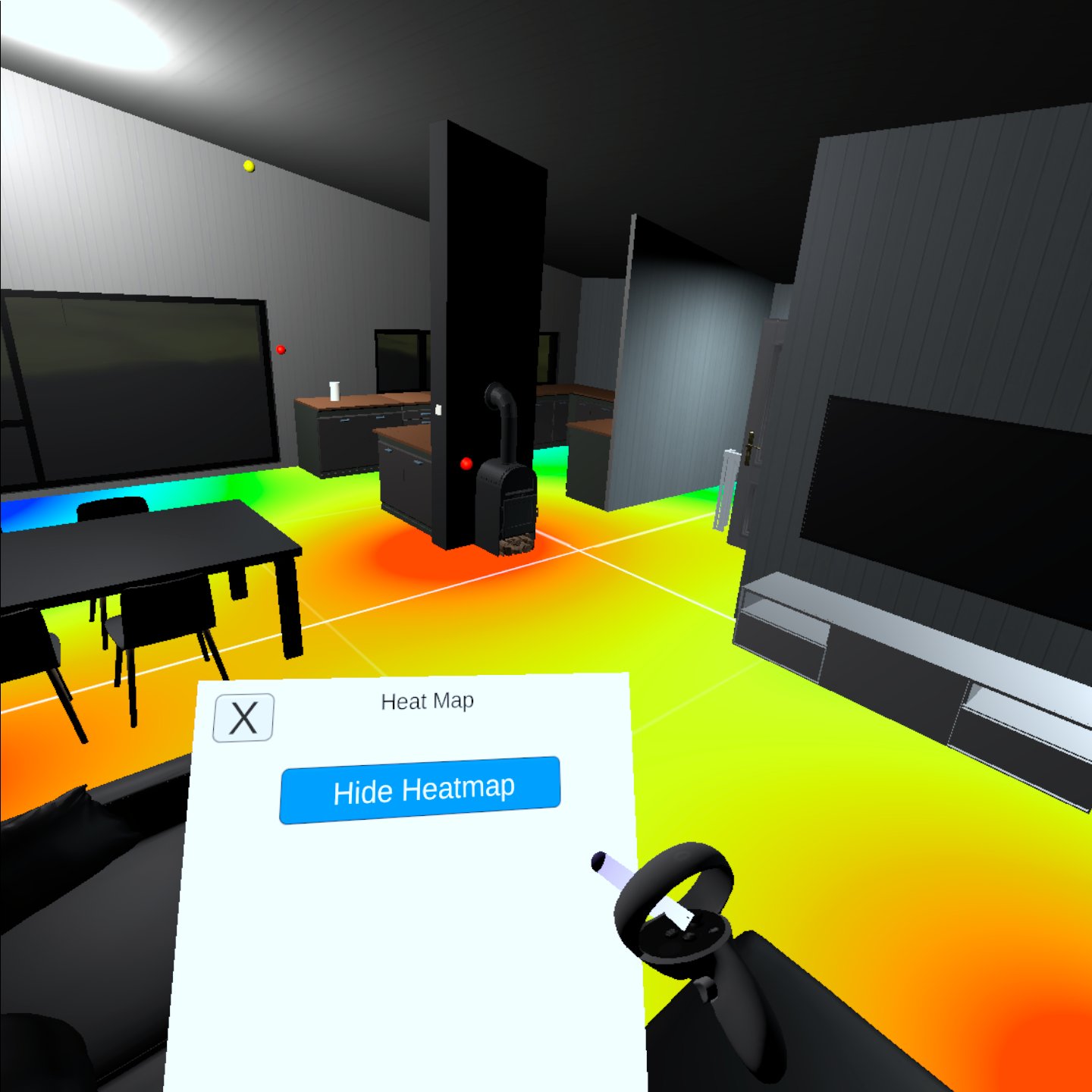}
     \caption{Heat Map.}
      \label{fig:ui_usagee}
   \end{subfigure}\\
     \caption{Navigating UI panels in Oculus Quest 2 VR.}
    \label{fig:ui_usage}
\end{figure*}

In Fig.~\ref{fig:ui_usage} one can see the enumeration of images featuring different navigation panels of the UI system. Fig.~ \ref{fig:ui_usagea} is the main menu which is at the center of all the other monitoring and troubleshooting systems for the diagnostic DT. Fig.~\ref{fig:ui_usageb} features the sensor inspector, providing critical information related to condition monitoring. By pointing at spherical probes around the house and scanning them with the "A" button of the right-hand controller, the sensor inspector displays relevant information relating to that probe. Real-time data from the specific sensor implemented for the descriptive DT is then revealed in the sensor inspector panel. Note that the probe positions are supposed to reflect the real three-dimensional positions of the sensors from Fig.~\ref{fig:floorplan-sensors}. Fig.~\ref{fig:ui_usaged} shows the UI for the sun control panel. The sliders can be used to simulate the lighting conditions depending on the time of the year. Fig.~\ref{fig:ui_usagee} demonstrates how data from multiple sensors can be combined to create a heat map that gives a better feeling of the indoor environment. 
\subsubsection{Temperature Heat Map} \label{diagnostic_section:temperatureHeatMap}
The temperature heat map is a visualization aspect of the diagnostic DT that conveniently presents valuable information. The heat map does not represent the temperature on the room's floor, but the temperature distribution of the temperature probes around the room. The heat map in Fig.~\ref{fig:temperature-forecast-vr} assumes that the heat distributes itself in a plane, and therefore is using a euclidean distance equation (Equation \ref{eq:weighted_path_finding}) for path-finding to render the temperature gradient radially outwards onto a Unity shader with the path limited to a reach based on $\frac{\alpha}{2}$. A shader is an object that communicates how to correctly color pixels onto a material in Unity \cite{UnityDocsShaders}. The implemented algorithm takes the current warmest and coldest temperature sensor measurements in a room and uses that interval to weight the radial output of the temperature gradient. The color mapping from temperature to color is the same as the one in Fig.~\ref{fig:diagnostic_temperature_fog}, i.e. if the minimum temperature is $16.5^\circ C$ and maximum $25.7^\circ C$ then that would be the blue and red colors respectively, and the other temperatures would fall within the color space in between.
\begin{align}\label{eq:weighted_path_finding}
    d\left( p,q\right)   = \frac{\alpha}{2}\sqrt {\sum _{i=1}^{n}  \left( q_{i}-p_{i}\right)^2 } 
\end{align}
\subsubsection{Fog Particle Effects}\label{fog_effects}
Another possible way to monitor the real-time data coming into Unity from the various data sources is by using the particle system in Unity.  In the current set-up, the red fog has its opacity adjusted based on if the real-time CO2 concentration is within a certain ppm interval, while the temperature is being mapped to color to represent the heat. The CO2 and temperature representations in Fig.~\ref{fig:diagnostic_DT_co2_results} and \ref{fig:diagnostic_temperature_fog} are implemented to visualize the Netatmo Weather Station placed in the living room, seen in Fig.~\ref{fig:floorplan-sensors}. The temperature reading is first transformed into an HSV value to be mapped into a color, and then from HSV into an RGB that is finally displayed as fog in Unity. It is important to note that the temperature intervals can be redefined, but for this example, the temperature interval is set to be between zero and 40 degrees Celsius. 
\subsection{Set-up for Predictive DT}\label{sec:Predictive}
This section is about the setup and methodology used to achieve a predictive DT. At this level, the DT can predict the system's future states or performance and support prognostic capabilities. To demonstrate the DT's predictive capability, we consider two cases, one related to the prediction of the inner state of the house in terms of the indoor temperature and another related to the external state in terms of the available solar potential. The reason for choosing the first case is that knowing the evolution of inside temperature can help develop cost and energy-effective control strategies. The second case is of great relevance for a country like Norway, where complex terrain can significantly affect solar exposure. 

This section also demonstrates two entirely different approaches to modeling that are relevant to DT-like technologies. One is pure data-driven modeling (DDM), while the other is physics-based modeling (PBM). DDM is effective when the physics governing a process is not entirely known, is computationally expensive to solve in real-time, or the values of physical parameters are not known accurately. PBM is more effective when the physics is known, or there is a need for generalization in unseen situations for which no data exists. We will use a DDM to predict the indoor second-floor temperature based on past experiences because the state-of-the-art building simulation model can not precisely describe the dynamics of the buildings, and the exact composition of the built material is unknown. On the other hand, we will use a PBM for sun position prediction because the equations governing the sun's movement are well-known and can be used to deterministically simulate any situation.
\subsubsection{Data Collection and Preprocessing}
First, the raw time series data (temperature, humidity, proximity) from Disruptive Technologies sensors are aligned and sampled at five-minute intervals to generate a  multivariate timeseries corresponding to 90 days, out of which 70 days were used for training, 10 for validation, and 10 for testing. The training of the ML models is done on a computer with the following specifications: Intel(R) Xeon(R) Gold 6140M CPU @ 2.30GHz for a CPU and NVIDIA Quadro RTX 5000 for GPU. The GPU is mainly used to speed up the training of the LSTM utilizing NVIDIA's CUDA library \cite{cudaToolkit, cuDNN}.
\subsubsection{Forecasting and Prediction Pipelines}
\begin{figure*}[!htb]
    \centering
    \includegraphics[width=\linewidth]{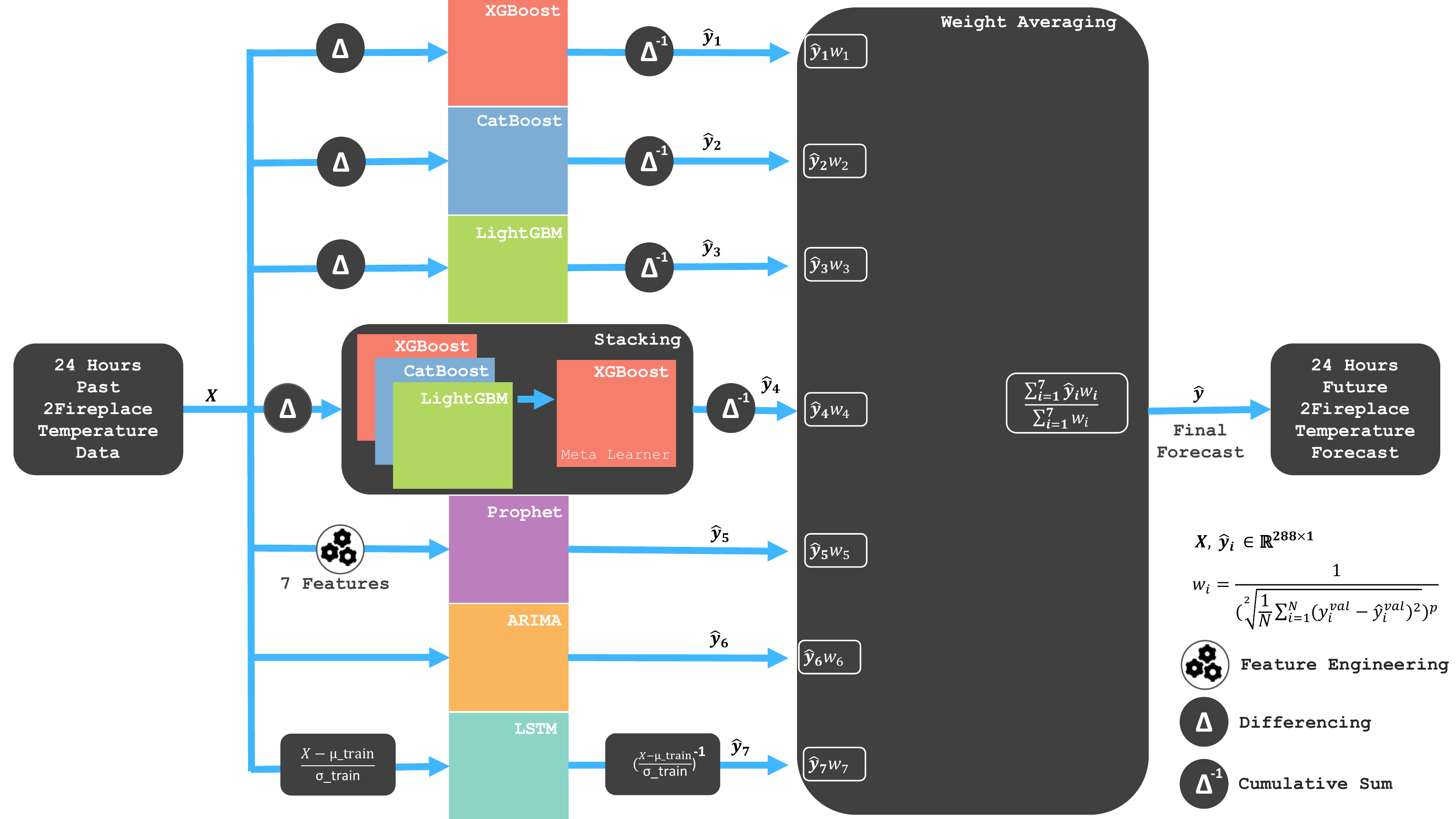}
    \caption{High level view of the forecasting pipeline.}
    \label{fig:forecasting_pipeline}
\end{figure*}

\begin{figure*}[!htb]
    \centering
    \includegraphics[width=\linewidth]{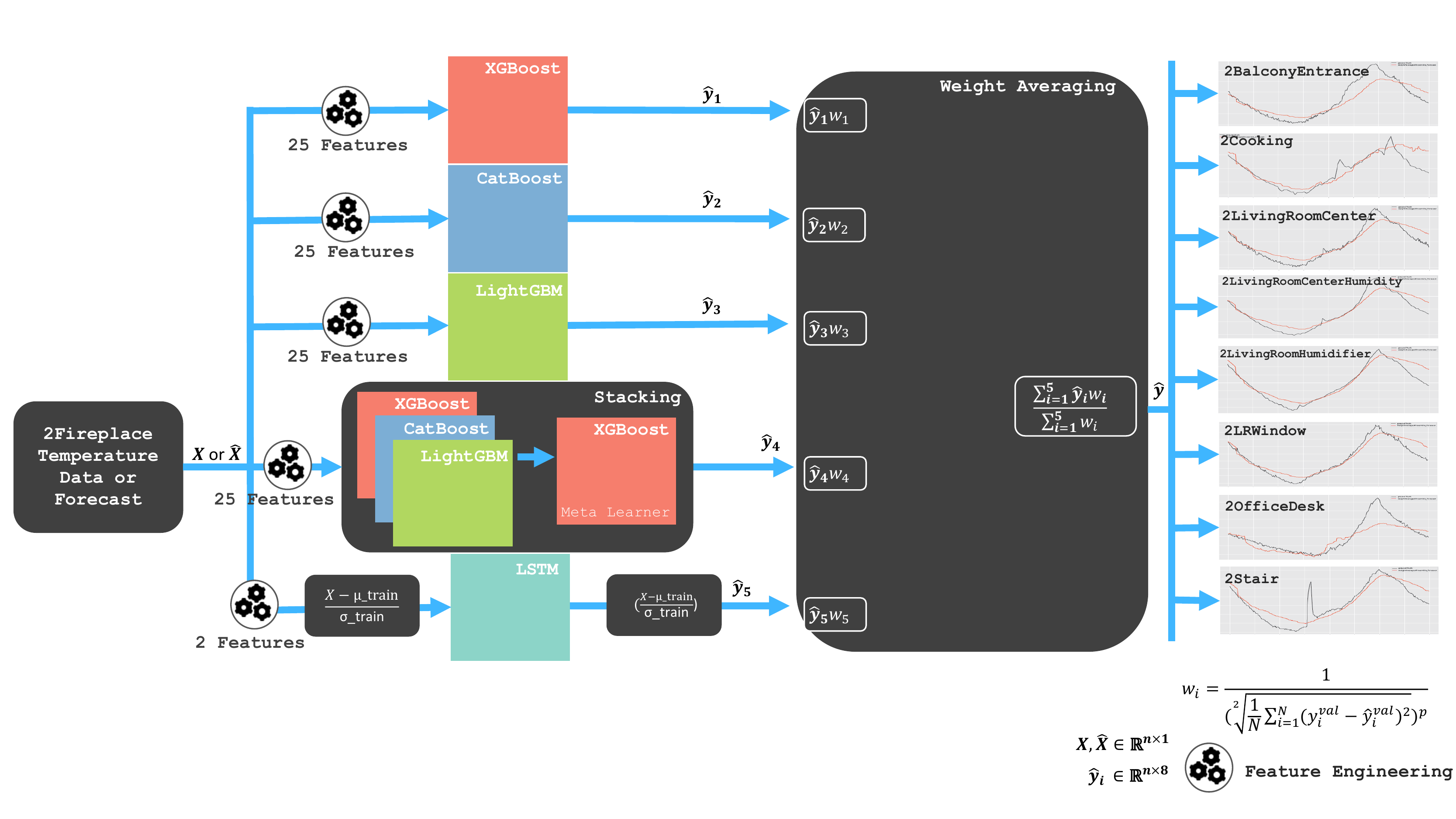}
    \caption{High level view of the prediction pipeline.}
    \label{fig:prediction_pipeline}
\end{figure*}

As seen in Fig.~\ref{fig:forecasting_pipeline}, a weight-averaging ensemble is used as the final model. Gradient boosting machines and the stacked model worked very well with a differencing transform of the target, while the LSTM model performed better with standard data normalization. For the Prophet model, providing extra features increased performance. Also based on the RMSE validation score, hyperparameters of every single model are optimized using the Optuna Python library. The final forecasts on the validation and test set are achieved by using the weighted average based on the validation RMSE of each model. All the forecast models except for LSTM are setup as incremental forecasting models, mainly because none of them support an instant multi-output target. 
In the prediction model, the data is preprocessed in a similar fashion to the forecasting model. In practice, the prediction pipeline in Fig.~\ref{fig:prediction_pipeline} can be used in two ways. First, it can be used to predict the potential state of an out-of-commission sensor, assuming that the fireplace sensor in the room is still operational with data $\mathbf{X}$. Or it can be used to fill in the eight forecasted temperature sensors using the forecasted output $\hat{\mathbf{y}}$ of Fig.~\ref{fig:forecasting_pipeline}, where in the context of Fig.~\ref{fig:prediction_pipeline} the forecasted fireplace data is denoted as input $\hat{\mathbf{X}}$. Therefore assuming a temperature sensor deployed in the second floor dies after a few years, instead of replacing it, these sensors can be accurately predicted based on the fireplace sensor. Otherwise, the prediction model can be seen as an extension of the forecasting model. 
\subsubsection{Hyperparameters of the LSTM}
The neural network is simply constructed of a LSTM layer, followed by one-dimensional global max pooling into a dense output layer. The following hyperparameters are used, hyperparameters not explicitly mentioned here can be assumed to be the default of the Keras/Tensorflow library:
\begin{itemize}
\setlength{\itemsep}{0mm}
    \item Learning Rate: $0.01$
    \item Batch Size: $4$
    \item Optimizer: ADAM
    \item Epochs: $5$
    \item Loss Function: Mean Squared Error
    \item LSTM Units: $64$
    \item Dense layer neurons: $64$
\end{itemize}
\subsection{Set-up for Prescriptive Digital Twin}\label{subsec:prescriptive}
This section is about the setup and methodology used to achieve the prescriptive DT. The prescriptive DT is the fourth capability of a DT pipeline. In this level, the DT is able to make recommendations based on what-if? scenarios, and through that support uncertainty quantification. What-if scenarios in our case can be potential weather forecasts or historical weather conditions in the past that may have led to certain behaviors. 

Behaviors can certainly be used to motivate future recommendations, given if such a situation would again occur sometime in the future. The recommendations can be used to give experts a decision support system, but for our case, such a system might be used for uncertainty quantification of the unpredictable parts of the future forecasts, that have been modeled and highlighted in the predictive DT. 

\subsubsection{Collaborative Filtering}
\begin{figure}[!htb]
    \centering
    \includegraphics[width=\linewidth]{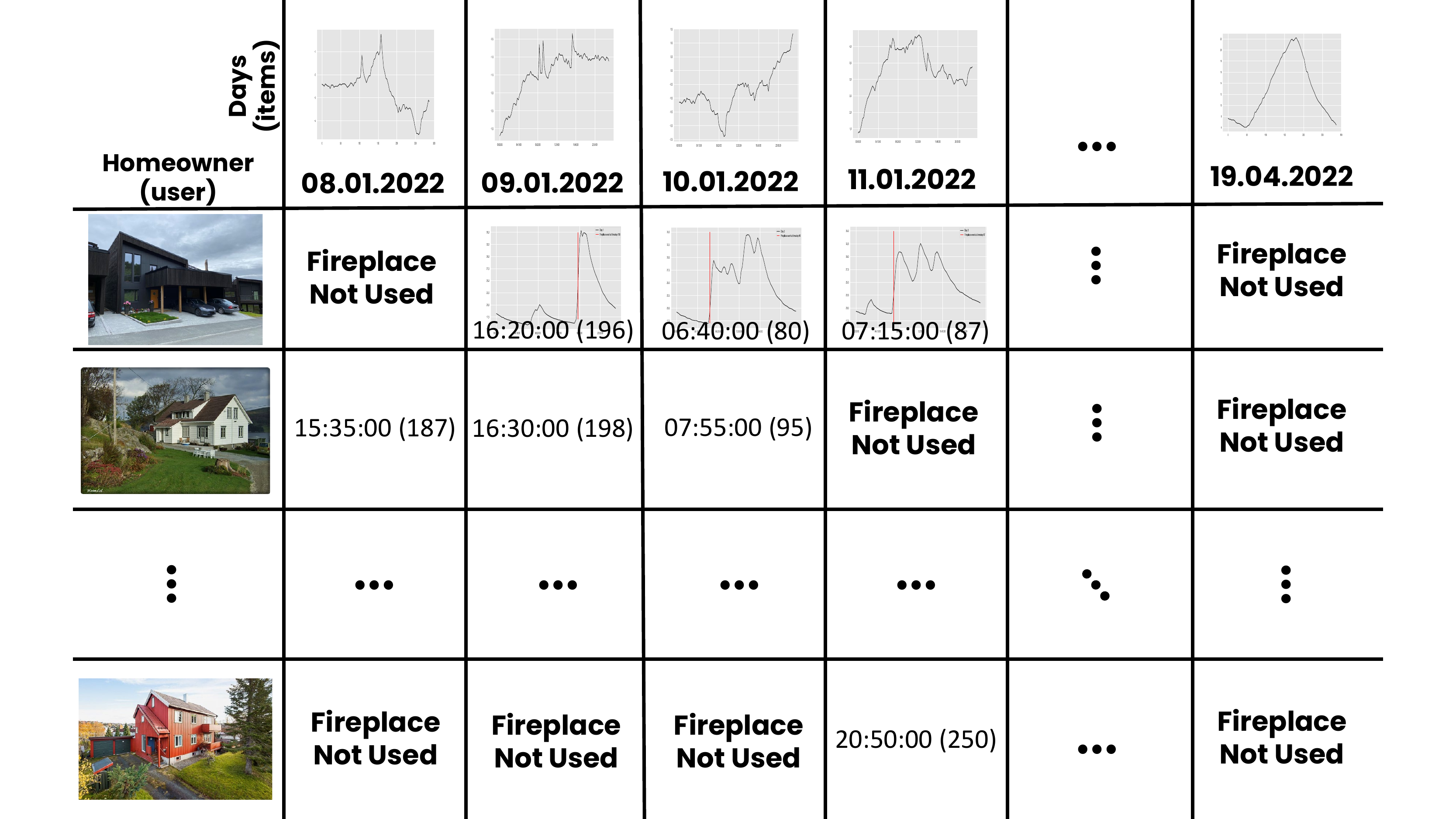}    
    \caption{Collaborative filtering matrix. Note that other than the house that we have collected data from in row one, the remaining houses and their data is artificial for the sake of demonstrating user based collaborative filtering. }
    \label{fig:UBCF_matrix}
\end{figure}

There are only eight recorded fireplace events for 102 days. This is a very small dataset to work from, making it hard to conclude anything in particular about the user's fireplace lighting behavior. Furthermore, no data has been collected from other homeowners and there are therefore no other users, from whom we can obtain our recommendations to the specific user. Hence there is insufficient data, as the sparsity of data comes from that most of the days the fireplace is simply not used. The motivation for demonstrating this particular method comes from its potential to become valuable, as more data is collected from this house and other locations.
The collaborative filtering for fireplace lighting events is set up with $\mathbf{U}$ homeowners (users) as the rows and $\mathbf{V}$ days (items) as columns. Note each day has specific information about the outdoor temperature of that certain day. In the collaborative filtering matrix from Fig.~\ref{fig:UBCF_matrix}, the fireplace timesteps for every single day are extracted and used as a row for a specific user. The timestep for when the fireplace was approximately turned on is inferred by finding the biggest difference between two points on a specific day. The timestep for a lit fireplace for a specific day can be seen in the first row, as a red vertical line in Fig.~\ref{fig:UBCF_matrix}.

\begin{figure*}[!htb]
    \includegraphics[width=1\linewidth]{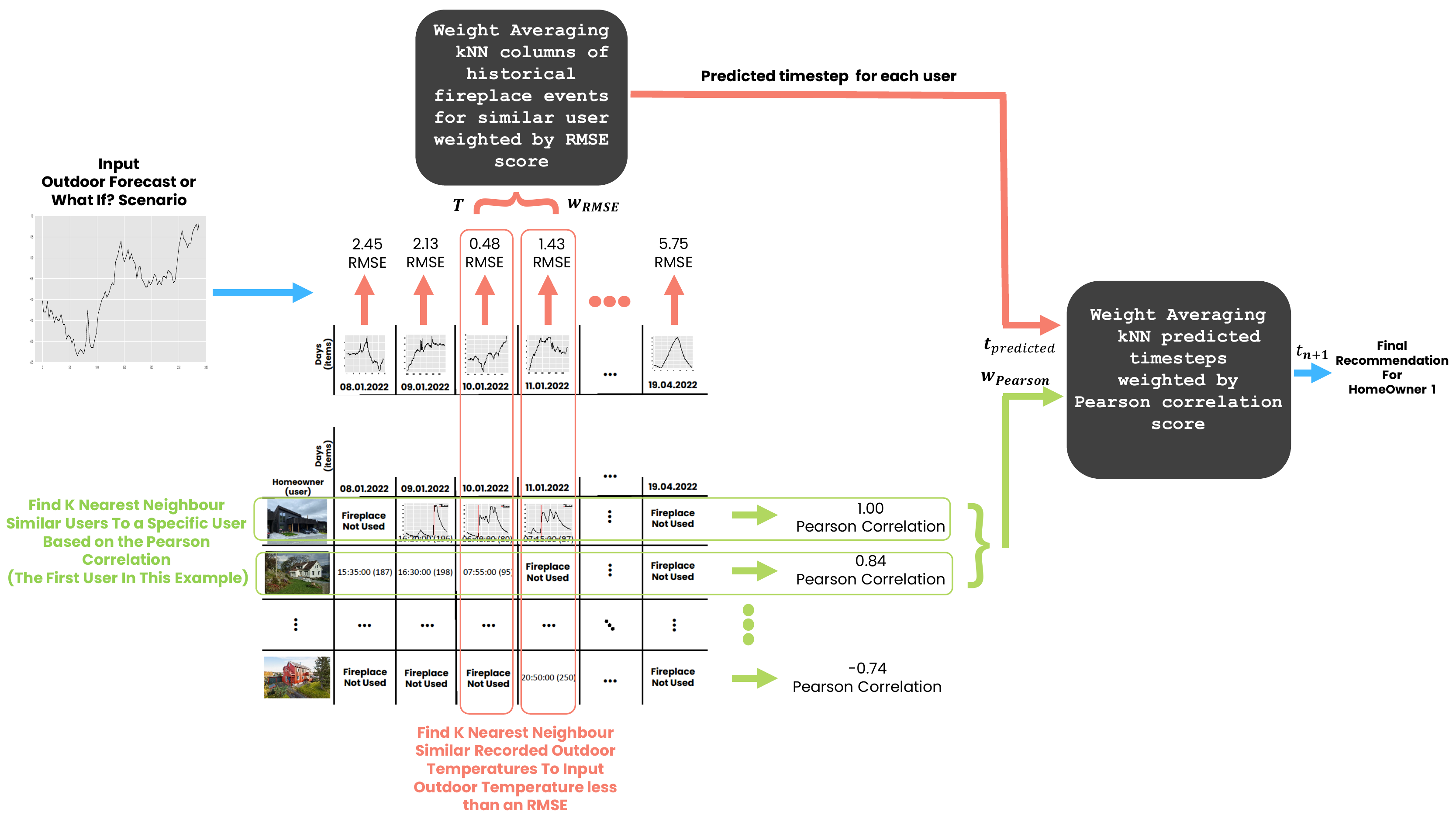}
    \caption{User based collaborative filtering pipeline.}
    \label{fig:UBCF_pipeline}
\end{figure*}

The user-based collaborative (UBCF) pipeline as implemented is visualized in Fig.~\ref{fig:UBCF_pipeline}. Given that an unknown input outdoor temperature forecast or scenario is given to the pipeline, first, the RMSE between the input and each column-specific outdoor temperature is calculated to find the most similar scenario to our input. For simplicity all users are assumed to have experienced the same outdoor temperature in Trondheim, therefore for users outside of Trondheim, RMSE specific to that user needs to be calculated. The remaining pipeline would still work the same way. Resulting in the RMSE weight vector $\mathbf{w}_{RMSE}$ seen in Equation \ref{eq:weight_averaging_RMSE_UBCF_rmse}. 

The recommendation pipeline seen in Fig.~\ref{fig:UBCF_pipeline} is supposed to be for the first user. In parallel with finding $\mathbf{w}_{RMSE}$ for the user, the Pearson correlation of all the fireplace lighting event actions of said user is calculated against the other users in the population, assuming these exist. Then the highest correlated users based on some threshold are extracted as a weight vector $\mathbf{w}_{pearson}$ seen in Equation \ref{eq:weight_averaging_RMSE_UBCF_pearson}. Where $\chi_{pq}$ is the set of days both users $p$ and $q$ that have fireplace behavioral data.

Knowing the most similar scenarios to the input scenario and the most correlated users to the user we wish to make recommendations for, now we extract a matrix denoted $\mathbf{T}$. This matrix is the region where both the green and red rectangles cross as seen in Fig.~\ref{fig:UBCF_pipeline}. Each row of said matrix is weighted by the RMSE score of the specific user in a weighted average. This gives the predicted fireplace lighting behavior of each individual user, given that specific scenario. That particular prediction vector $\mathbf{t}_{predicted}$ is then weighted by the most similar users to the particular user we are recommending, to produce the final recommendation time to light the fireplace $t_{n+1}$.

\begin{equation}
    \mathbf{w}_{RMSE}=\left[ \begin{matrix}
    \frac{1}{(\sqrt{\frac{1}{N} \sum^N_{k=1}(T^{item1}_k - T^{input})^2})} & \\ 
    \hdots \frac{1}{(\sqrt{\frac{1}{N} \sum^N_{k=1}(T^{itemV}_k - T^{input})^2}} & 
    \end{matrix}\right]
    \label{eq:weight_averaging_RMSE_UBCF_rmse}
\end{equation}

\begin{equation}
    \mathbf{w}_{pearson} =\left[ \begin{matrix}
    \frac{\sum_{j \in \chi_{1i}} (t_{1j}-\Bar{t}_1)(t_{ij}-\Bar{t}_{i})}{\sqrt{\sum_{j \in \chi_{1i}} (t_{1j}-\Bar{t}_{1})^2}\sqrt{\sum_{j \in \chi_{1i}} (t_{ij}-\Bar{t}_{i})^2}} & \\ 
    \hdots \frac{\sum_{j \in \chi_{1U}} (t_{1j}-\Bar{t}_1)(t_{Uj}-\Bar{t}_{U})}{\sqrt{\sum_{j \in \chi_{1U}} (t_{1j}-\Bar{t}_{1})^2}\sqrt{\sum_{j \in \chi_{1U}} (t_{Uj}-\Bar{t}_{U})^2}} & 
    \end{matrix} \right]
    \label{eq:weight_averaging_RMSE_UBCF_pearson}
\end{equation}
 
\subsection{Set-up for Autonomous Digital Twin}
In the current work we have not demonstrated the autonomous capability of the DT. 

\section{Results and Discussions}
\label{sec:resultsanddiscussions}
In this section, we aim to showcase the potential and value of various types of DTs. To illustrate each capability level, we begin with a scenario that presents specific challenges, and then use the DT to demonstrate how these challenges can be addressed effectively. 
\subsection{Standalone DT}
\begin{figure}[!htb]
    \centering
    \includegraphics[width=\linewidth]{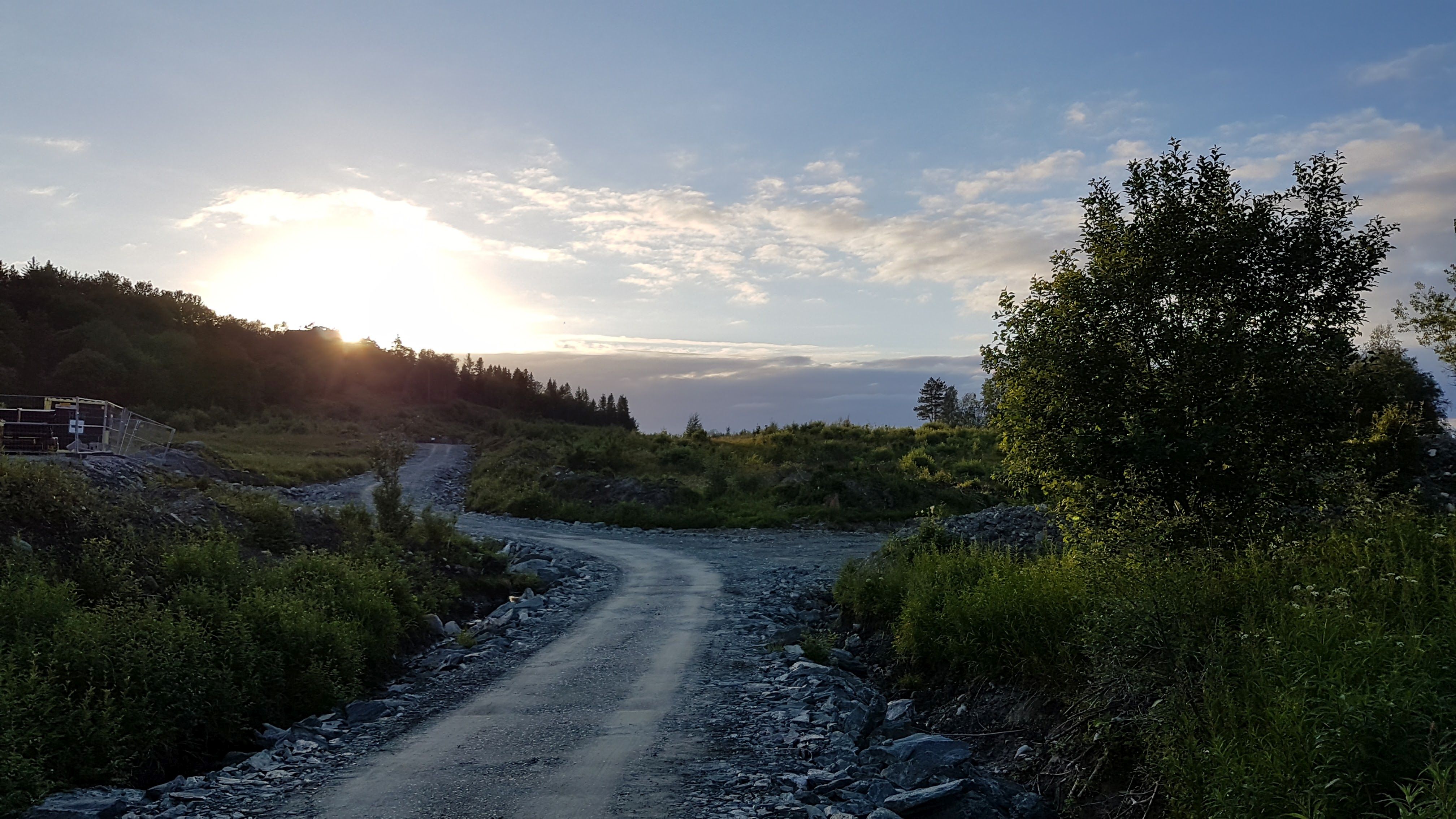} \caption{A visit to the site before the deal of the house was finalized.}
    \label{fig:sitebeforeconstruction}
\end{figure}

\textit{Scenario:} A potential homebuyer is interested in purchasing a yet-to-be-constructed house. The real estate company provides the buyer with a tour of the construction site (Fig.~\ref{fig:sitebeforeconstruction}) and several 2D sketches (Fig.~\ref{fig:floorplan-sensors}) depicting the different levels of the future house. Unfortunately, the provided documents do not offer any insight into how the neighborhood will look once the construction is complete. Despite the lack of information, the buyer commits to the purchase. Later on, the buyer is invited to customize the house, but once again, lacks insight into how their choices will look and feel in the real world. In this scenario, a virtual tour using a standalone DT could provide the buyer with the necessary information to make a more informed decision, and improve communication between the seller and buyer.

\textit{Solution:} The image provided by the real estate agent in Fig.~\ref{fig:sitebeforeconstruction} only provides a limited idea of the site before construction. However, with a standalone DT, it is possible to create a more immersive experience of the interior and exterior environment. This would allow the buyer to make informed decisions about customizing the interior, estimating solar potential, and assessing the recreational activities in the area. By using a DT, the buyer can gain a better understanding of the property and make more informed decisions.  

\begin{figure}
\begin{subfigure}{0.475\linewidth}
  \centering
  \includegraphics[width=\linewidth]{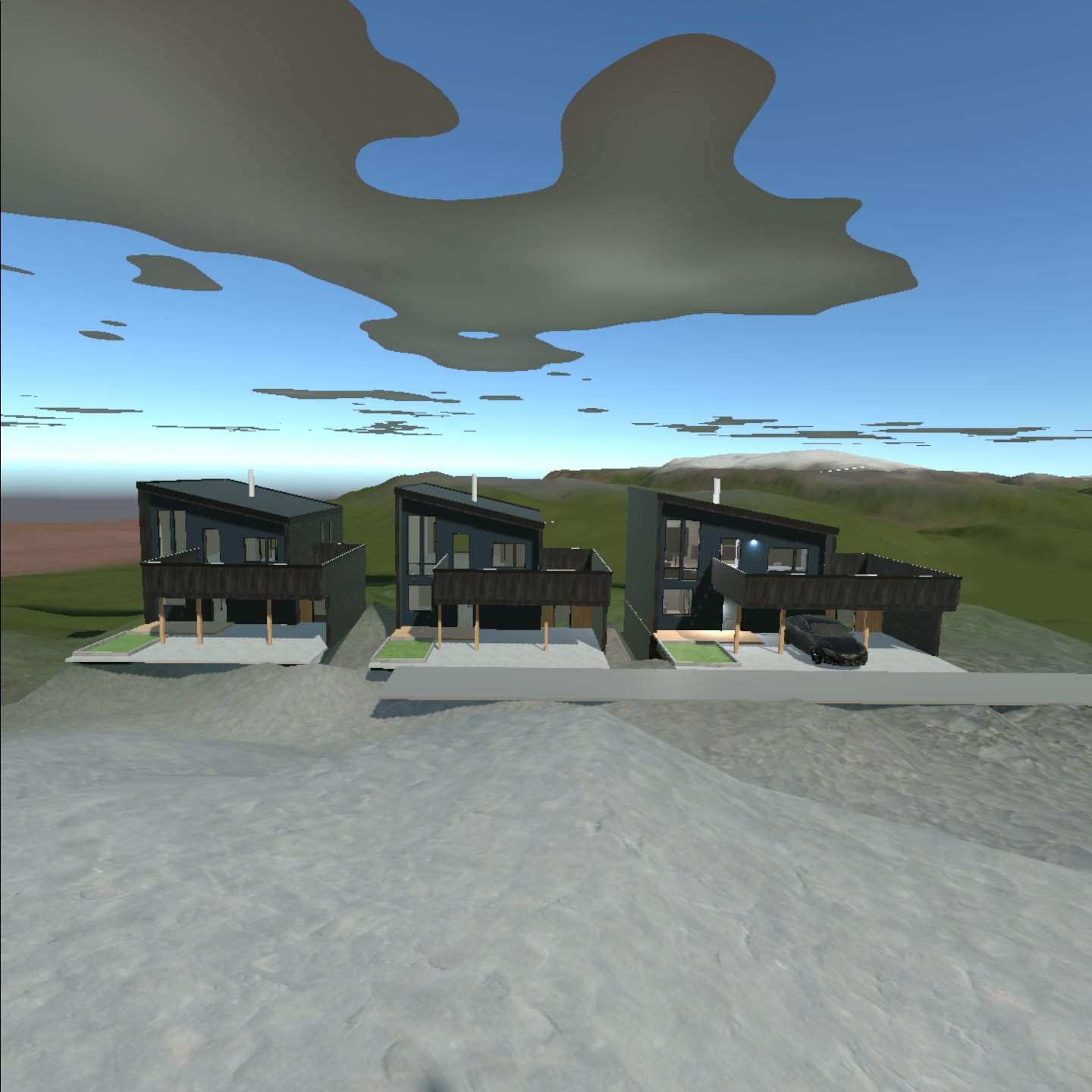}
  \caption{Front view.}
  \label{fig:sfig1}
\end{subfigure}
\begin{subfigure}{.475\linewidth}
  \centering
  \includegraphics[width=\linewidth]{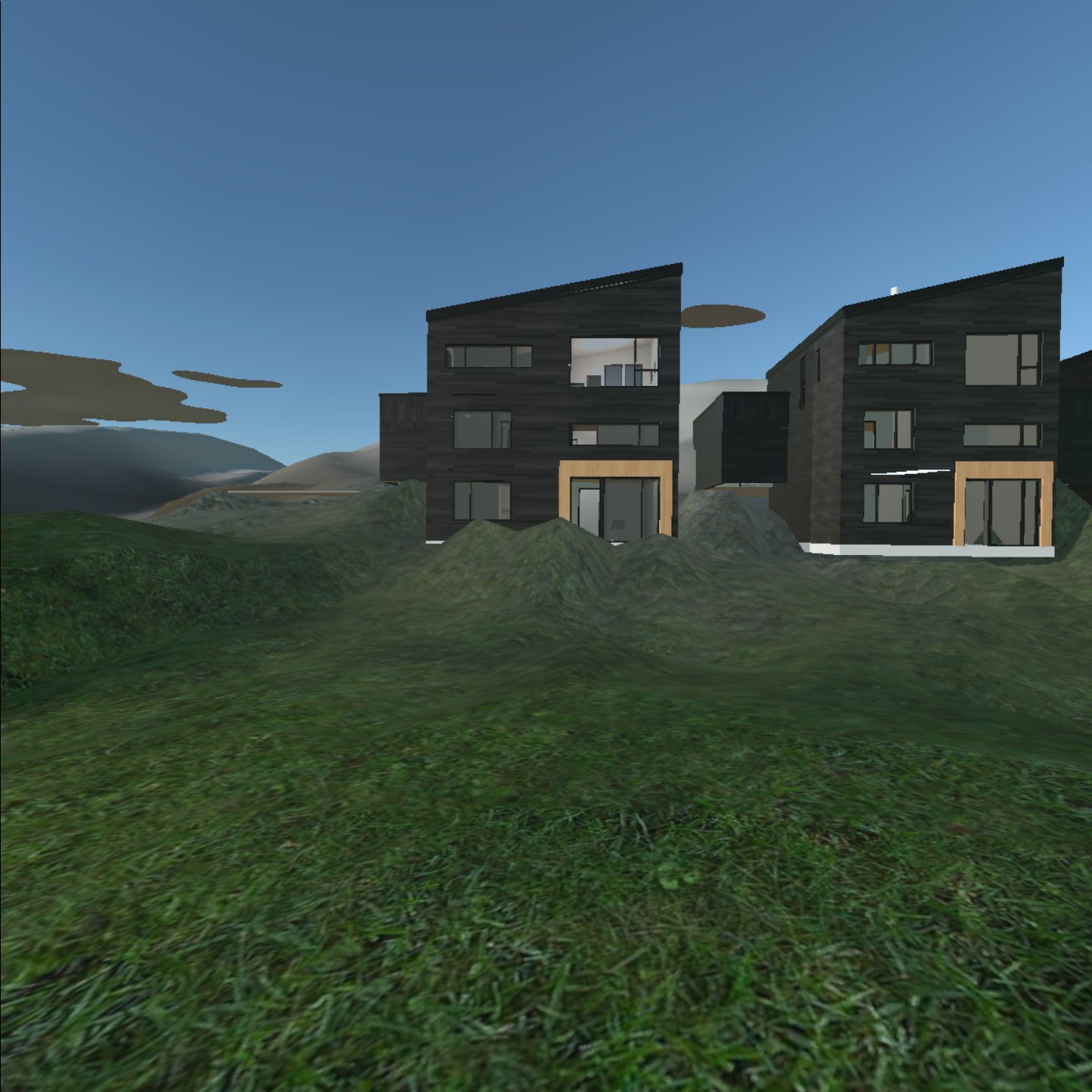}
    \caption{Back view.}
  \label{fig:sfig8}
\end{subfigure}\\
\begin{subfigure}{.475\linewidth}
  \centering
  \includegraphics[width=\linewidth]{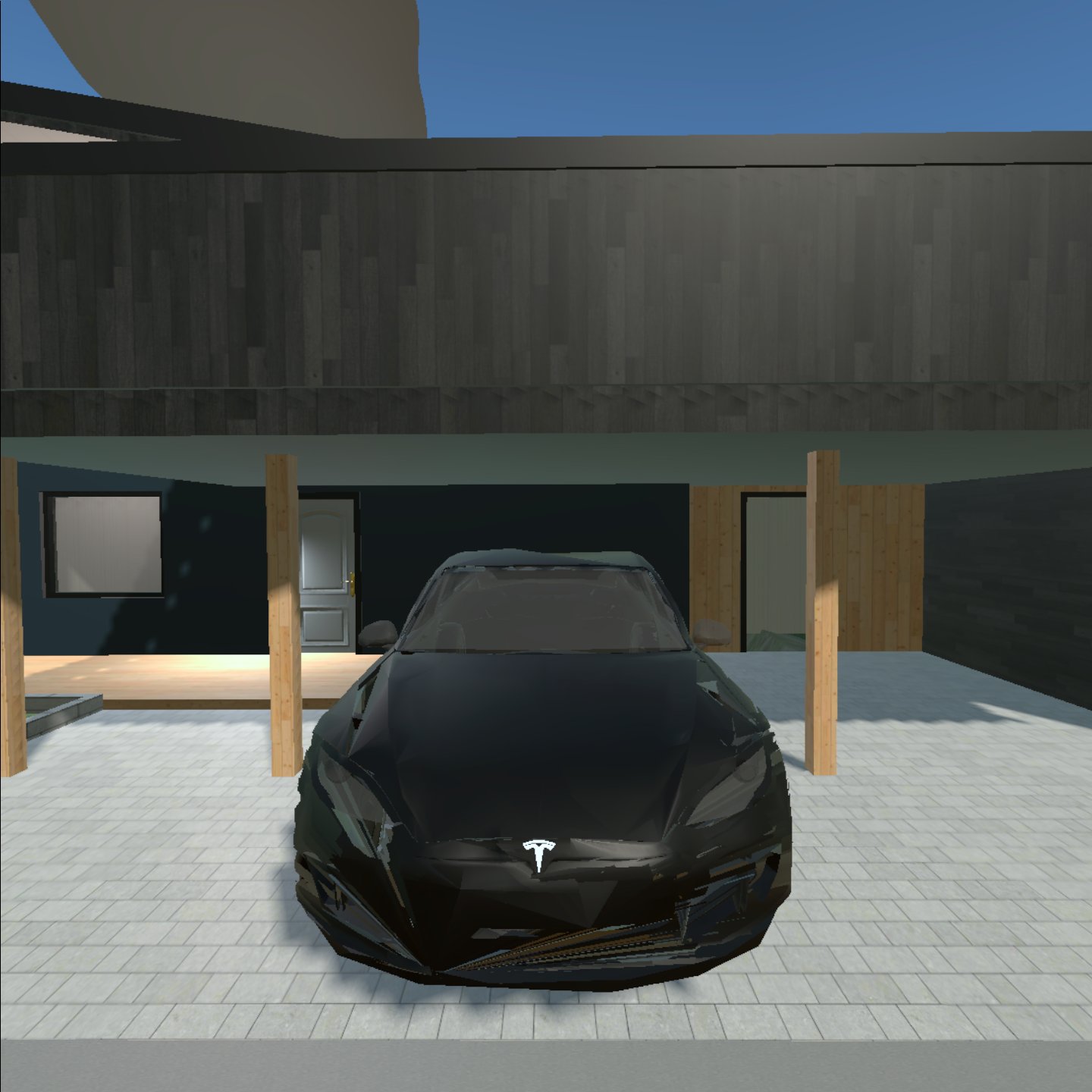}
  \caption{Driveway.}
  \label{fig:tesla2}
\end{subfigure}
\begin{subfigure}{.475\linewidth}
  \centering
  \includegraphics[width=\linewidth]{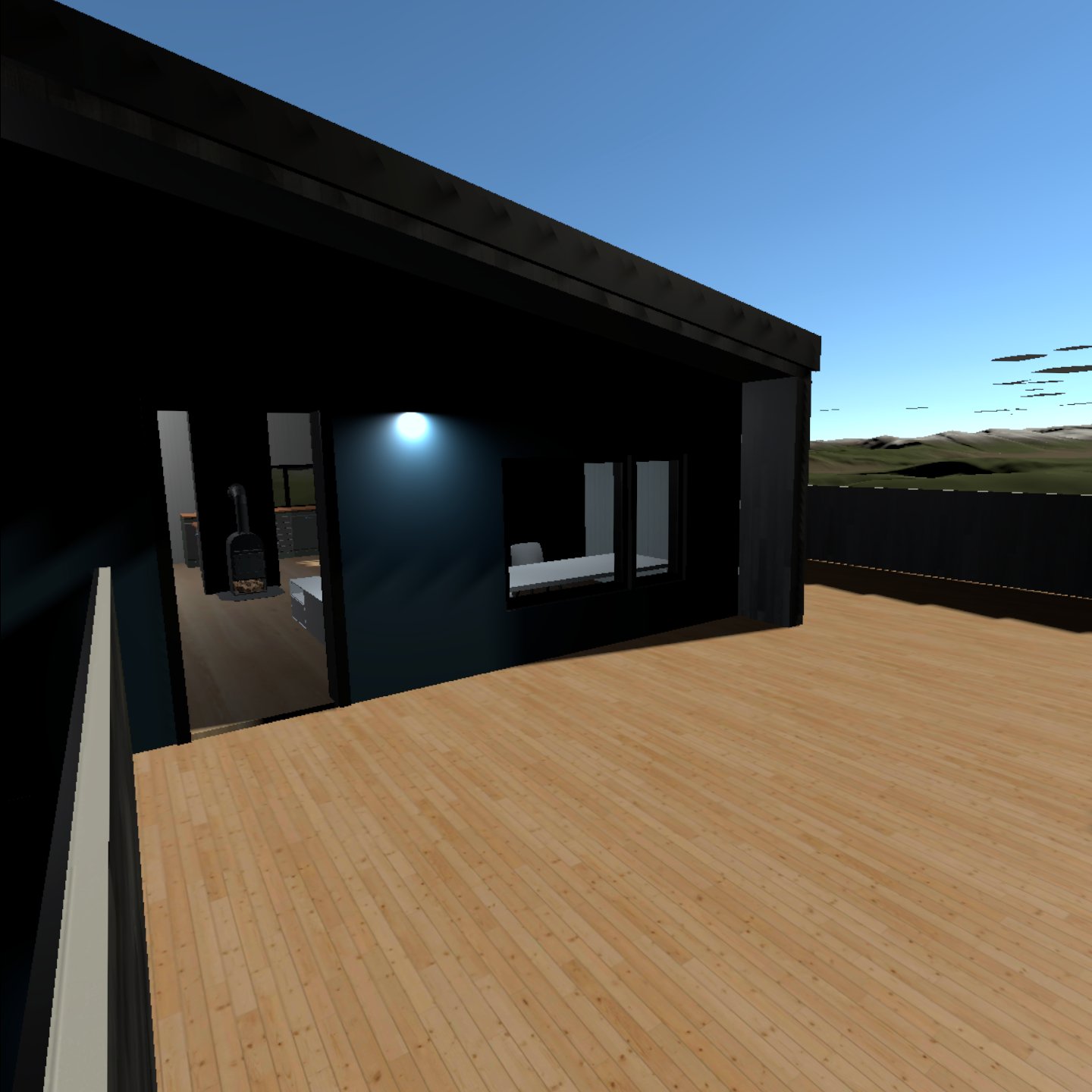}
    \caption{Balcony.}
  \label{fig:balcony}
\end{subfigure}
\caption{Demonstration of the external environment using a standalone DT.}
\label{fig:standaloneDemo_exterior}
\end{figure}

\begin{figure}
\begin{subfigure}{.495\linewidth}
  \centering
  \includegraphics[width=\linewidth]{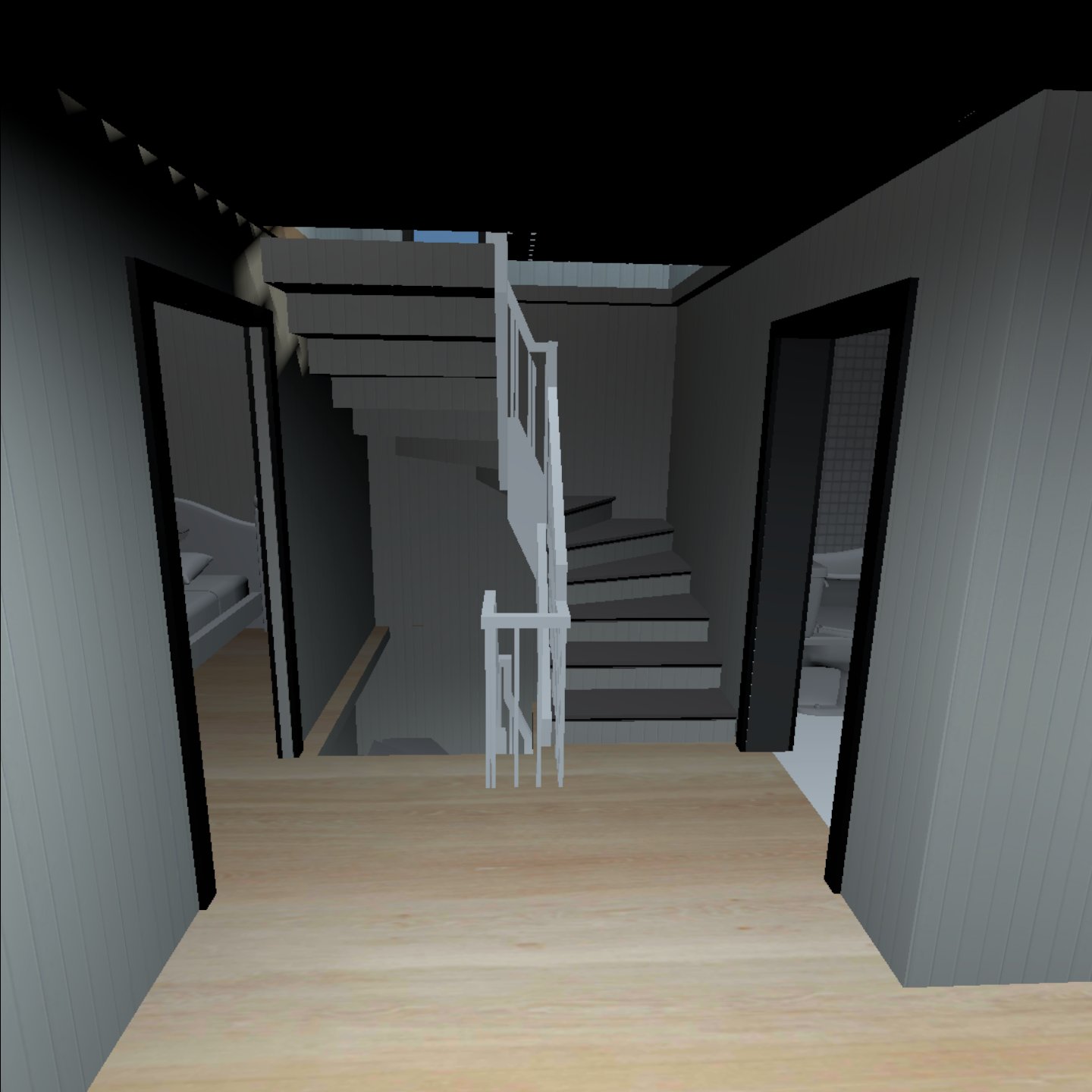}
    \caption{Entrance.}
  \label{fig:sfig5}
\end{subfigure}
\begin{subfigure}{.495\linewidth}
  \centering
  \includegraphics[width=\linewidth]{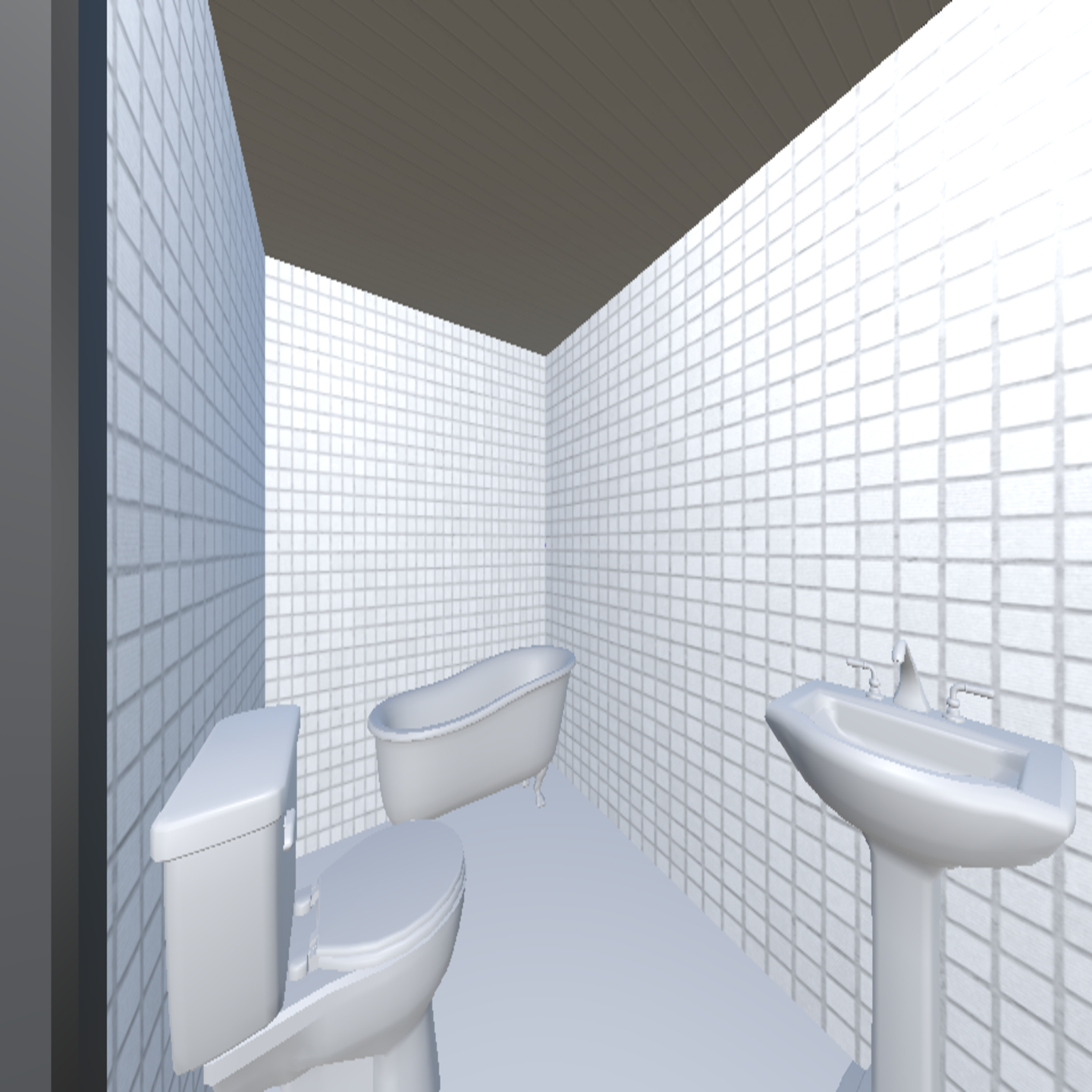}
  \caption{Bathroom.}
  \label{fig:sfig3}
\end{subfigure}\\
\begin{subfigure}{.495\linewidth}
  \centering
  \includegraphics[width=\linewidth]{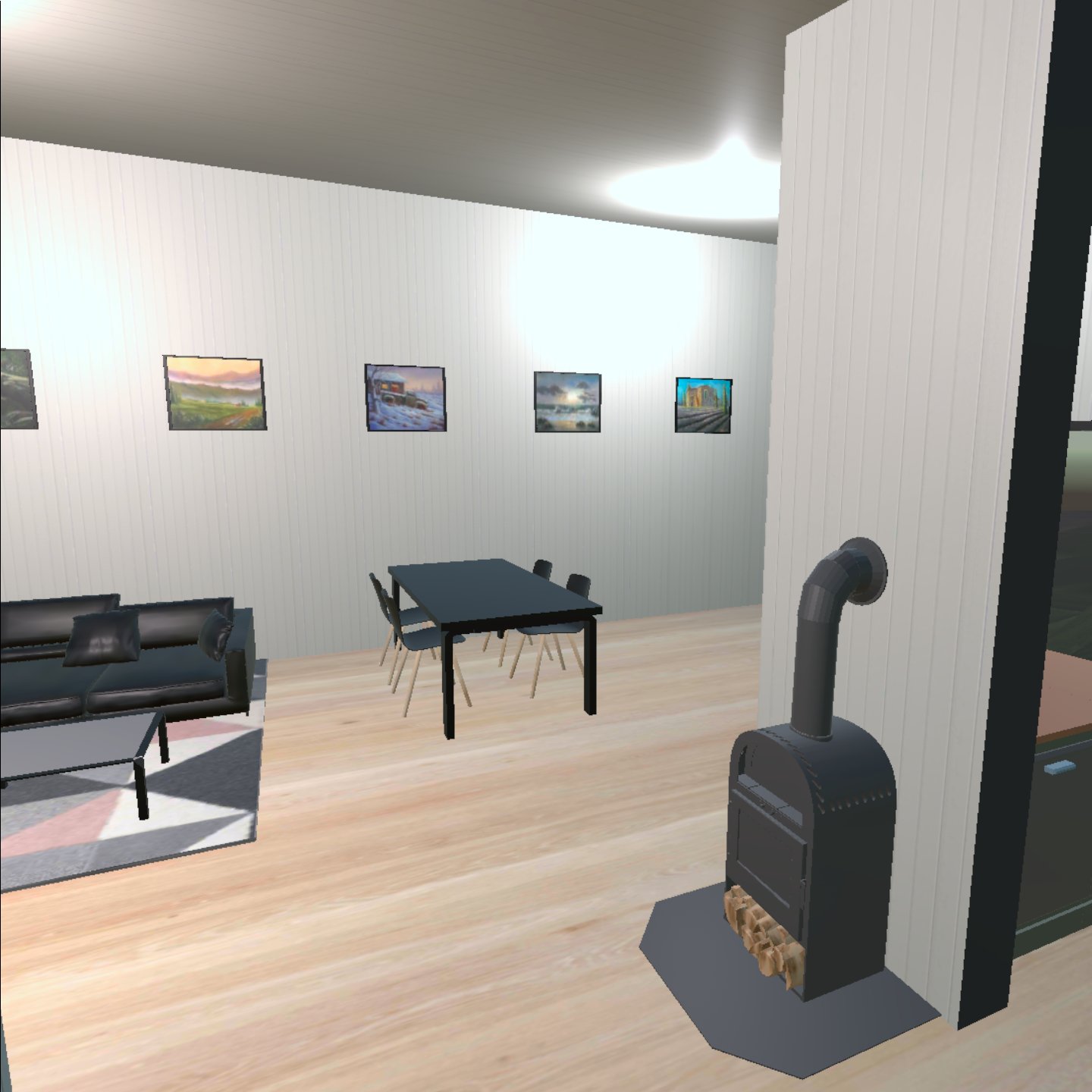}
    \caption{Living room.}
  \label{fig:sfig6}
\end{subfigure}
\begin{subfigure}{.495\linewidth}
  \centering
  \includegraphics[width=\linewidth]{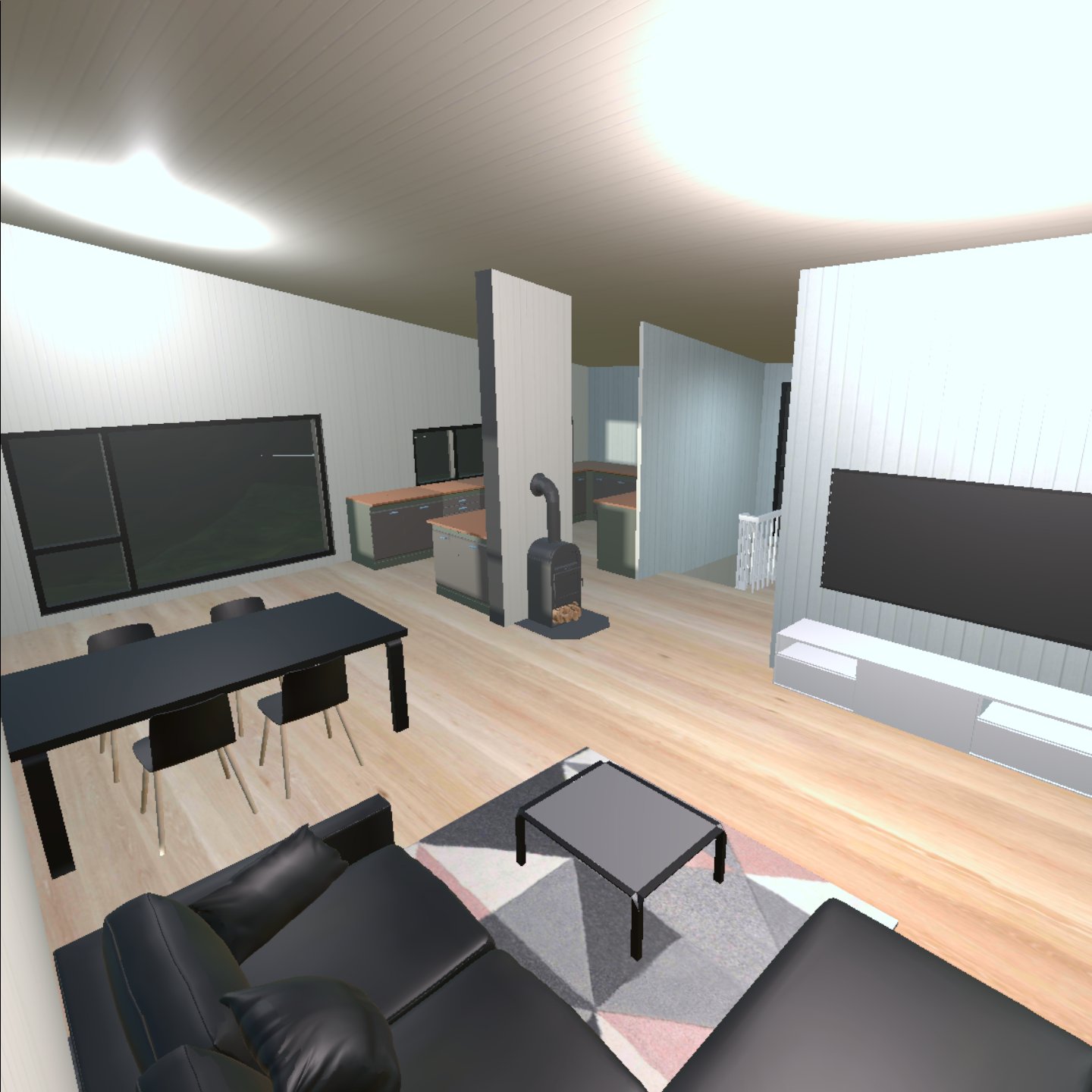}
    \caption{Living room.}
\end{subfigure}
\caption{Demonstration of the internal environment using a standalone DT.}
\label{fig:standaloneDemo_interior}
\end{figure}

It is apparent that Figs.~\ref{fig:standaloneDemo_exterior} and \ref{fig:standaloneDemo_interior} provide more comprehensive information in contrast to Fig.~\ref{fig:sitebeforeconstruction}. Figs.~ \ref{fig:sfig1} and \ref{fig:sfig8} offer a better understanding of the surroundings after construction, facilitating the selection of building materials that complement the natural environment. The available space in the driveway, as shown in Fig.~\ref{fig:tesla2}, can assist in planning which car sizes can be accommodated, or whether the driveway design should be altered to allow for larger cars. In addition, the balcony lighting simulation (illustrated in Fig.~\ref{fig:balcony}) can provide an estimation of the sunlight availability for any day and time of the year. Furthermore, visualization of the interior environment (depicted in Fig.~\ref{fig:standaloneDemo_interior}) can aid in optimal placement of lighting fixtures, selection of wall colors, flooring, furniture, and other relevant objects.

\subsection{Descriptive DT}
\textit{Scenario:} Suppose a standalone DT of the house was available, allowing the buyer to make an informed decision and customize the house to their liking. With digitalization in mind, advanced sensors and controllers were installed, providing real-time information about various aspects of the house, including indoor air quality, water leakage, door status, occupancy, security breaches, and external weather conditions. The homeowner wants to monitor the house remotely while away and desires more insight into the indoor environment while present inside the house.

\textit{Solution:} We now show how a descriptive DT can provide additional information regarding the house using real-time data from the installed sensors.

\begin{figure}[!htb]
   \begin{subfigure}{0.495\linewidth}
     \centering
     \includegraphics[width=\linewidth]{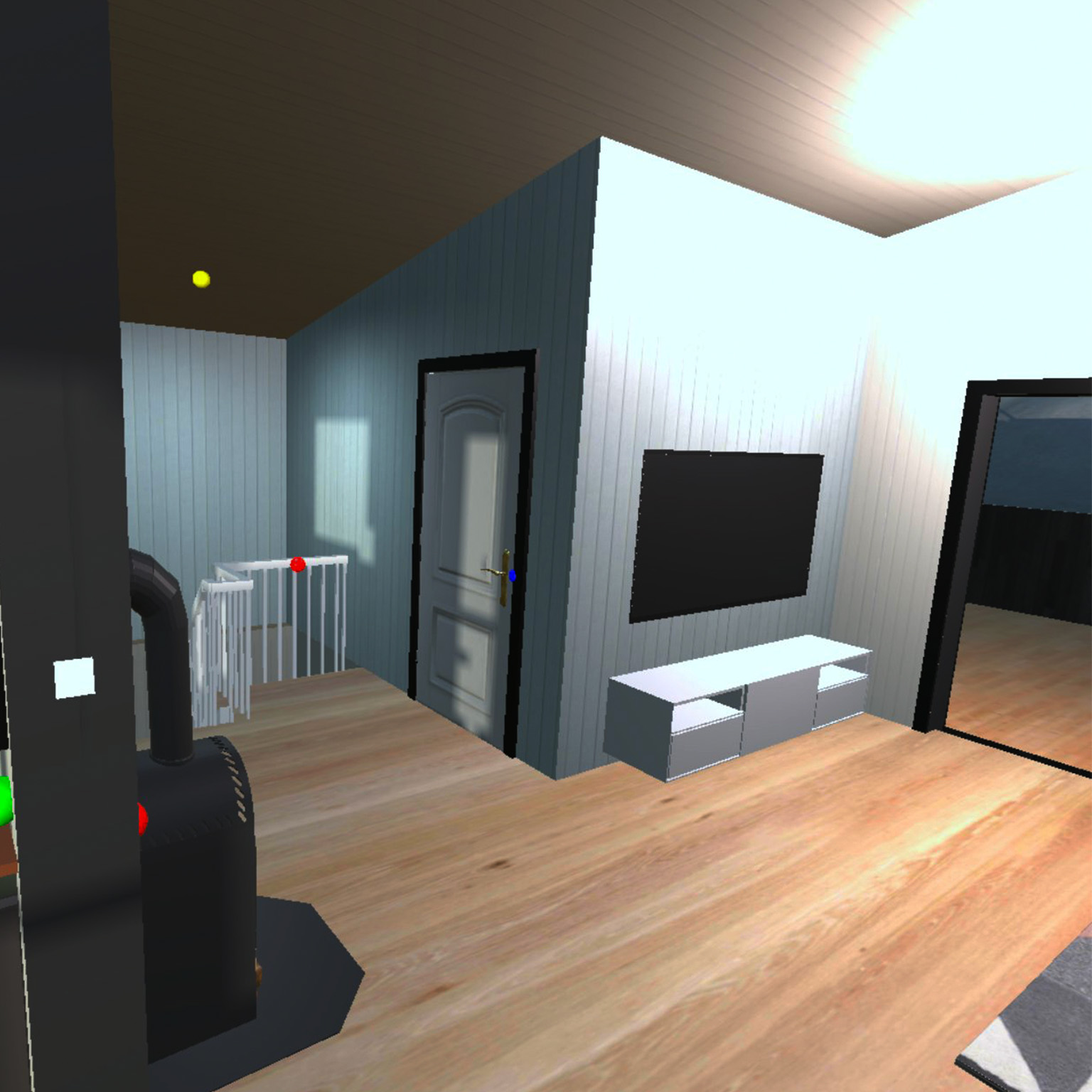}
     \caption{Virtual office door.}
   \end{subfigure}\hfill
   \begin{subfigure}{0.495\linewidth}
     \centering
     \includegraphics[width=\linewidth]{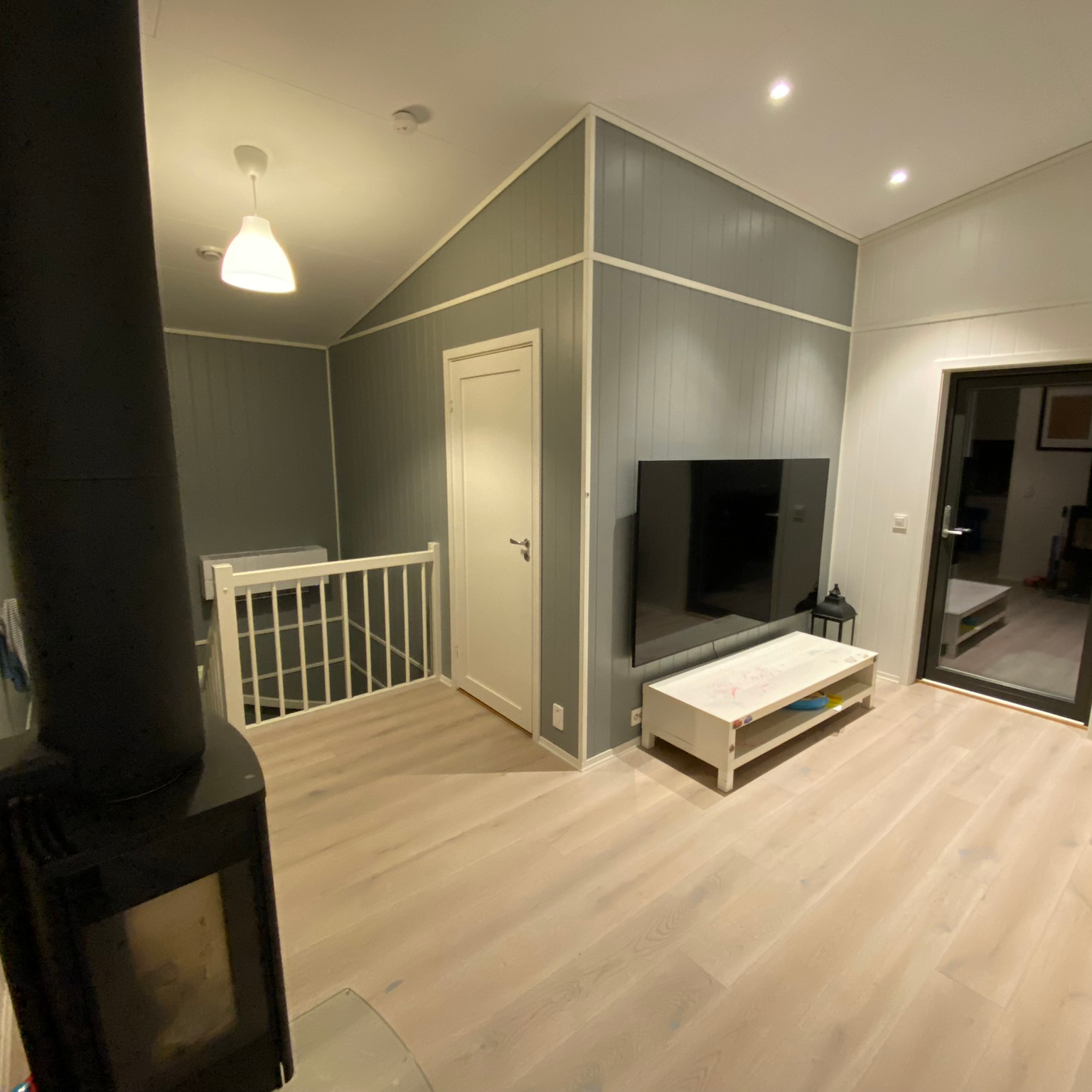}
     \caption{Real office door.}
   \end{subfigure}
    \caption{Real-time states of 2OfficeDoor, 2Stair, HueColorLamp2 located in the second floor, observed 21.02.2022. Note that the color of the spheres follows the same standards set in Fig.~\ref{fig:floorplan-sensors} and their positions represent the 3D position of the sensor in the physical asset.}
    \label{fig:real_time_states_2ndfloor}
\end{figure}

\begin{figure}[!htb]
   \begin{subfigure}{0.475\linewidth}
     \centering
     \includegraphics[width=\linewidth]{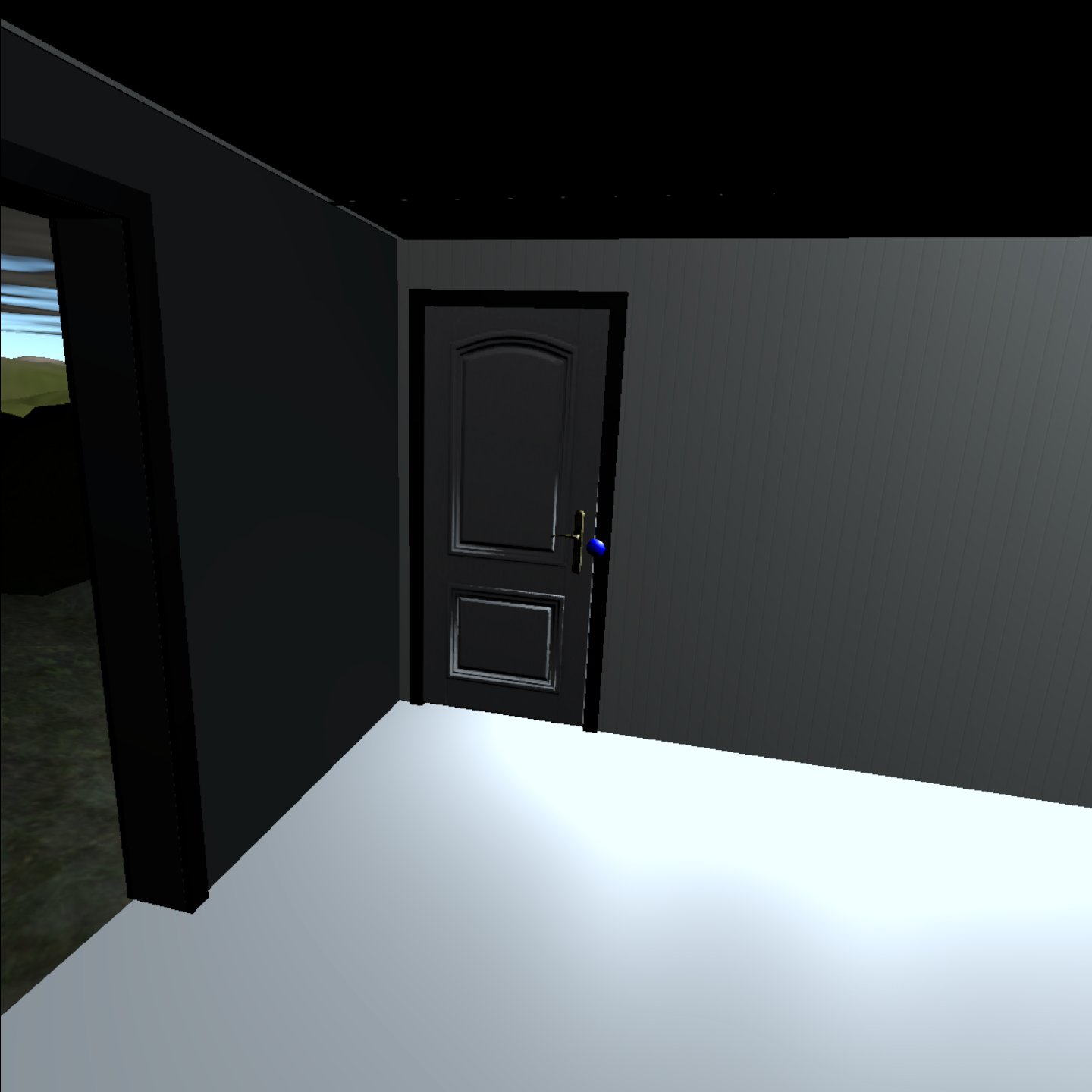}
     \caption{0BGuestDoor.}
   \end{subfigure}
   \begin{subfigure}{0.475\linewidth}
     \centering
     \includegraphics[width=\linewidth]{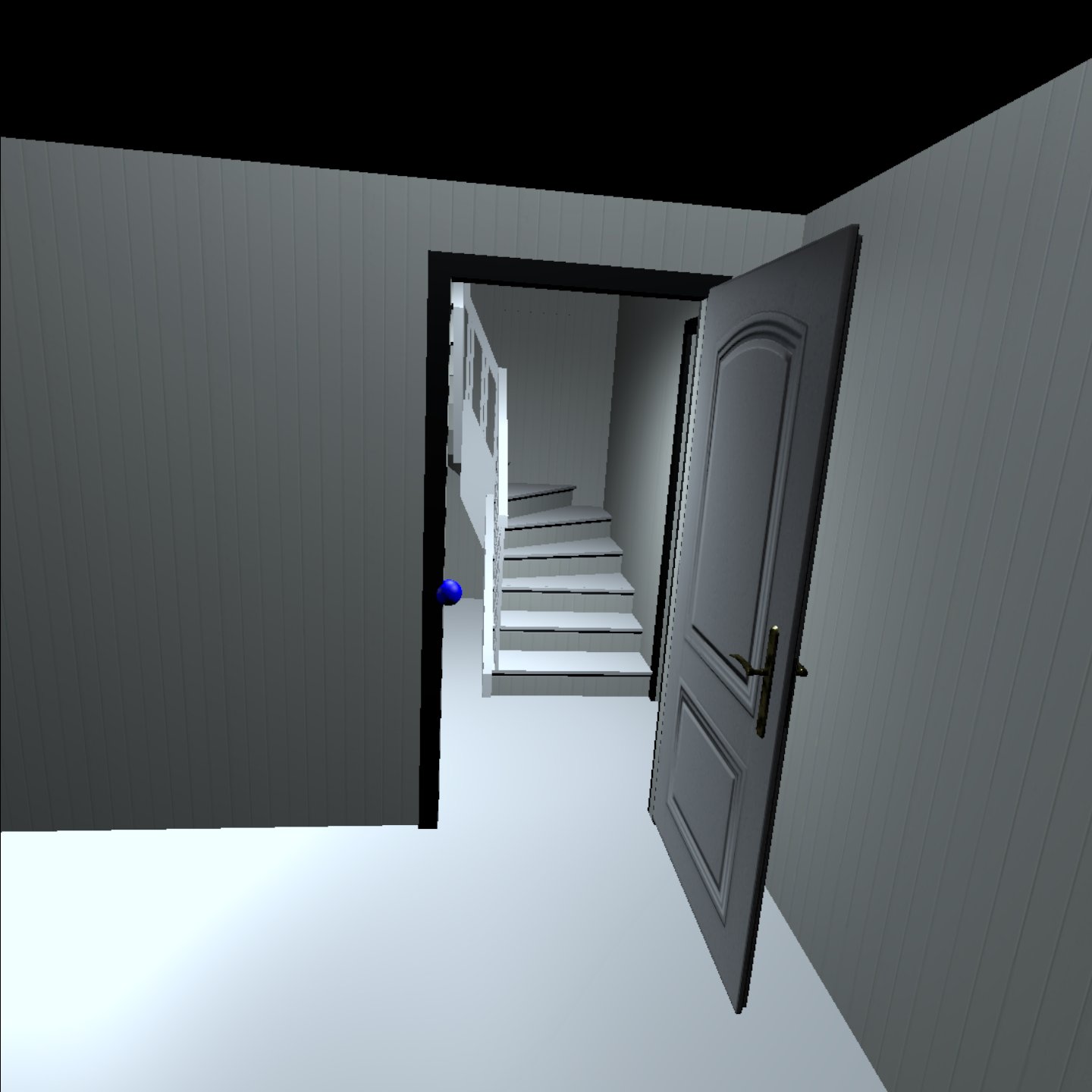}
     \caption{0BDoor.}
   \end{subfigure}\\
   \begin{subfigure}{0.475\linewidth}
     \centering
     \includegraphics[width=\linewidth]{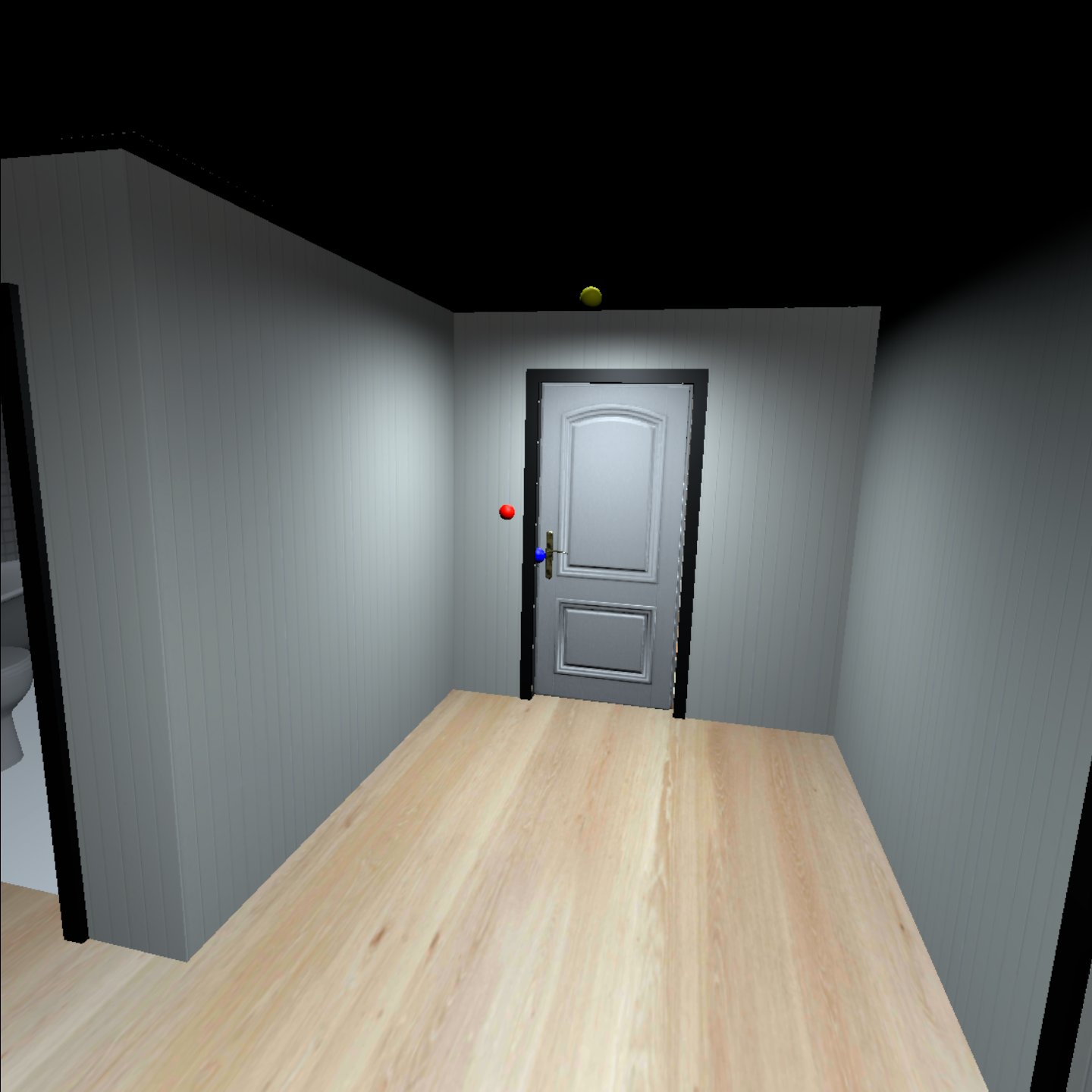}
     \caption{1MainDoor, Entrance1Ceiling.}
   \end{subfigure}
   \begin{subfigure}{0.475\linewidth}
     \centering
     \includegraphics[width=\linewidth]{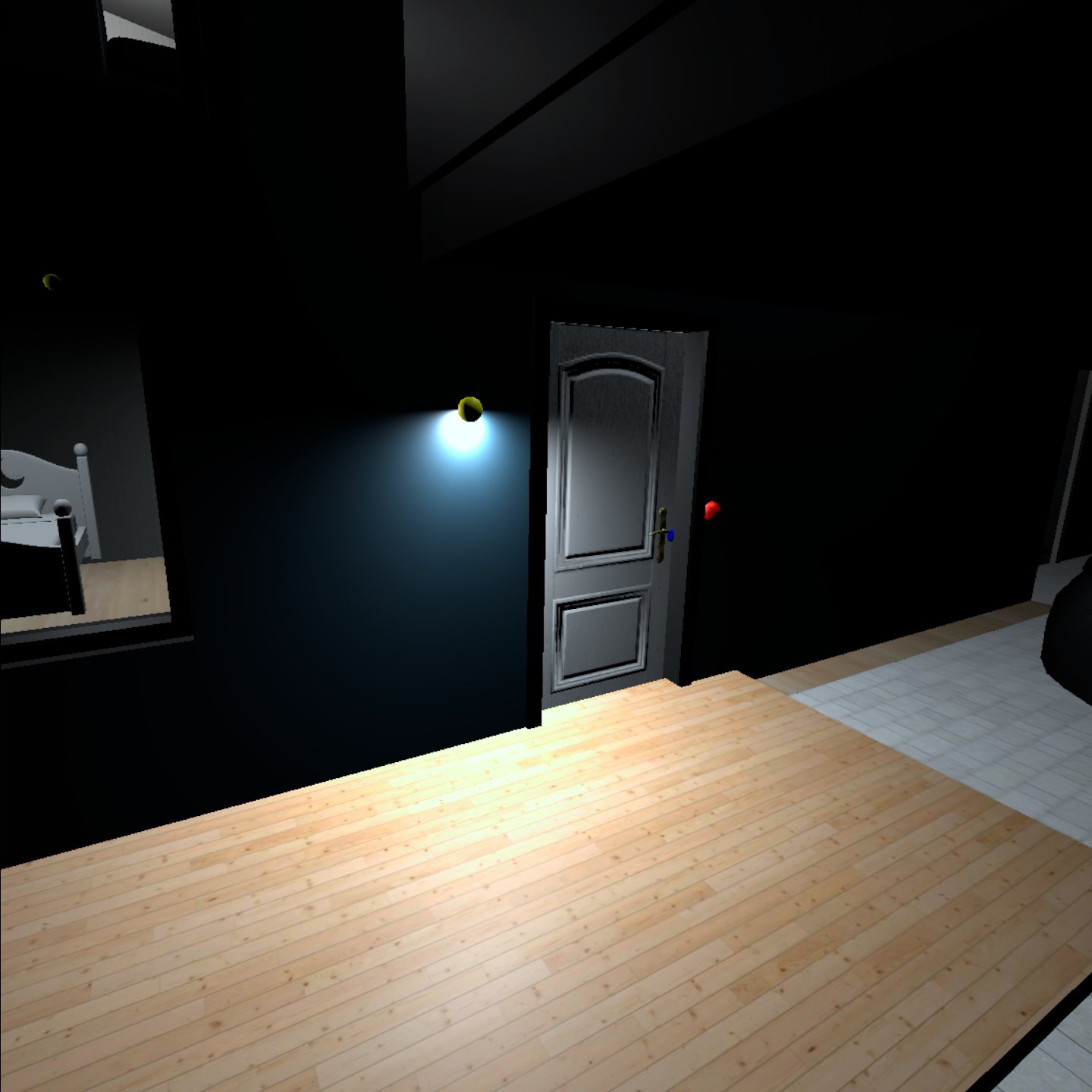}
     \caption{1MainDoor, Outdoor1.}
   \end{subfigure}\\
   \begin{subfigure}{0.475\linewidth}
     \centering
     \includegraphics[width=\linewidth]{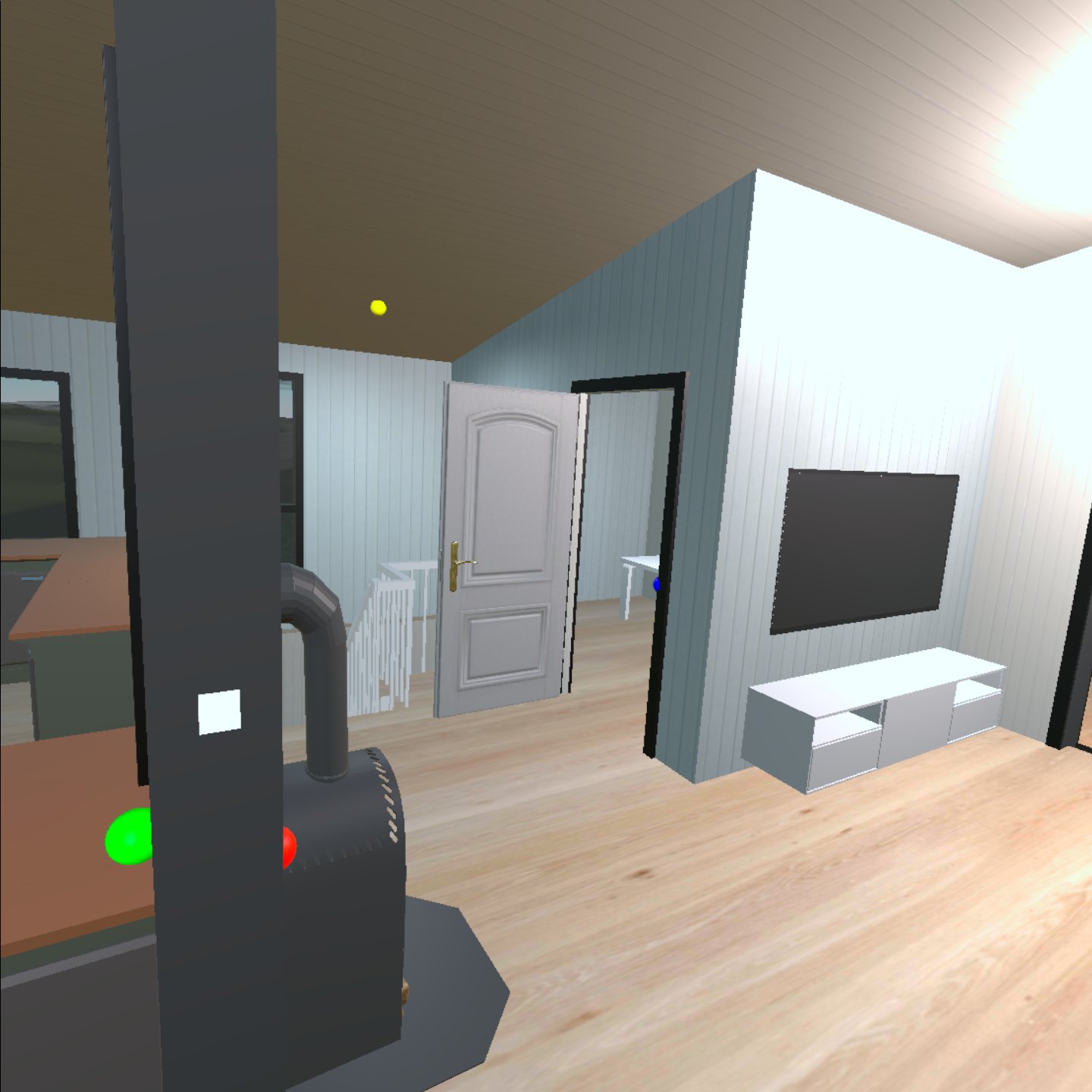}
     \caption{2OfficeDoor, HueColorLamp2.}
   \end{subfigure}
   \begin{subfigure}{0.475\linewidth}
     \centering
     \includegraphics[width=\linewidth]{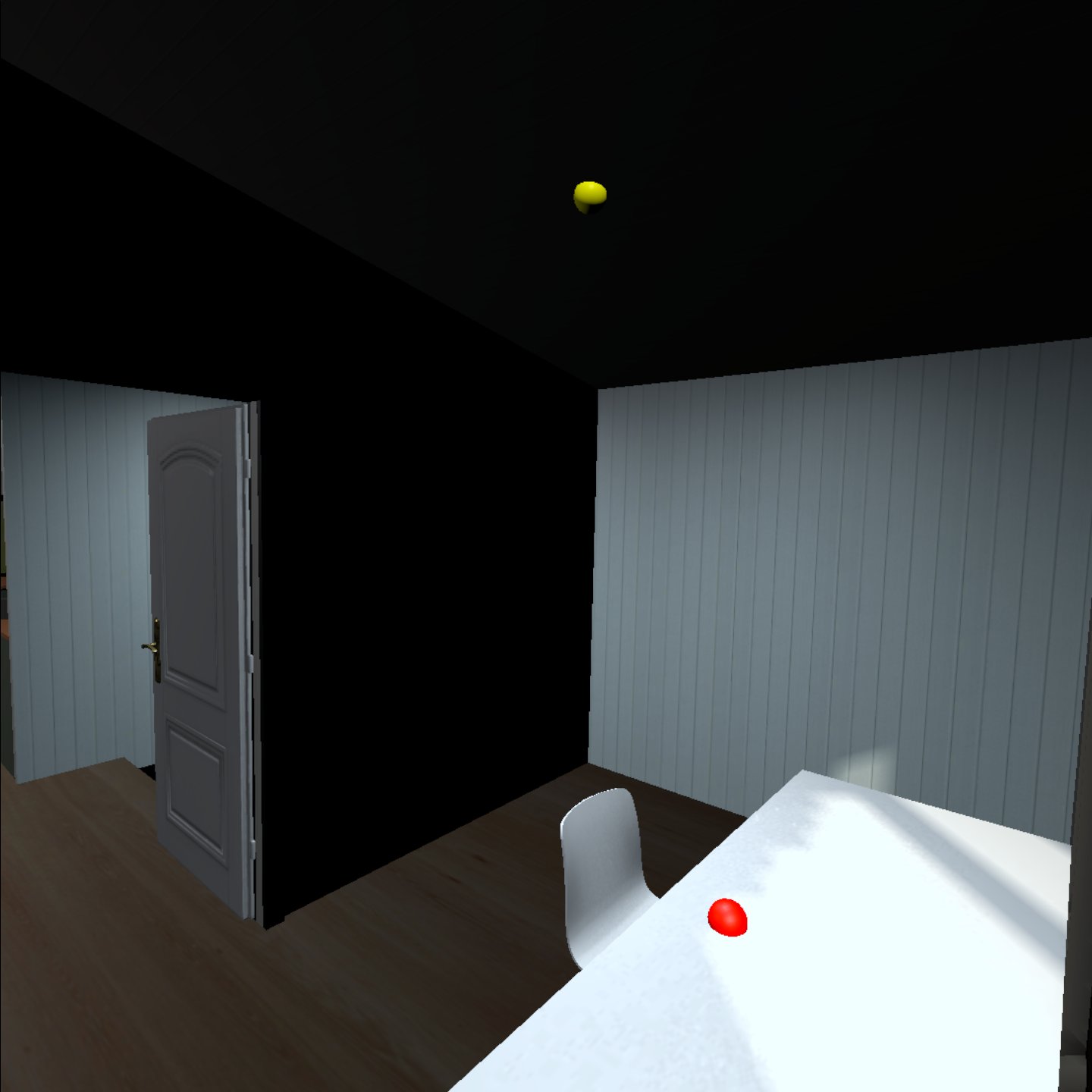}
     \caption{2OfficeDoor, OfficeCeiling.}
   \end{subfigure}
    \caption{Demonstration of various door and light sensor states in the descriptive DT. Observations recorded throughout the day on 21.05.2022.}
    \label{fig:real_time_states_demo}
\end{figure}

\begin{figure}[!htb]
   \begin{subfigure}{0.475\linewidth}
     \centering
     \includegraphics[width=\linewidth]{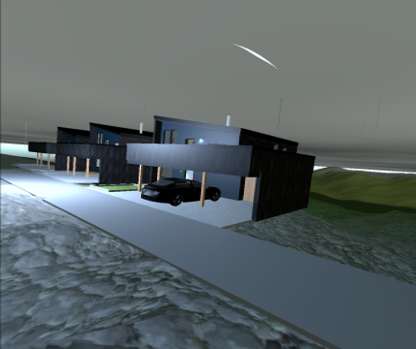}
     \caption{Rain observed 24.03.2022.}
   \end{subfigure}
   \begin{subfigure}{0.475\linewidth}
     \centering
     \includegraphics[width=\linewidth]{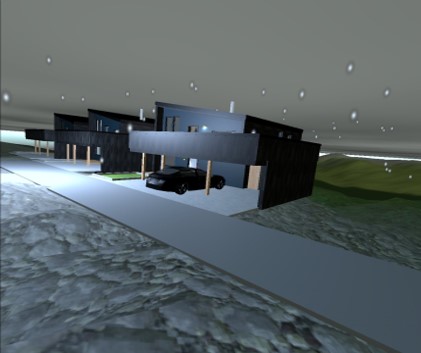}
     \caption{Snow observed 03.04.2022.}
   \end{subfigure}\\
      \begin{subfigure}{0.475\linewidth}
     \centering
     \includegraphics[width=\linewidth]{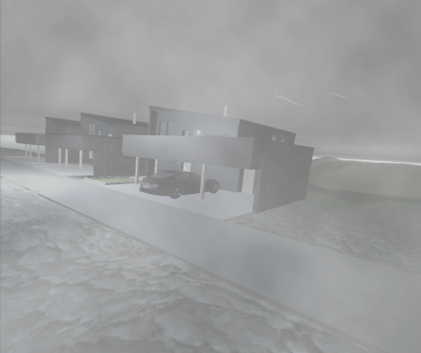}
     \caption{Fog observed 25.04.2022.}
   \end{subfigure}
      \begin{subfigure}{0.475\linewidth}
     \centering
     \includegraphics[width=\linewidth]{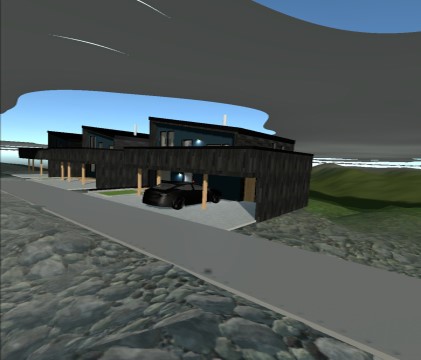}
     \caption{Cloudy observed 03.05.2022.}
   \end{subfigure}\\
      \begin{subfigure}{0.475\linewidth}
     \centering
     \includegraphics[width=\linewidth,height=\linewidth]{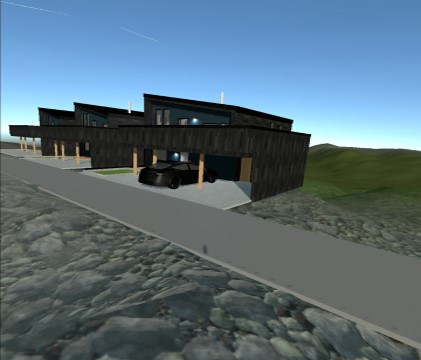}
     \caption{Clear sky observed 21.04.2022.}
   \end{subfigure}
   \begin{subfigure}{0.475\linewidth}
     \centering
     \includegraphics[width=\linewidth]{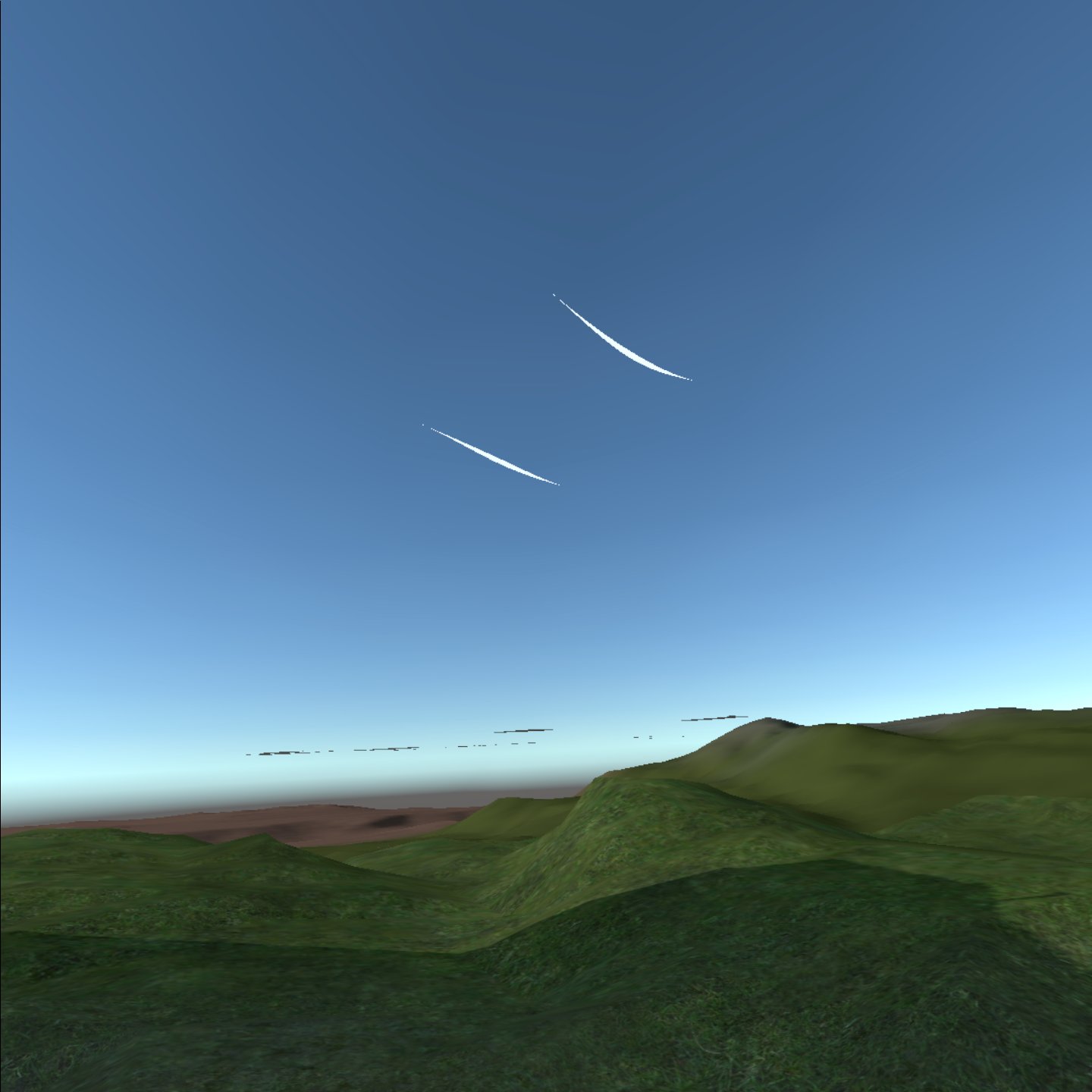}
     \caption{Wind effect.}
     \label{fig:wind_effect}
   \end{subfigure}
     \caption{Demonstration of weather conditions taking place in the descriptive DT. }
    \label{fig:diagnostic_weather_condition}
\end{figure}

\begin{figure}[!htb]
      \begin{subfigure}{0.475\linewidth}
     \centering
     \includegraphics[width=\linewidth]{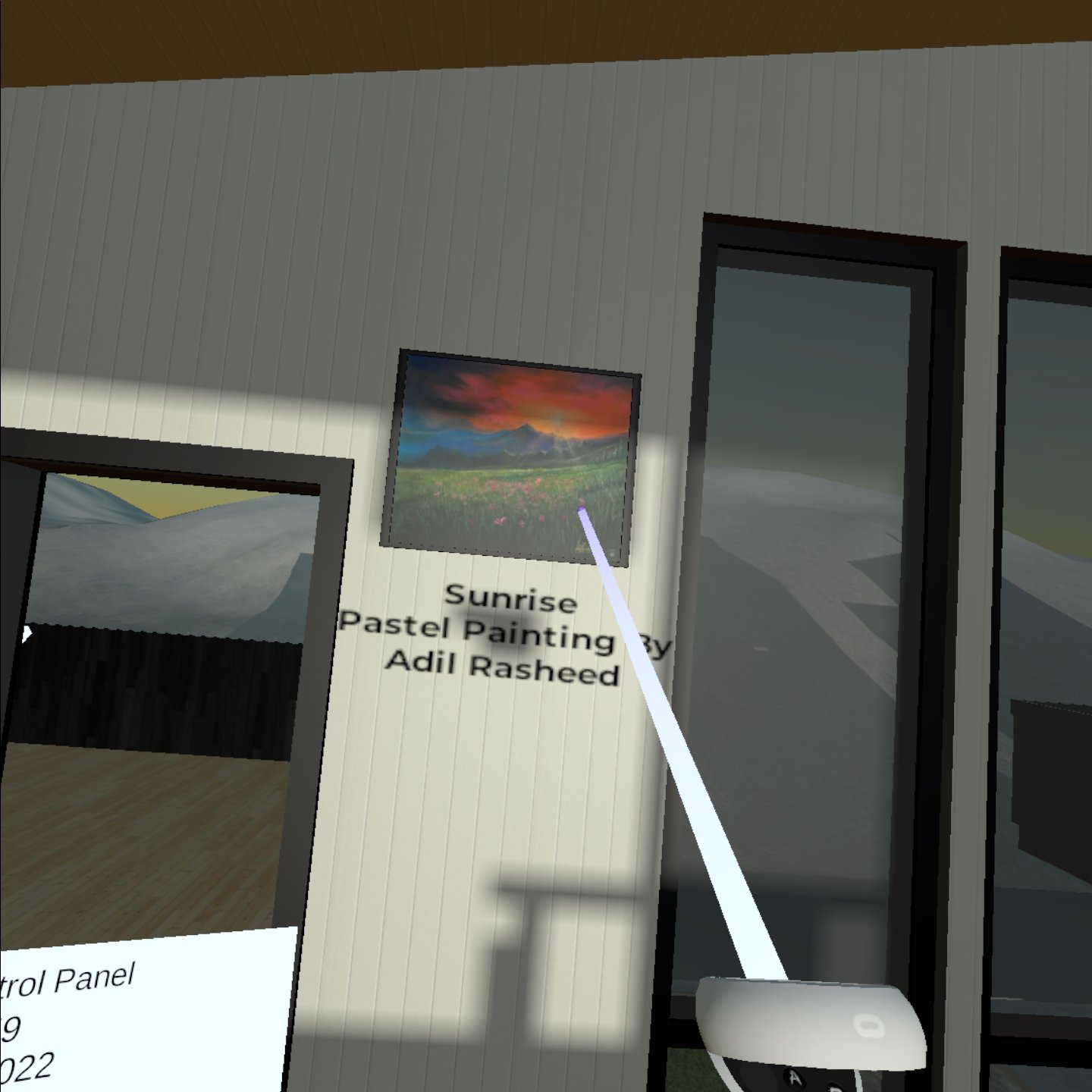}
     \caption{Pastel painting.}
   \end{subfigure}
   \begin{subfigure}{0.475\linewidth}
     \centering
     \includegraphics[width=\linewidth]{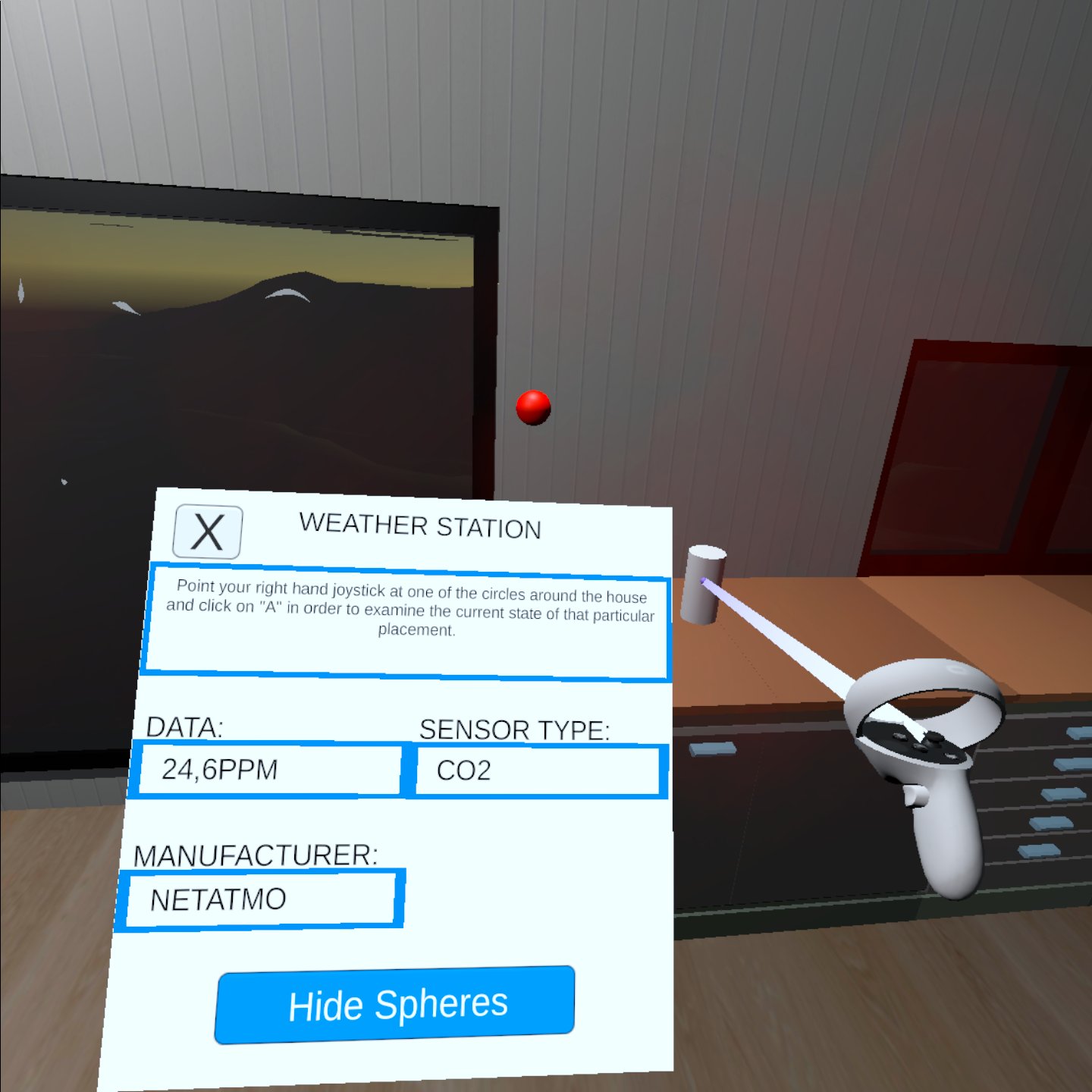}
     \caption{Real-time CO2 data.}
   \end{subfigure}\\
      \begin{subfigure}{0.475\linewidth}
     \centering
     \includegraphics[width=\linewidth]{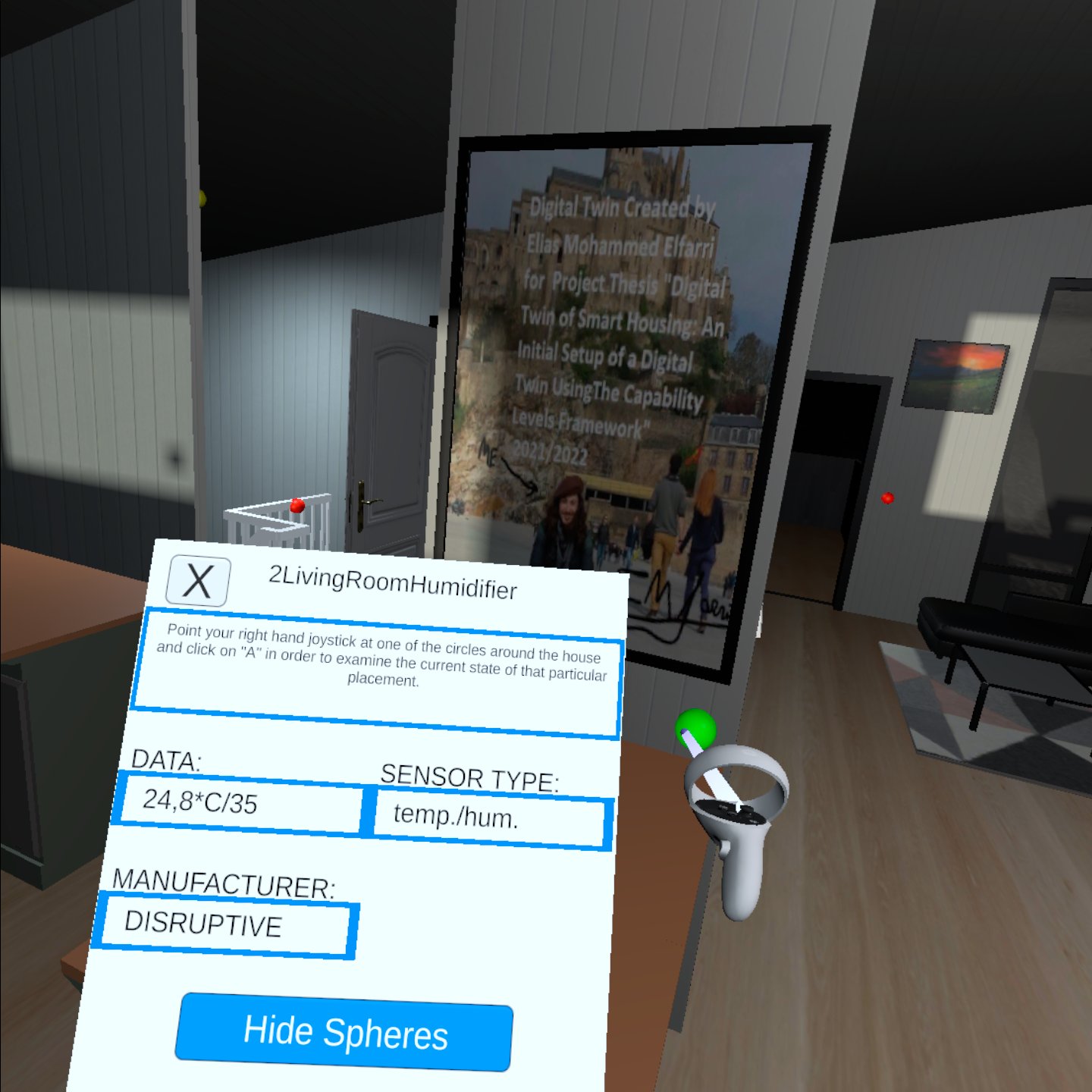}
     \caption{Real-time temperature/humidity data.}
   \end{subfigure}
   \begin{subfigure}{0.475\linewidth}
     \centering
     \includegraphics[width=\linewidth]{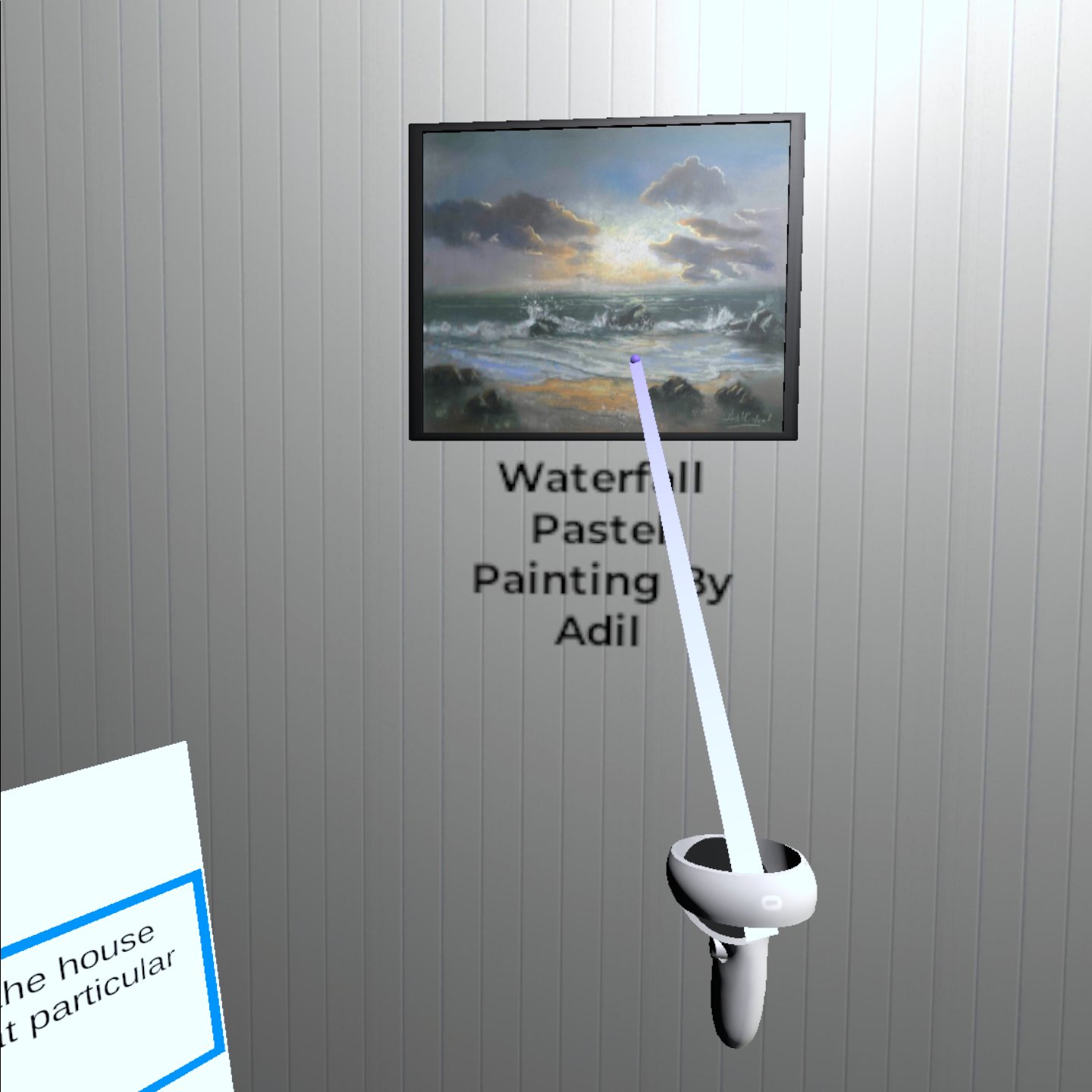}
     \caption{Another pastel painting.}
   \end{subfigure}
    \caption{Information from the remaining sensors and other objects, obtained using a virtual reality interface.}
    \label{fig:diagnostic_painting}
\end{figure}

In Figs.~\ref{fig:real_time_states_2ndfloor} and \ref{fig:real_time_states_demo}, the real-time states of the house are displayed, based on the data collected by various sensors and lights. The presented data is raw and unprocessed, intended for human interpretation to draw relevant conclusions. Fig.~\ref{fig:diagnostic_weather_condition} provides an example of rendering the prevailing weather conditions on specific days. Additionally, Fig.~\ref{fig:diagnostic_painting} shows real-time data from sensors measuring CO2, humidity, temperature, and paintings, which are otherwise not visually representable in the descriptive DT. The homeowner regularly updates the data from the sensors, and the DT database stores this information. This enables the DT to reflect the up-to-date state of the objects in the house, such as the name of the selected painting. Although monitoring the name of a painting may seem trivial, the same concept can be applied to other objects in the house, such as monitoring resource inventory levels or the amount of wood remaining in the fireplace. Advanced object detection and classification algorithms are necessary to monitor such details.

\subsection{Diagnostic DT}

\textit{Scenario:} Imagine two scenarios, one where the homeowner is physically present inside the house and the other where the homeowner is remotely located and does not have physical access to the house. With the multitude of sensors installed in the house, the diagnostic DT provides not only real-time updates on the current state of the house, as with the descriptive DT, but also critical alerts in case of any changes. When the homeowner is physically present, the diagnostic DT can be utilized to gain insight into the internal environment using virtual reality to visualize air quality, noise, or temperature maps, which are not visible to the naked eye. On the other hand, when the homeowner is remotely located, the diagnostic DT can provide analytics on the current situation in the house, such as temperature increases, room occupancy, and other relevant information by fusing data from various sensor sources.

\subsubsection{Remote Location}
\begin{figure}[!htb]
    \centering
    \includegraphics[width=\linewidth]{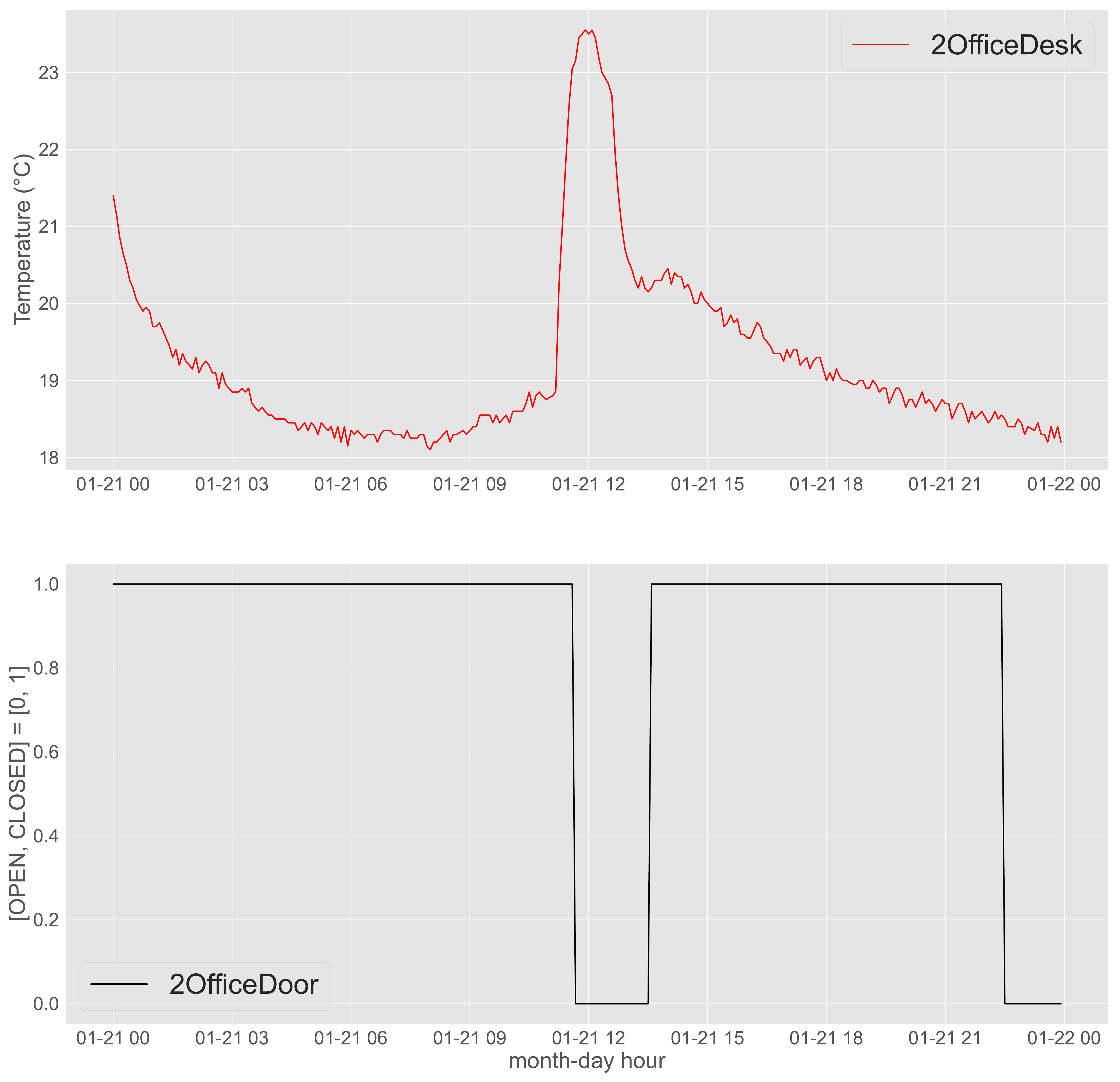}
    \caption{Diagnostic information from fusing office temperature and office door proximity sensor. Scenario one recorded 21.01.2022. }
    \label{fig:diagnostic_remote_location_1}
\end{figure}

When the homeowner is remotely located, they can diagnose whether a room is occupied or not by analyzing temperature and door sensors related to a specific room. Fig.~\ref{fig:diagnostic_remote_location_1} shows a clear temperature rise in the office when the door was opened. Since the sensor was placed under the table, it can sense the body heat of the occupant, and by using simple diagnostic tools, these peaks corresponding to occupancy can be automatically detected and communicated. Thus, one can diagnose that the room was occupied between 11:00-14:00.

\subsubsection{Inside The House}
\begin{figure}[!htb]
   \begin{subfigure}{0.475\linewidth}
     \centering
     \includegraphics[width=\linewidth]{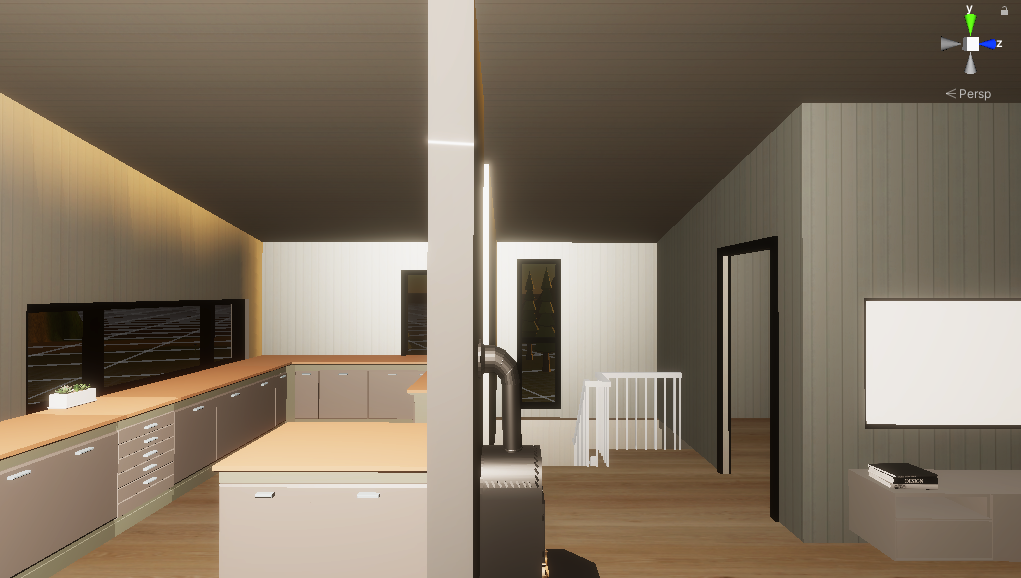}
     \caption{[0, 400) ppm.}
   \end{subfigure}
   \begin{subfigure}{0.475\linewidth}
     \centering
     \includegraphics[width=\linewidth]{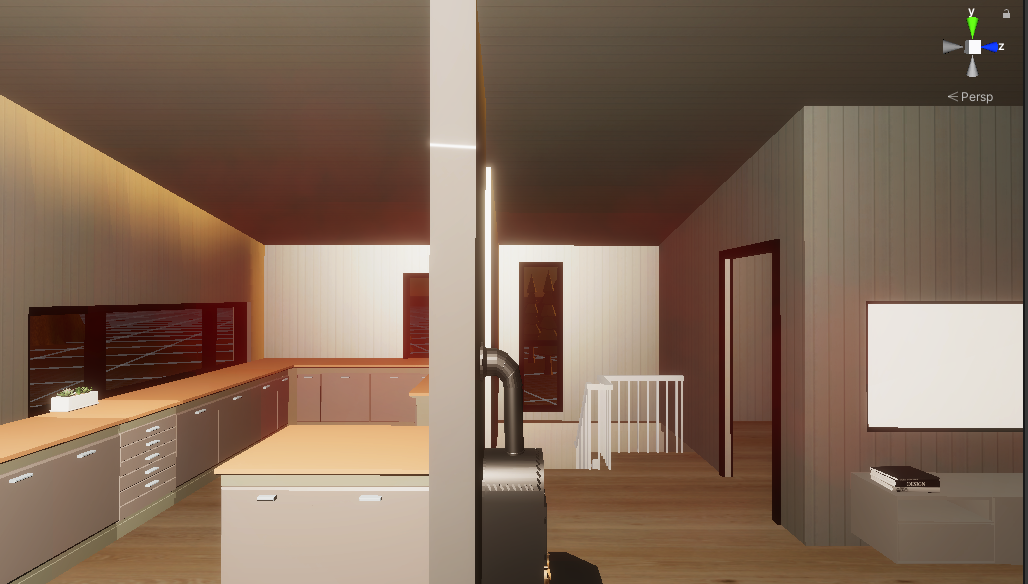}
     \caption{[400, 600) ppm.}
   \end{subfigure}\\
      \begin{subfigure}{0.475\linewidth}
     \centering
     \includegraphics[width=\linewidth]{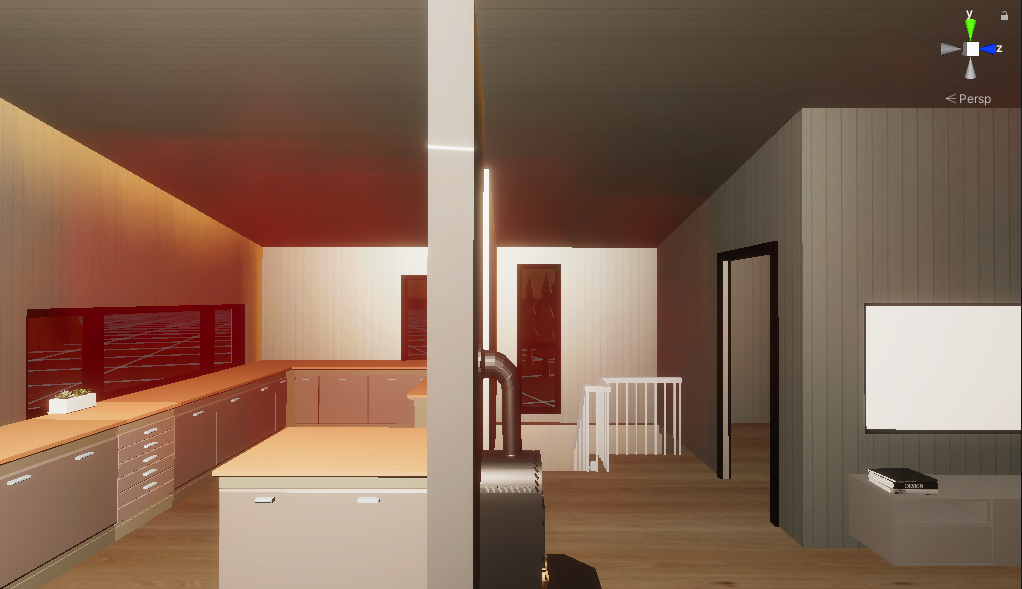}
     \caption{[600, 800) ppm.}
   \end{subfigure}
   \begin{subfigure}{0.475\linewidth}
     \centering
     \includegraphics[width=\linewidth]{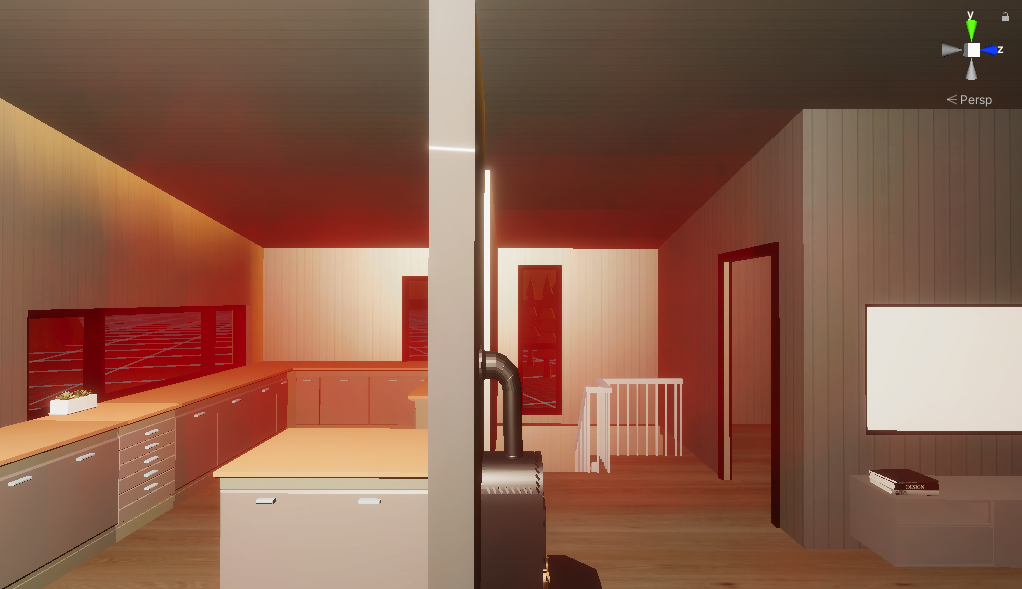}
     \caption{[800, $\infty$) ppm.}
   \end{subfigure}
    \caption{Visualizing CO2 concentration from Netatmo Weather Station as fog in the Unity Game Engine based on 4 predefined intervals. Recorded 27.11.2021.}
    \label{fig:diagnostic_DT_co2_results}
\end{figure}

\begin{figure}[!htb]
    \centering
    \includegraphics[width=\linewidth]{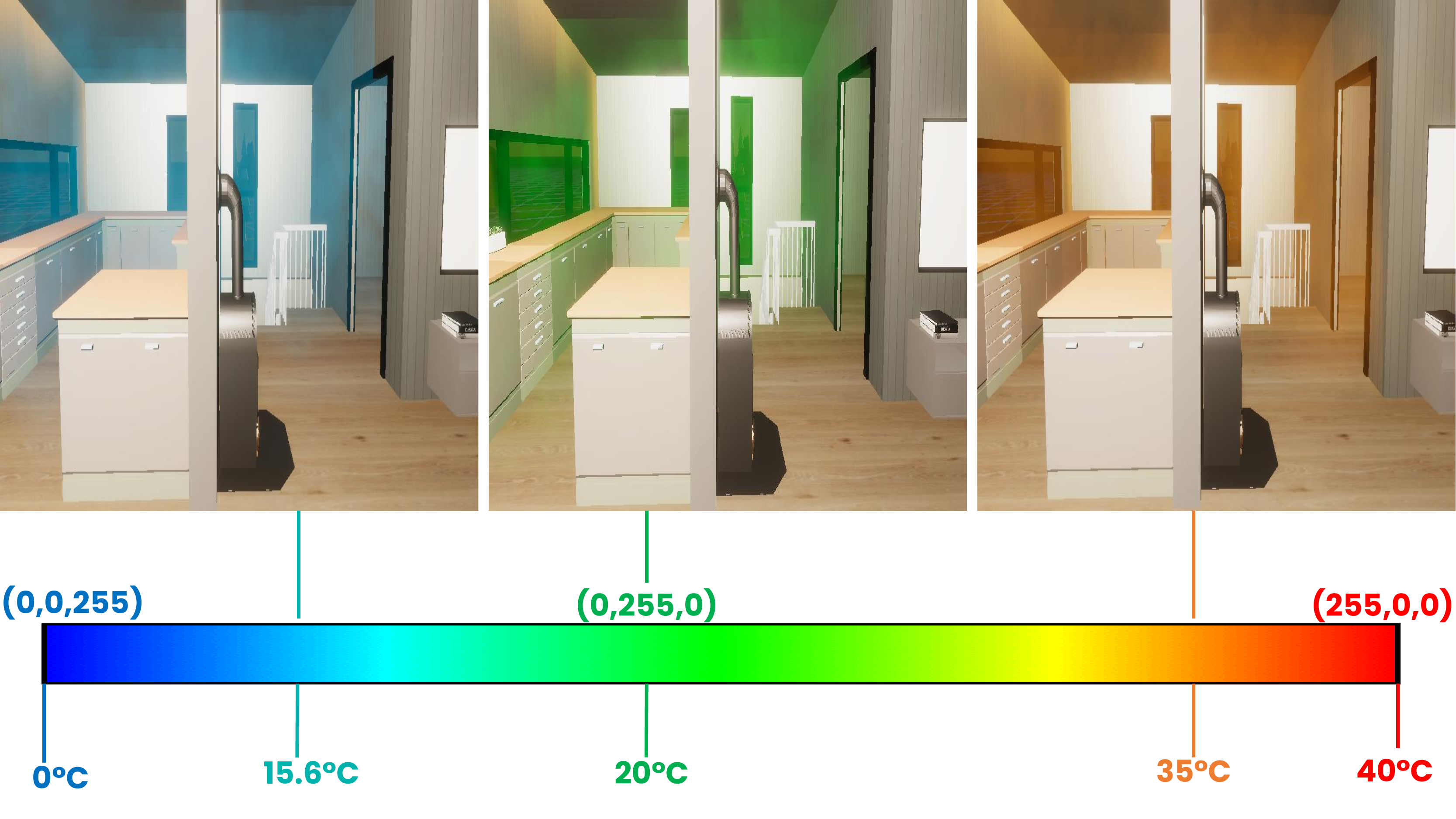} \caption{Visualizing indoor temperature from Netatmo Weather station as fog by converting temperature in Celsius to an RGB color representation.   $15.6^\circ$C  was observed 30.11.2021 at 07:30 AM, $20^\circ$C at the same date 10:30 AM and $35^\circ$C was not an observation but a simulated scenario to display how that would look like in the event of such an occurrence.}
    \label{fig:diagnostic_temperature_fog}
\end{figure}

When the homeowner is present inside the house, they can visualize diagnostic information such as temperature and CO2 density, as shown in Figs.~\ref{fig:diagnostic_temperature_fog} and \ref{fig:diagnostic_DT_co2_results}, respectively. This information can be used to gain insight into those aspects of the indoor environment which are otherwise invisible to the naked eye, such as CO2 and CO concentration. Note that at the descriptive level, these data could only be presented as numbers, while the diagnostic DT allows for visualization and interpretation of the data to gain a deeper understanding of the indoor environment.
 
\subsection{Predictive DT}
\textit{Scenario:} The homeowner now has access to a significant amount of high-quality sensor data from both past and present, but is interested in knowing the future state of the house to plan for more efficient and cost-effective utilization of energy. They may also want to know about the availability of natural sunlight due to potential future developments, such as the construction of a high-rise building.

\subsubsection{Temperature prediction and forecasting}\label{predictive:DDM_prediction_model}
\begin{figure*} 
   \vspace*{-0.5cm}
   \makebox[\linewidth]{%
   \begin{subfigure}[b]{0.5\linewidth}
     \centering
     \includegraphics[width=\linewidth]{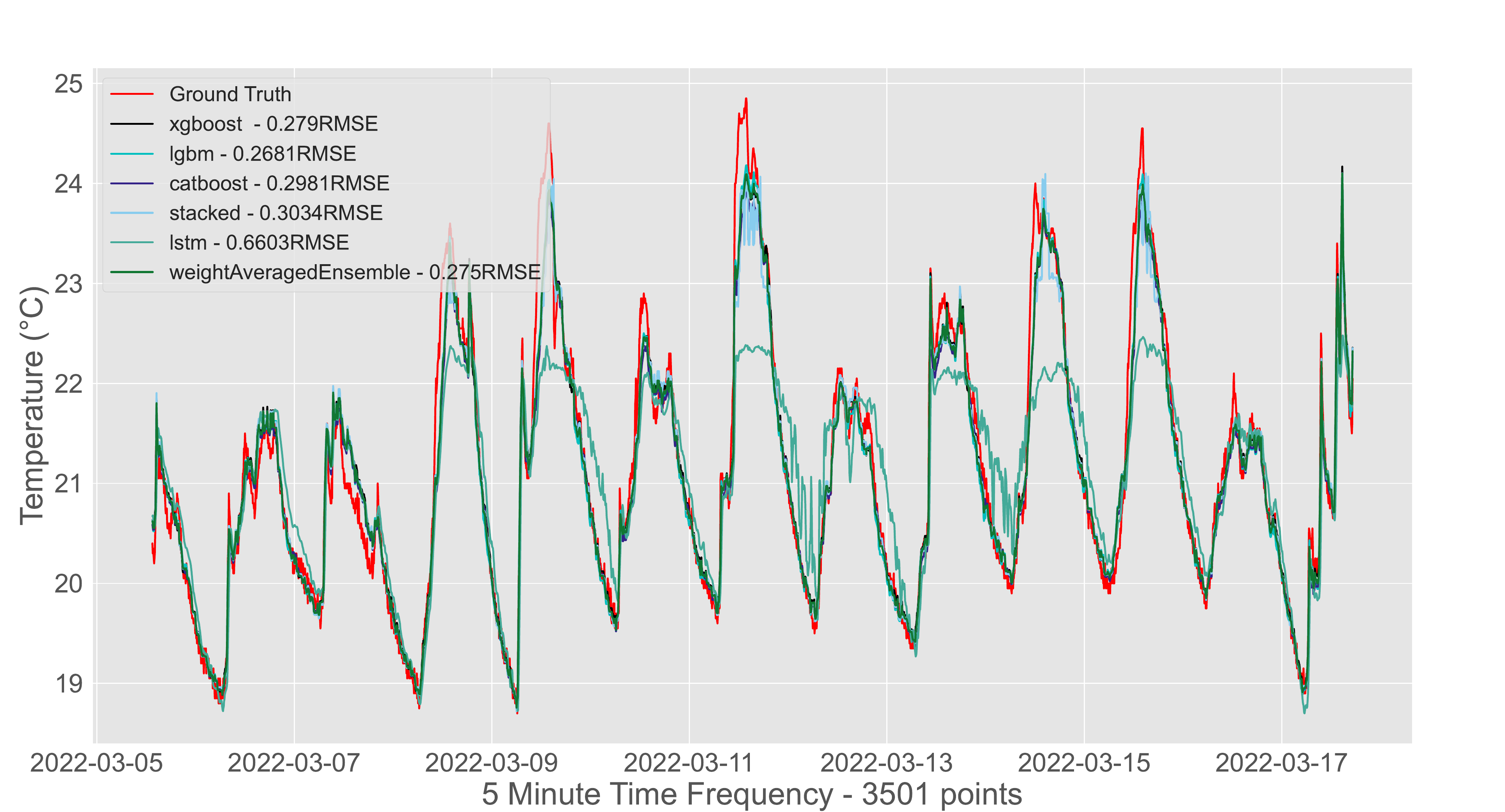}
     \caption{\scriptsize{2BalconyEntrance}}
   \end{subfigure}
   \begin{subfigure}[b]{0.5\linewidth}
     \centering
     \includegraphics[width=\linewidth]{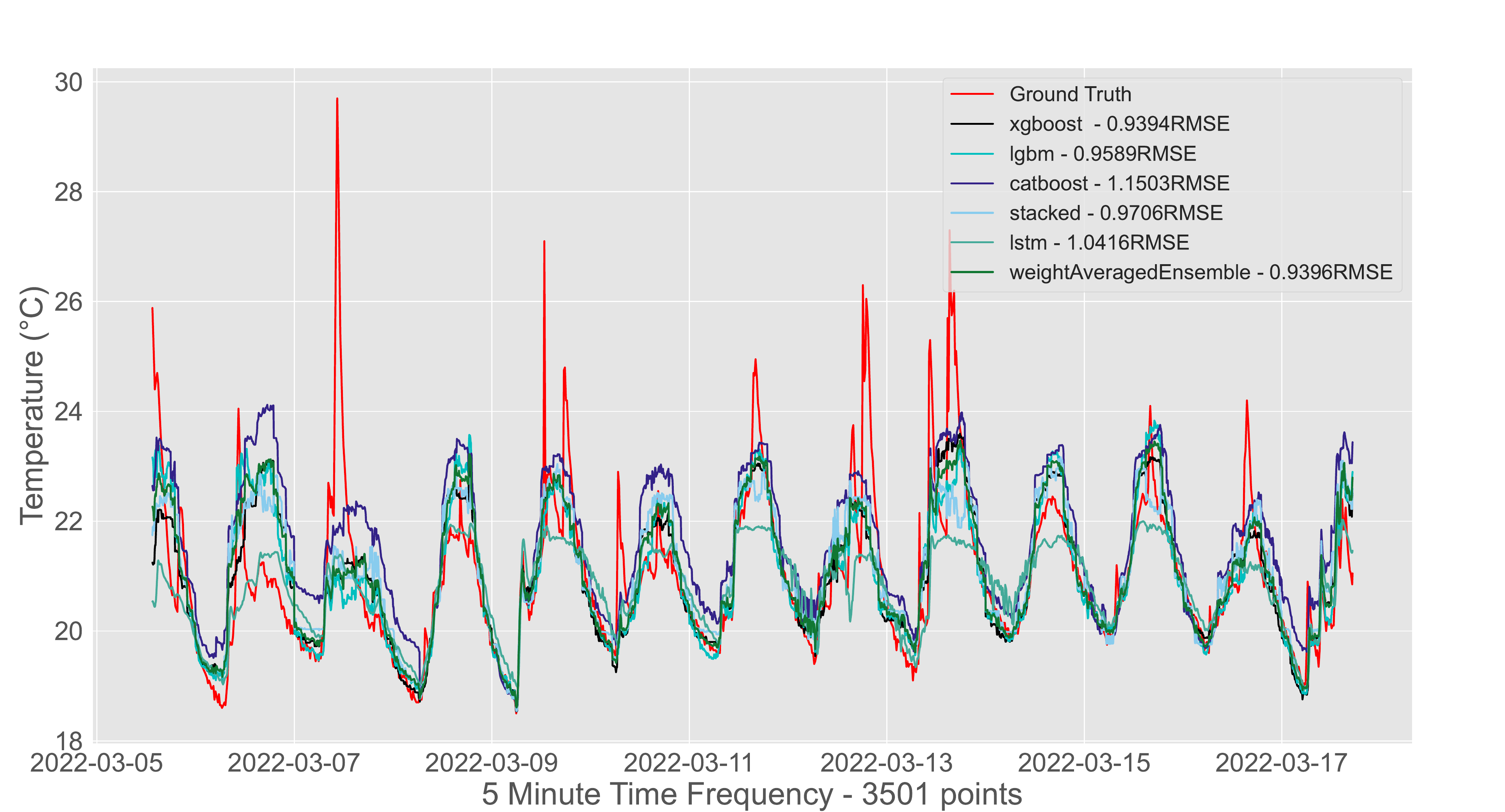}
     \caption{\scriptsize{2Cooking}}
   \end{subfigure}
   }
   
   \makebox[\linewidth]{%
      \begin{subfigure}[b]{0.5\linewidth}
     \centering
     \includegraphics[width=\linewidth]{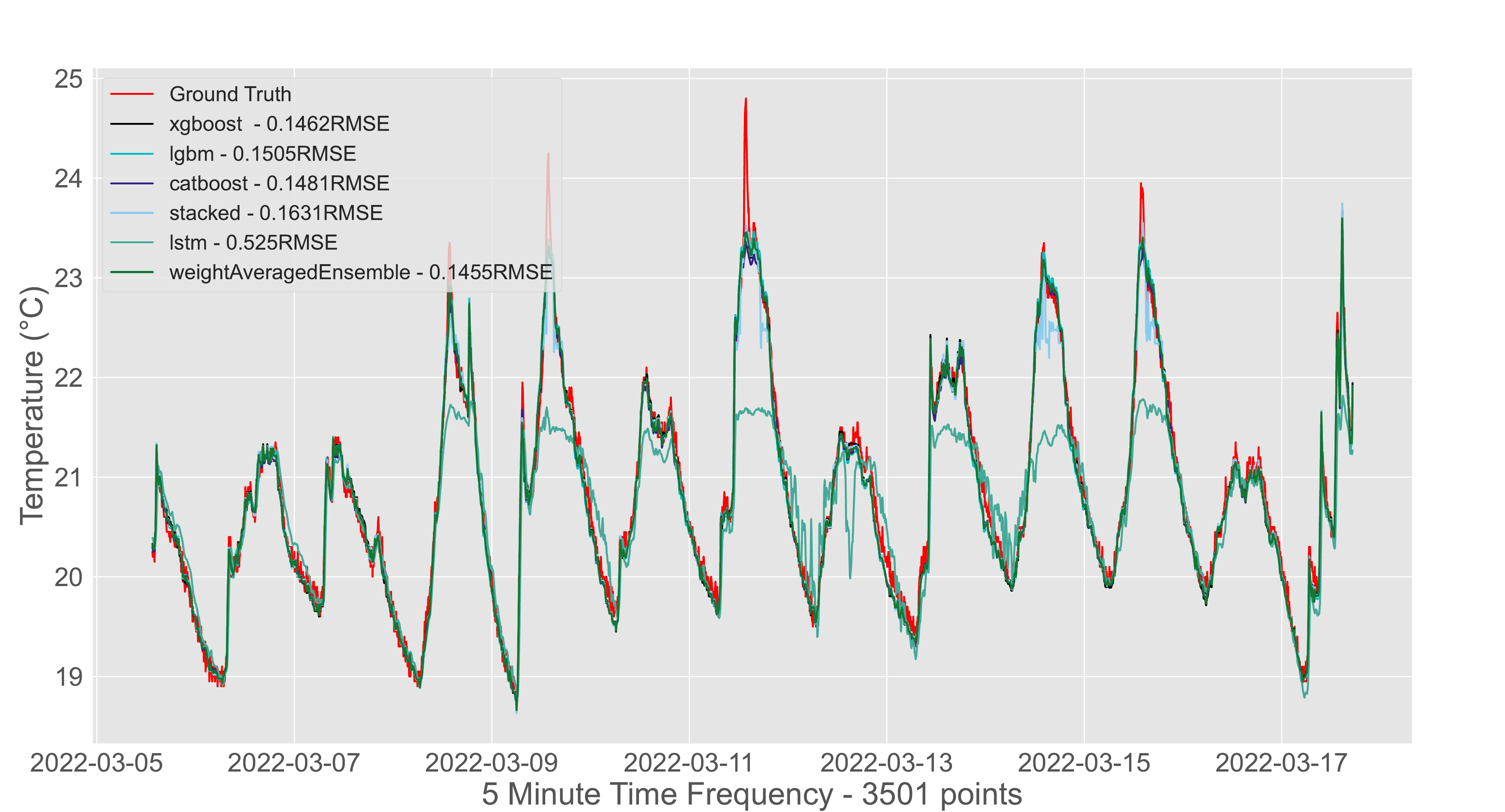}
     \caption{\scriptsize{2LivingRoomCenter }}
   \end{subfigure}
   \begin{subfigure}[b]{0.5\linewidth}
     \centering
     \includegraphics[width=\linewidth]{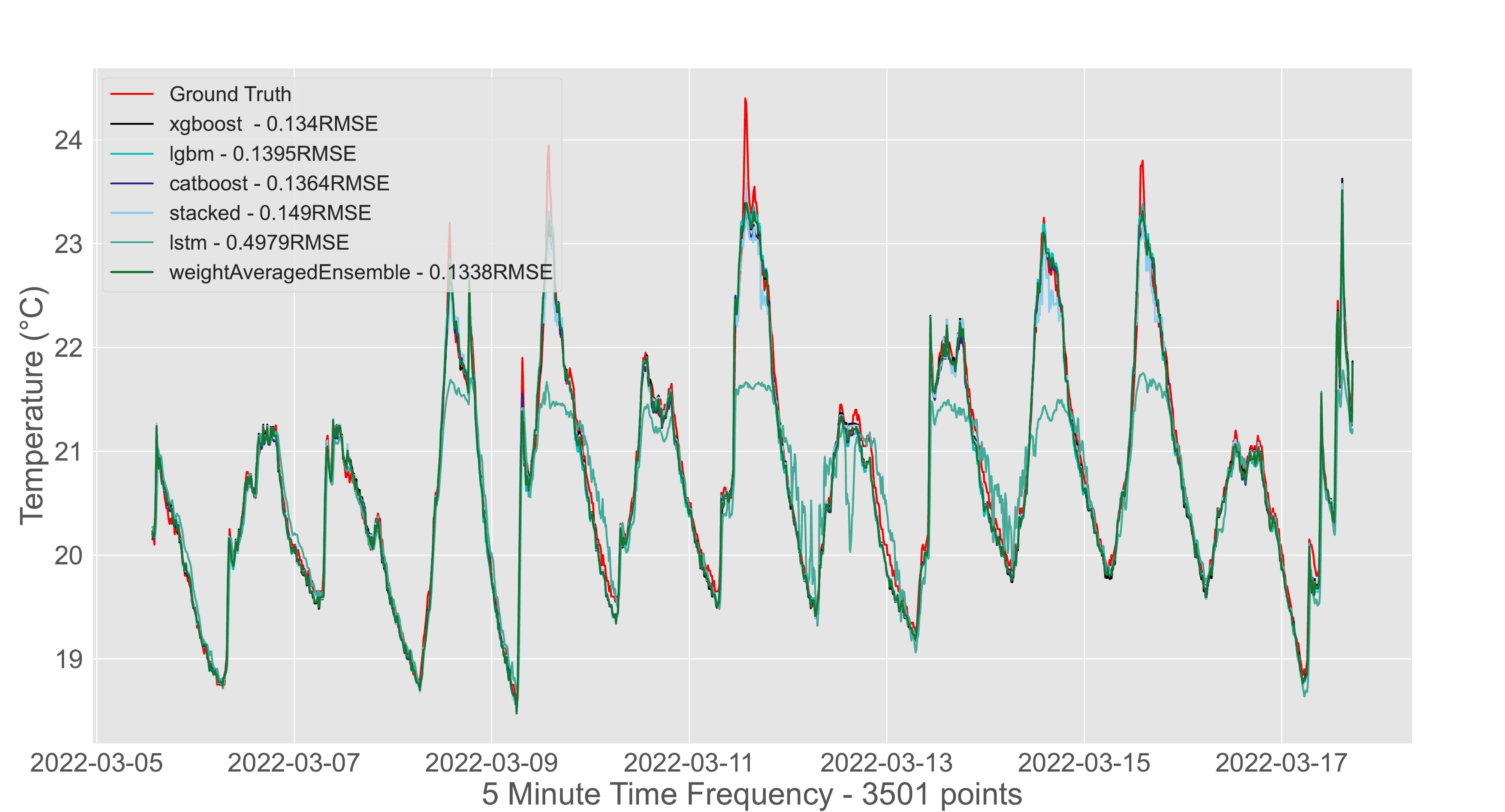}
     \caption{\scriptsize{2LivingRoomCenterHumidity }}
   \end{subfigure}
   }
   
 \makebox[\linewidth]{%
   \begin{subfigure}[b]{0.5\linewidth}
     \centering
     \includegraphics[width=\linewidth]{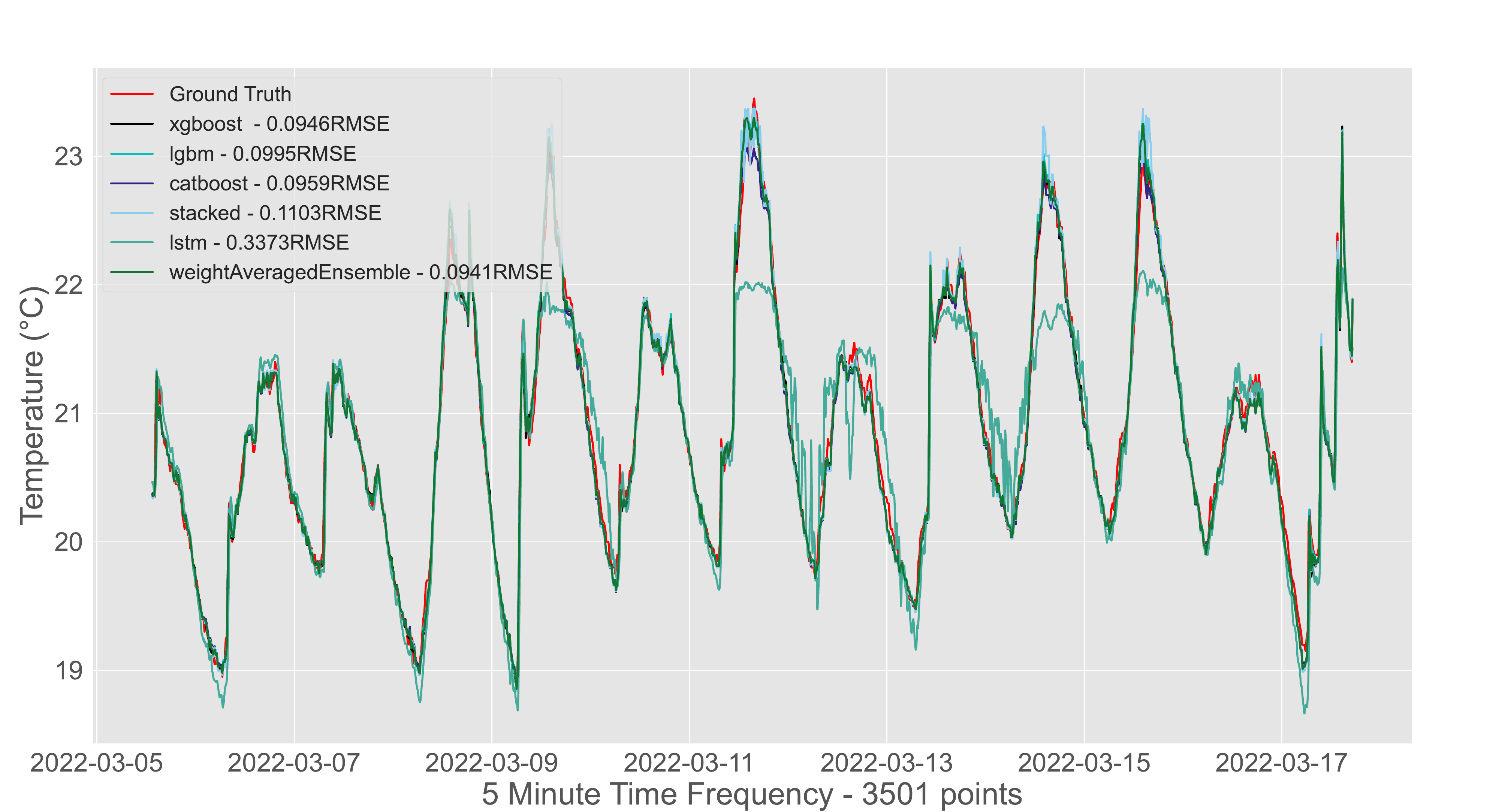}
     \caption{\scriptsize{2LivingRoomHumidifier }}
   \end{subfigure}
   \begin{subfigure}[b]{0.5\linewidth}
     \centering
     \includegraphics[width=\linewidth]{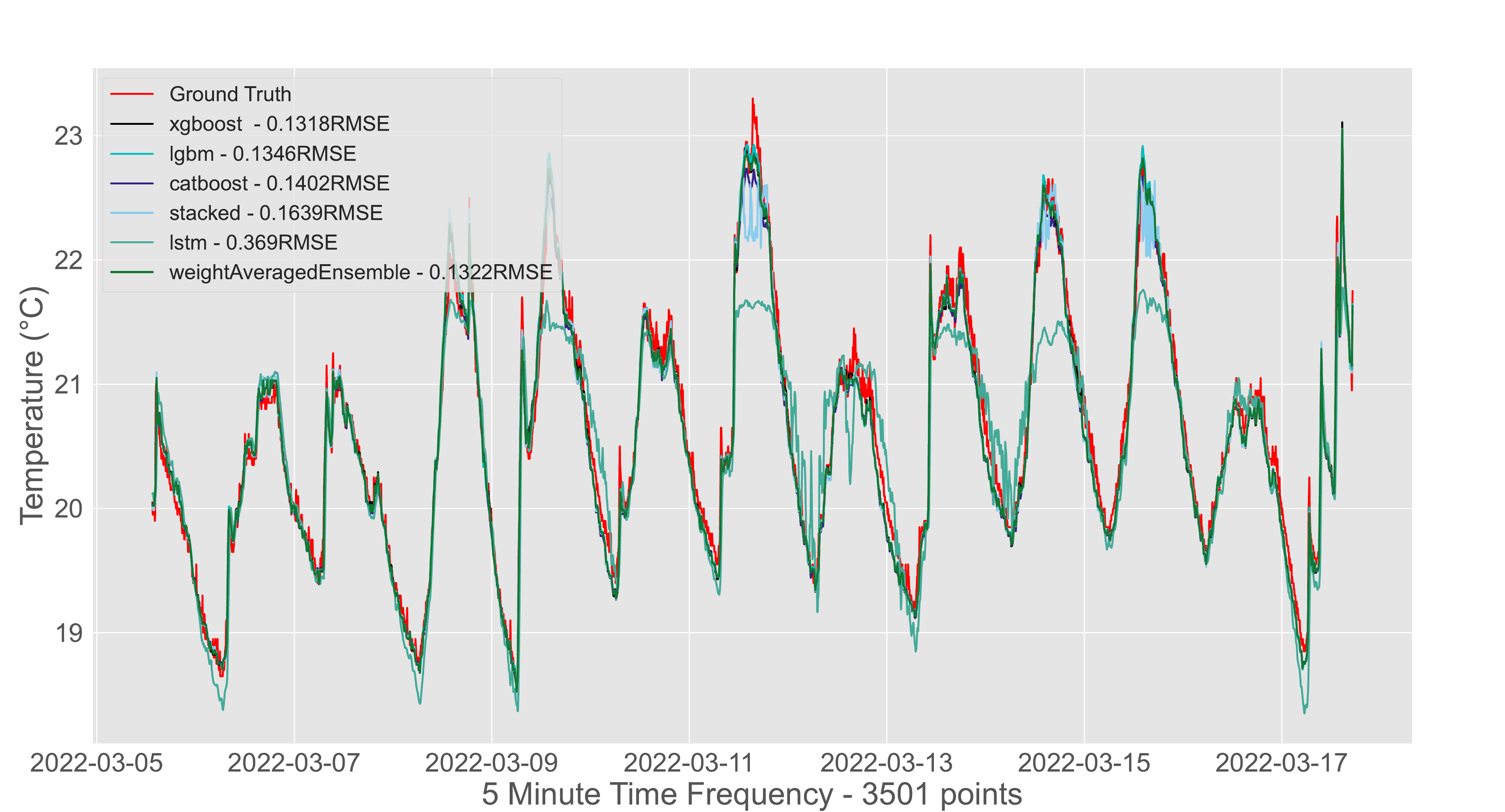}
     \caption{\scriptsize{2LRWindow}}
   \end{subfigure}
  }
  \makebox[\linewidth]{%
   \begin{subfigure}[b]{0.5\linewidth}
     \centering
     \includegraphics[width=\linewidth]{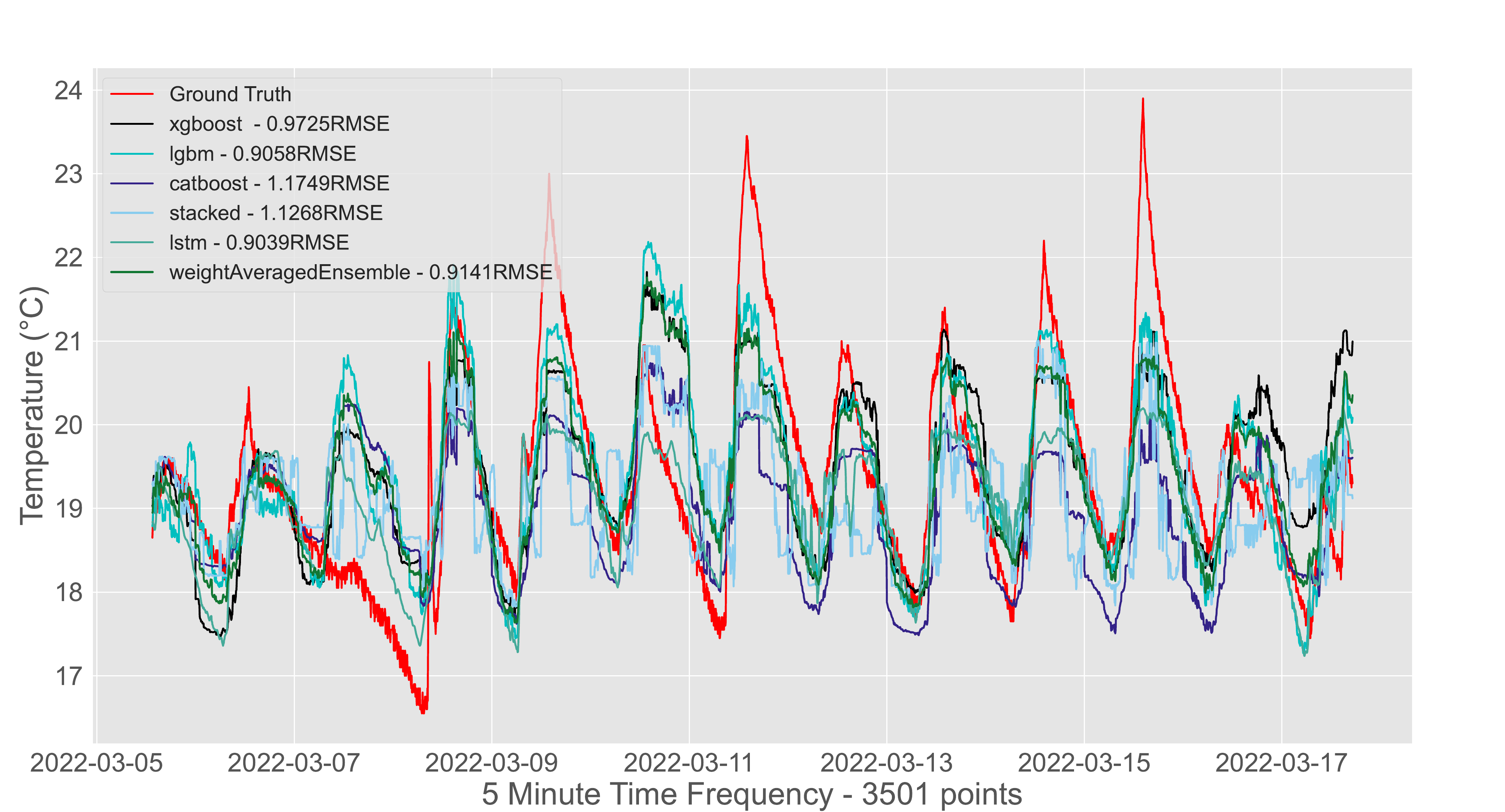}
     \caption{\scriptsize{2OfficeDesk}}
   \end{subfigure}
   \begin{subfigure}[b]{0.5\linewidth}
     \centering
     \includegraphics[width=\linewidth]{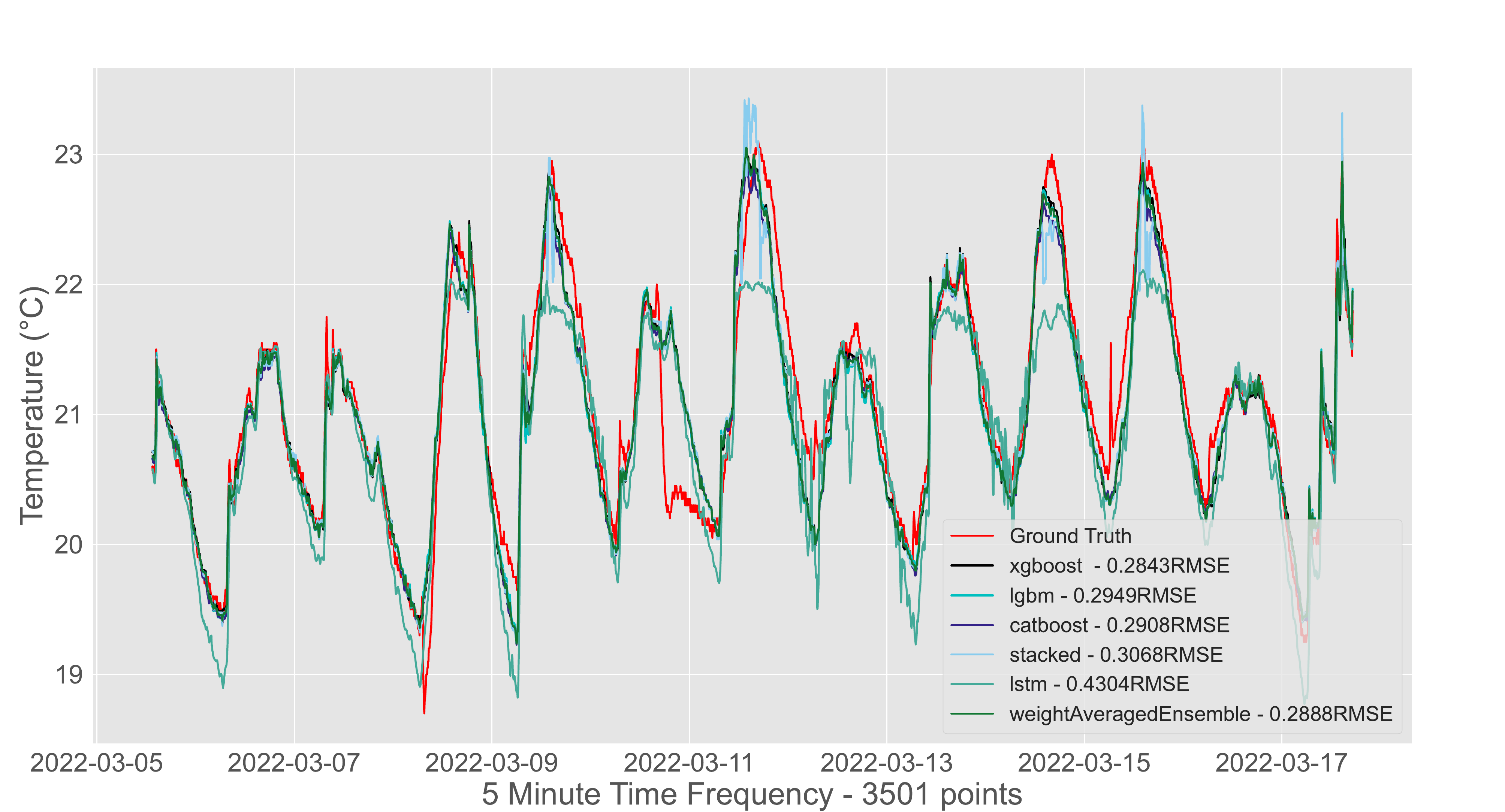}
     \caption{\scriptsize{2Stair}}
   \end{subfigure}
  }
    \caption{Temperature predictions at different sensor locations when the temperature profile at the fireplace is known. This approach can be used to fill in missing data due to sensor malfunction or should need to decommission some of the sensors arises in the future. The Fig.~shows the performance of different regression models. The best model is the weight averaging ensemble from Fig.~\ref{fig:prediction_pipeline}, which weighs the contribution of each model based on their validation set RMSE.}
    \label{fig:missing_data_model_performance}
\end{figure*}

To fill in missing temperature values for eight sensors when only the fireplace sensor is available, the predictive model pipeline can be used, as shown in Fig.~\ref{fig:missing_data_model_performance}. This ensures that the temperature in the room can still be tracked even if some sensors go out of commission in the future.

\begin{figure}
    \includegraphics[width=\linewidth]{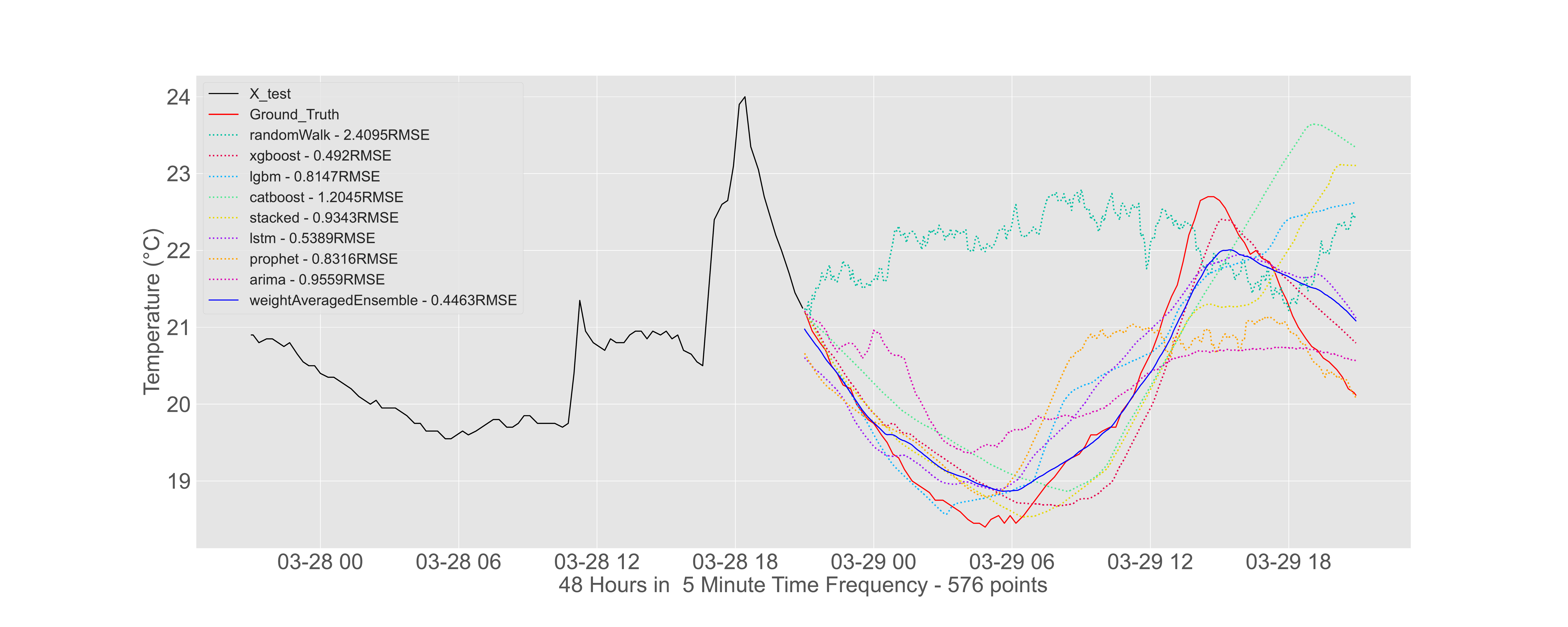}
    \caption{2Fireplace temperature forecast. Given 24 hours of past fireplace sensor data from Disruptive Technologies temperature sensor "2Fireplace" (black graph), these models are constructed to forecast 24 hours into the future of said sensor. The best forecasting model is the weight averaging ensemble (blue graph), weighing the contribution of each model seen in Fig.~\ref{fig:forecasting_pipeline}. Note that the random walk is not part of the forecasting model, but merely a way to demonstrate that the models can learn a pattern better than a baseline coin flip forecast. The ground truth is the red graph where the models are compared to that.}
    \label{fig:forecasting_fireplace}
\end{figure}

\begin{figure} 
   \makebox[\linewidth]{%
   \begin{subfigure}[b]{0.5\linewidth}
     \centering
     \includegraphics[width=\linewidth]{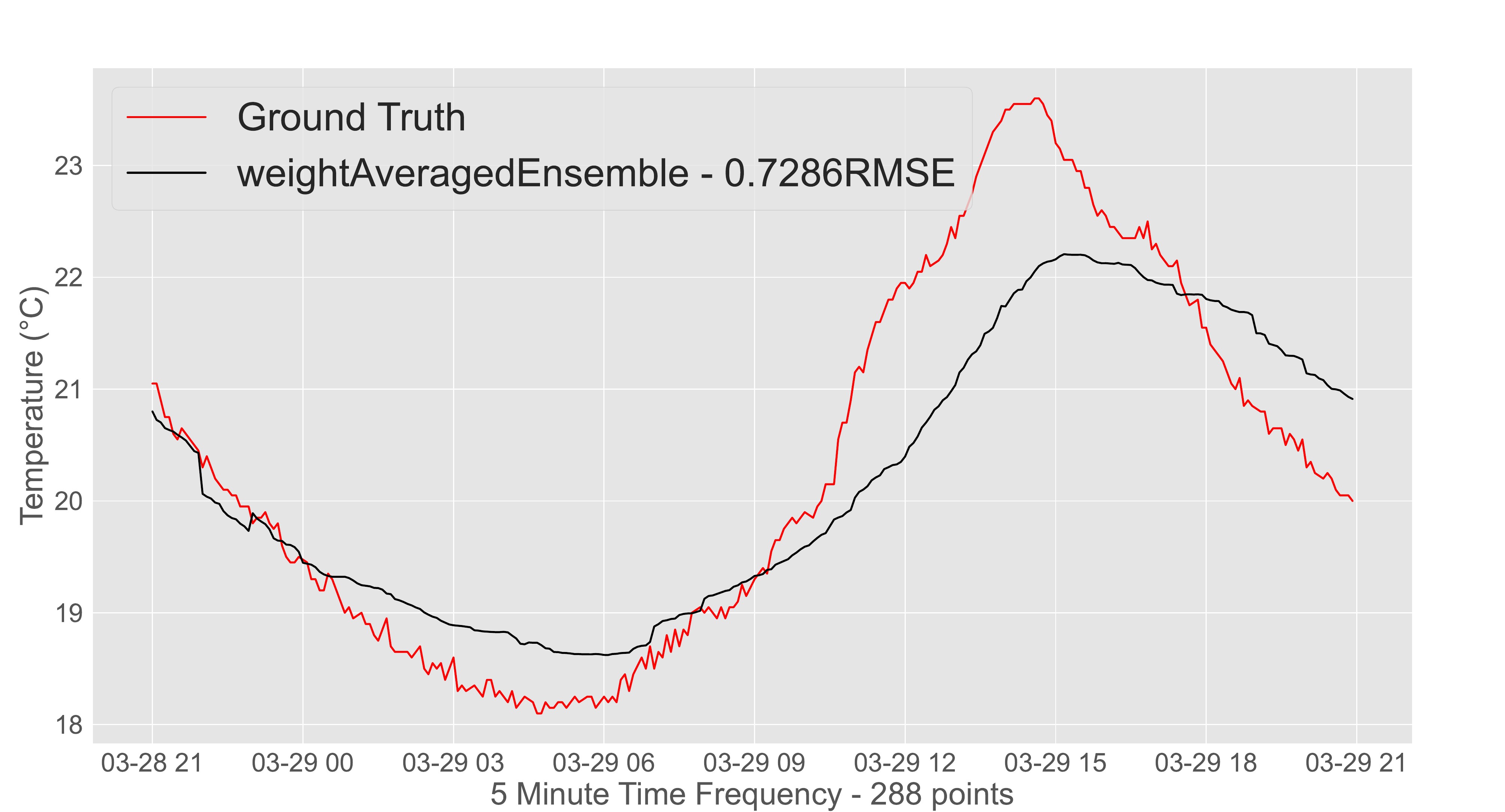}
     \caption{\scriptsize{2BalconyEntrance}}
   \end{subfigure}
   \begin{subfigure}[b]{0.5\linewidth}
     \centering
     \includegraphics[width=\linewidth]{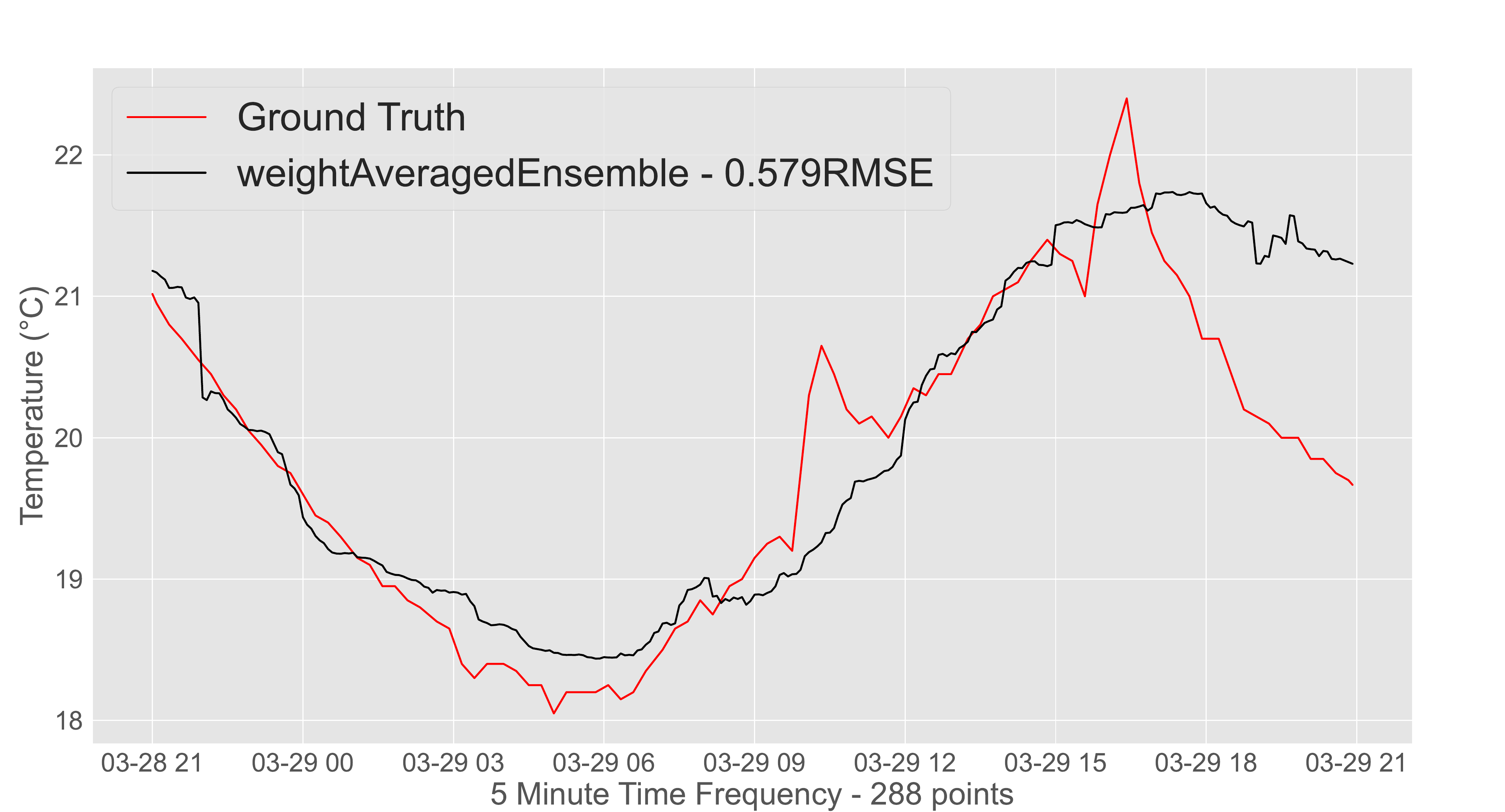}
     \caption{\scriptsize{2Cooking}}
   \end{subfigure}
   }
   
   \makebox[\linewidth]{%
      \begin{subfigure}[b]{0.5\linewidth}
     \centering
     \includegraphics[width=\linewidth]{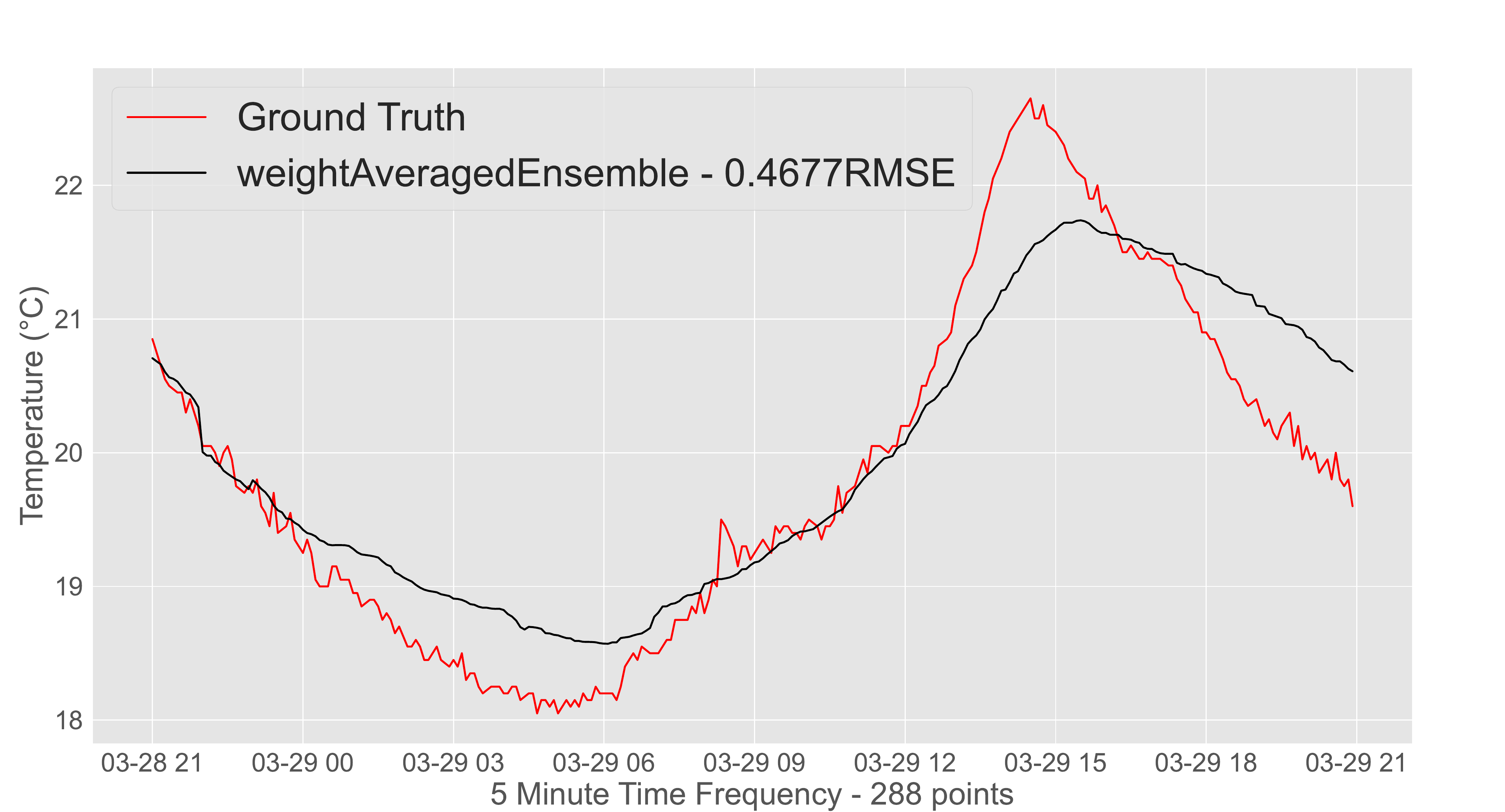}
     \caption{\scriptsize{2LivingRoomCenter}}
   \end{subfigure}
   \begin{subfigure}[b]{0.5\linewidth}
     \centering
     \includegraphics[width=\linewidth]{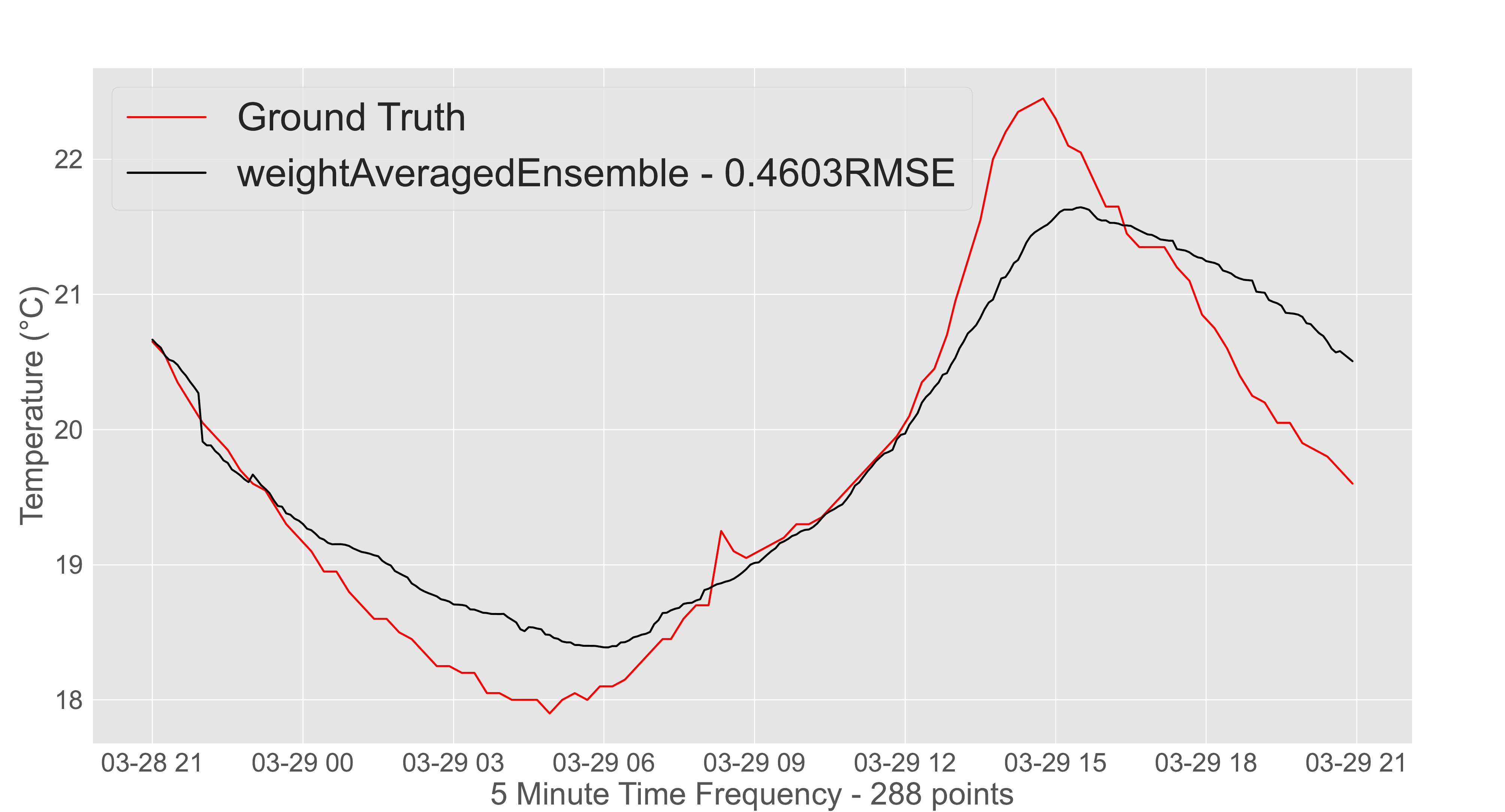}
     \caption{\scriptsize{2LivingRoomCenterHumidity}}
   \end{subfigure}
   }
   
 \makebox[\linewidth]{%
   \begin{subfigure}[b]{0.5\linewidth}
     \centering
     \includegraphics[width=\linewidth]{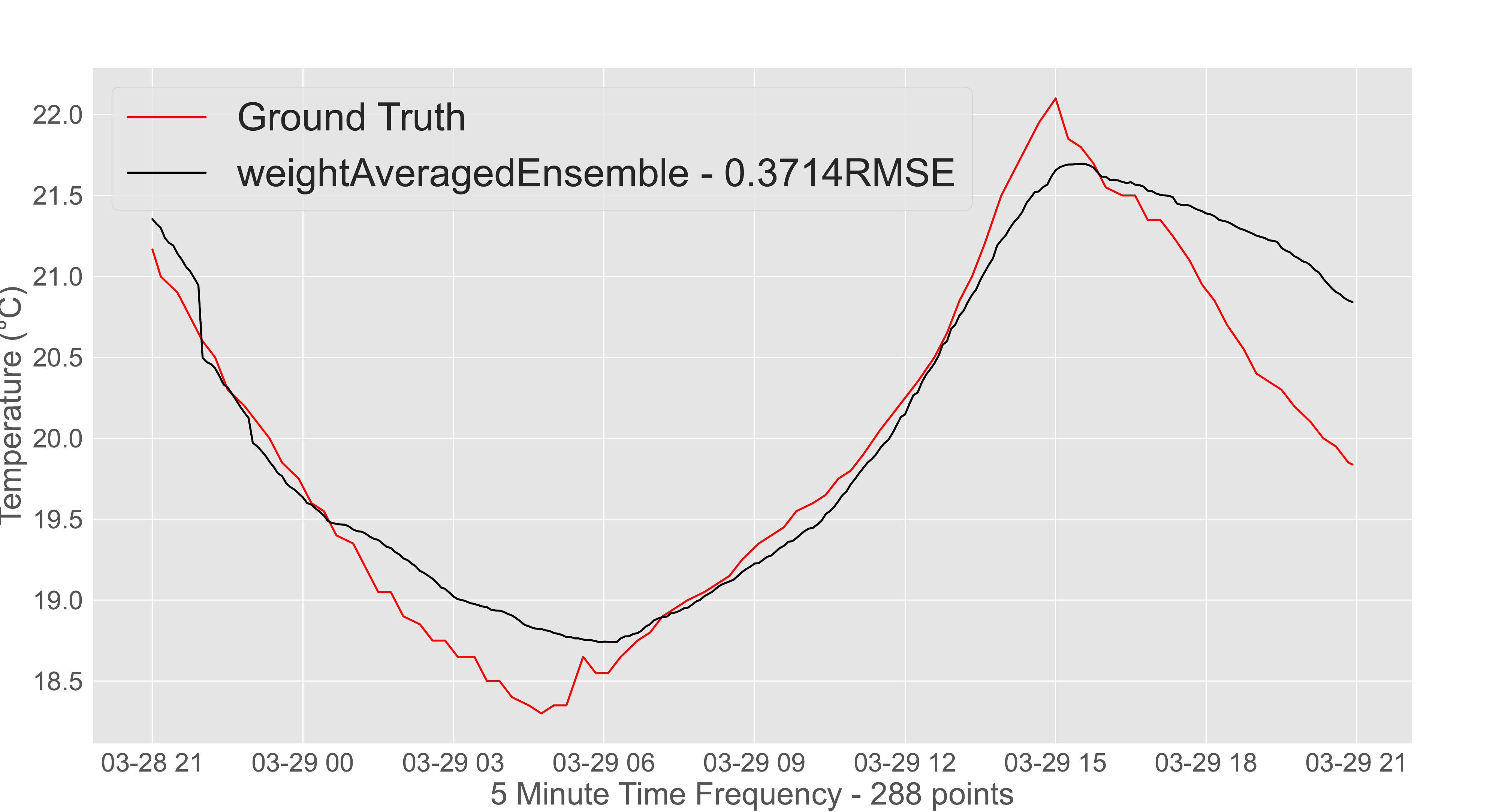}
     \caption{\scriptsize{2LivingRoomHumidifier }}
   \end{subfigure}
   \begin{subfigure}[b]{0.5\linewidth}
     \centering
     \includegraphics[width=\linewidth]{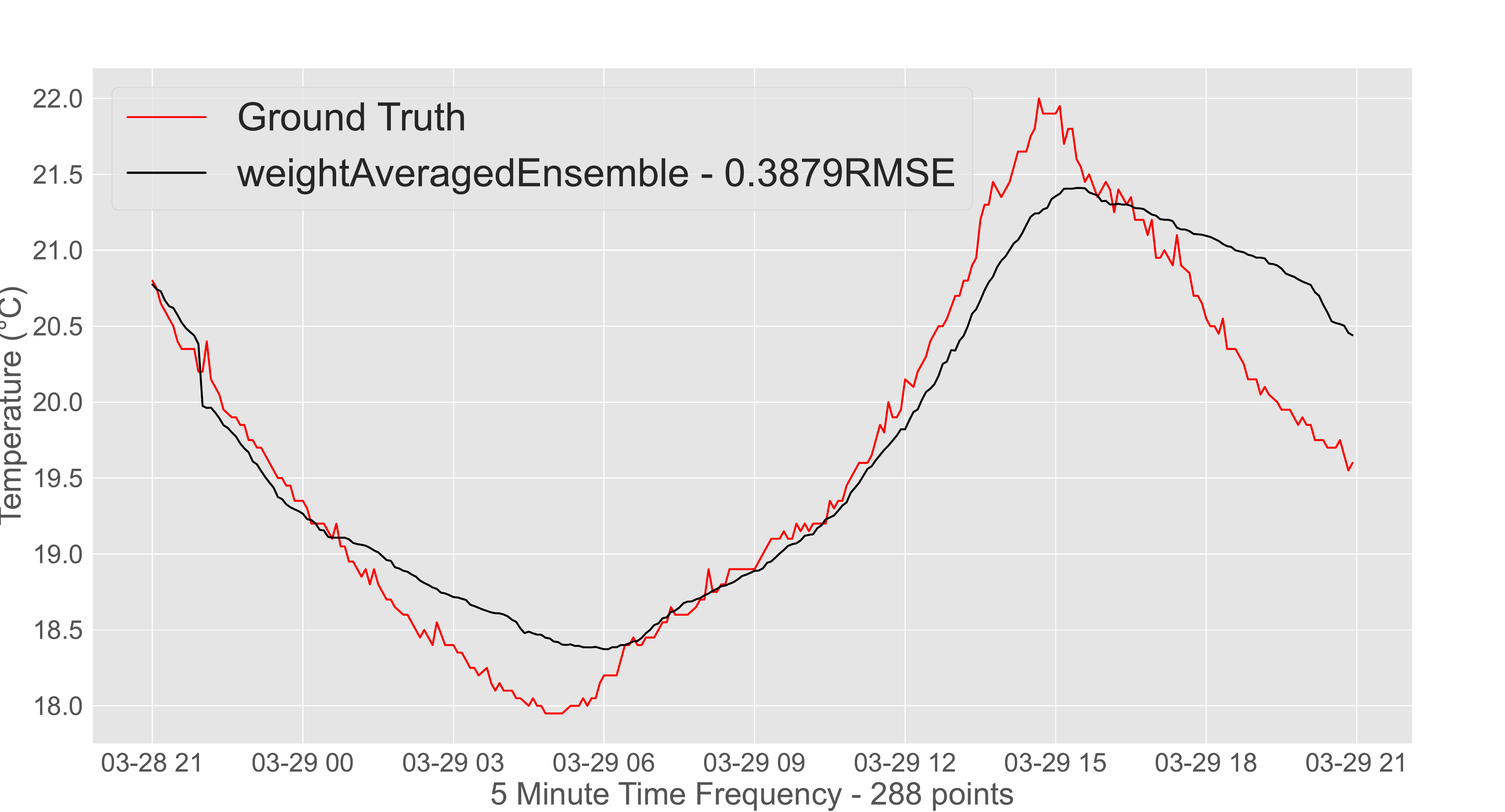}
     \caption{\scriptsize{2LRWindow}}
   \end{subfigure}
  }
  \makebox[\linewidth]{%
   \begin{subfigure}[b]{0.5\linewidth}
     \centering
     \includegraphics[width=\linewidth]{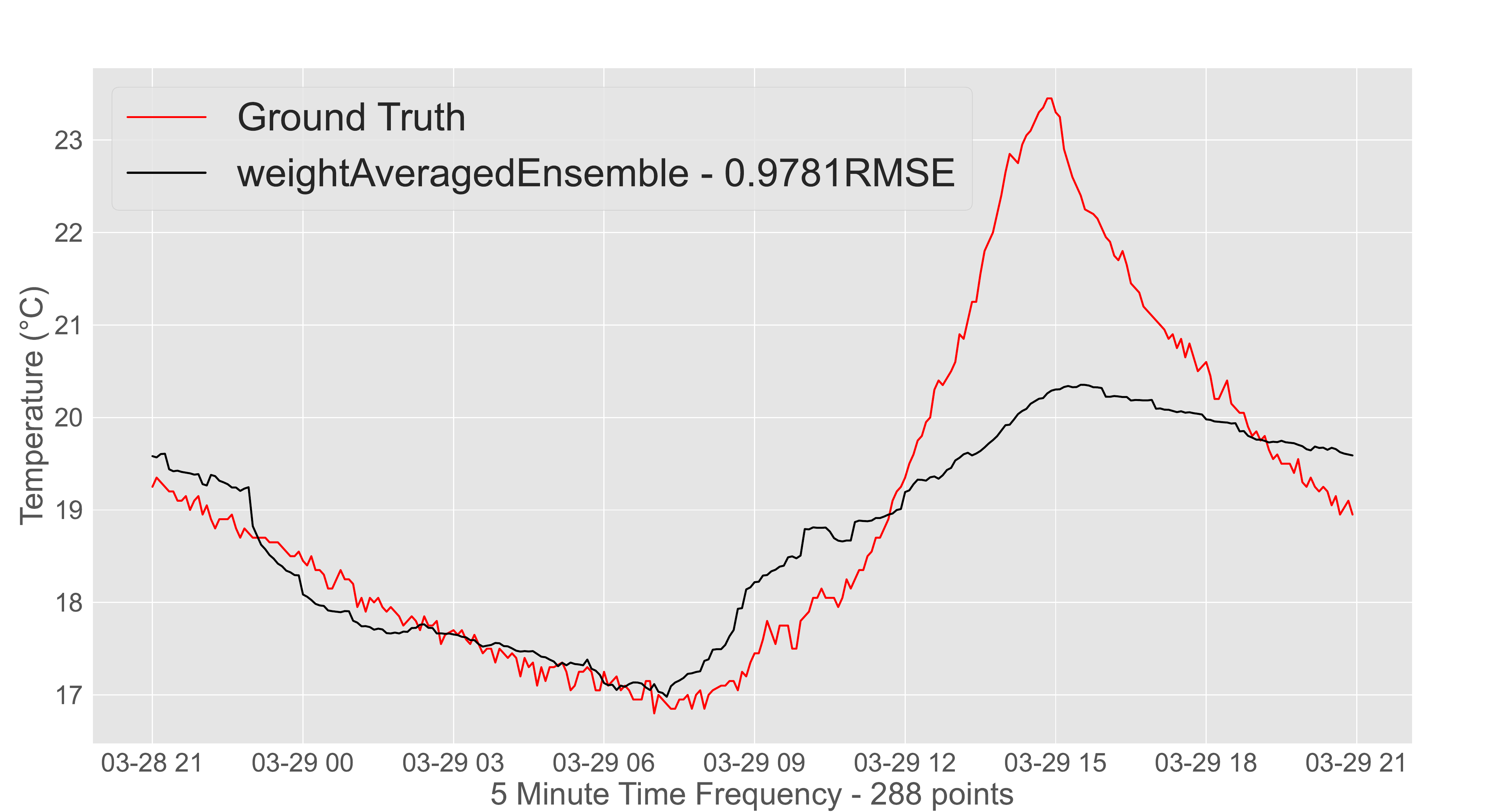}
     \caption{\scriptsize{2OfficeDesk}}
   \end{subfigure}
   \begin{subfigure}[b]{0.5\linewidth}
     \centering
     \includegraphics[width=\linewidth]{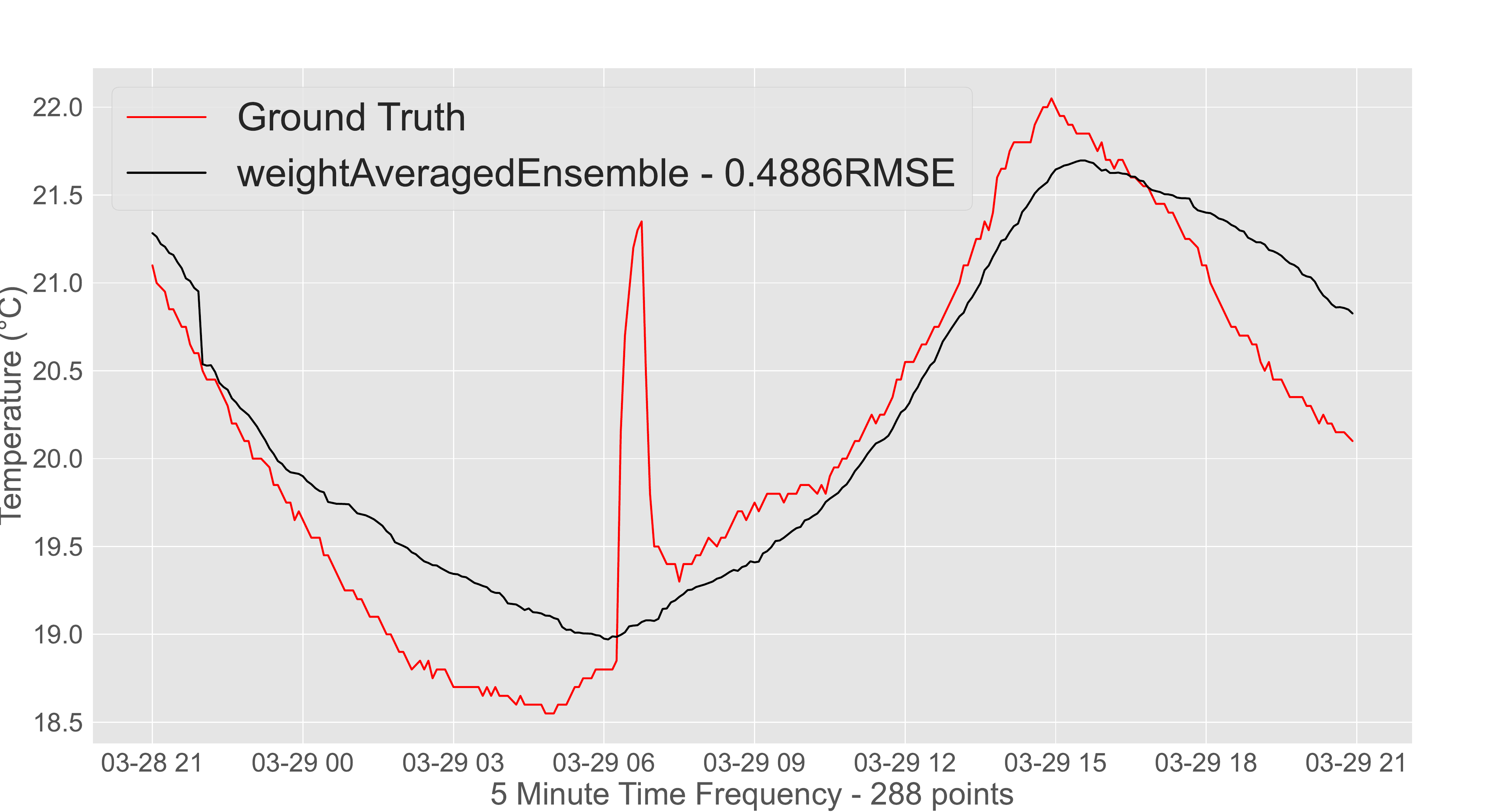}     \caption{\scriptsize{2Stair}}
   \end{subfigure}
  }
    \caption{Using the output of the fireplace forecaster in Fig.~\ref{fig:forecasting_fireplace}, we can feed it to the weighted average prediction model from Fig.~\ref{fig:prediction_pipeline}, similarly to the one used to output Fig.~\ref{fig:missing_data_model_performance} and generate forecasts for the remaining second floor temperatures.}
    \label{fig:forecasting_restof_secondfloor_usingfireplace}
\end{figure}

\begin{figure}
   \begin{subfigure}{0.475\linewidth}
     \centering
     \includegraphics[width=\linewidth]{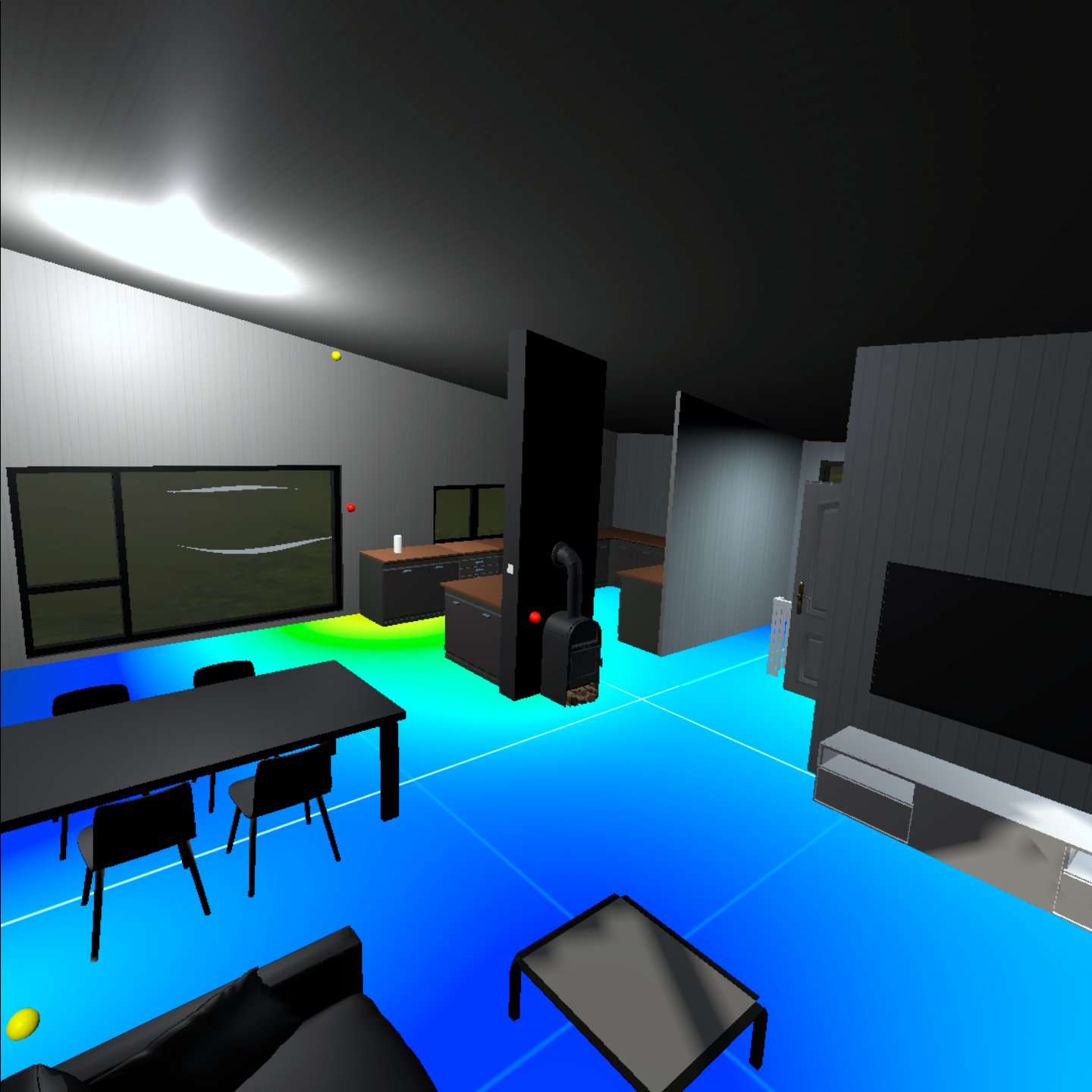}
     \caption{00:00.}
   \end{subfigure}
   \begin{subfigure}{0.475\linewidth}
     \centering
     \includegraphics[width=\linewidth]{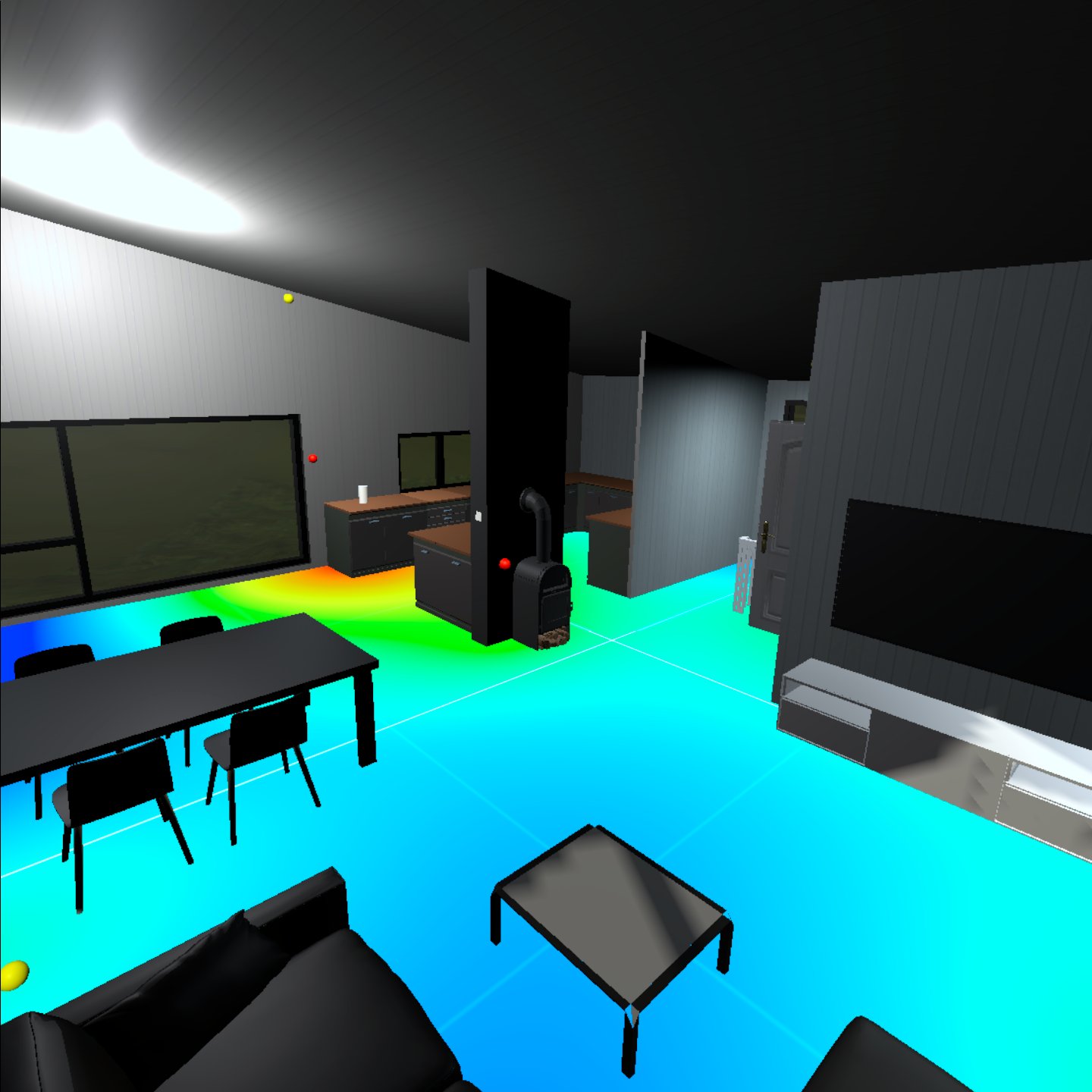}
     \caption{04:00.}
   \end{subfigure}\\
      \begin{subfigure}{0.475\linewidth}
     \centering
     \includegraphics[width=\linewidth]{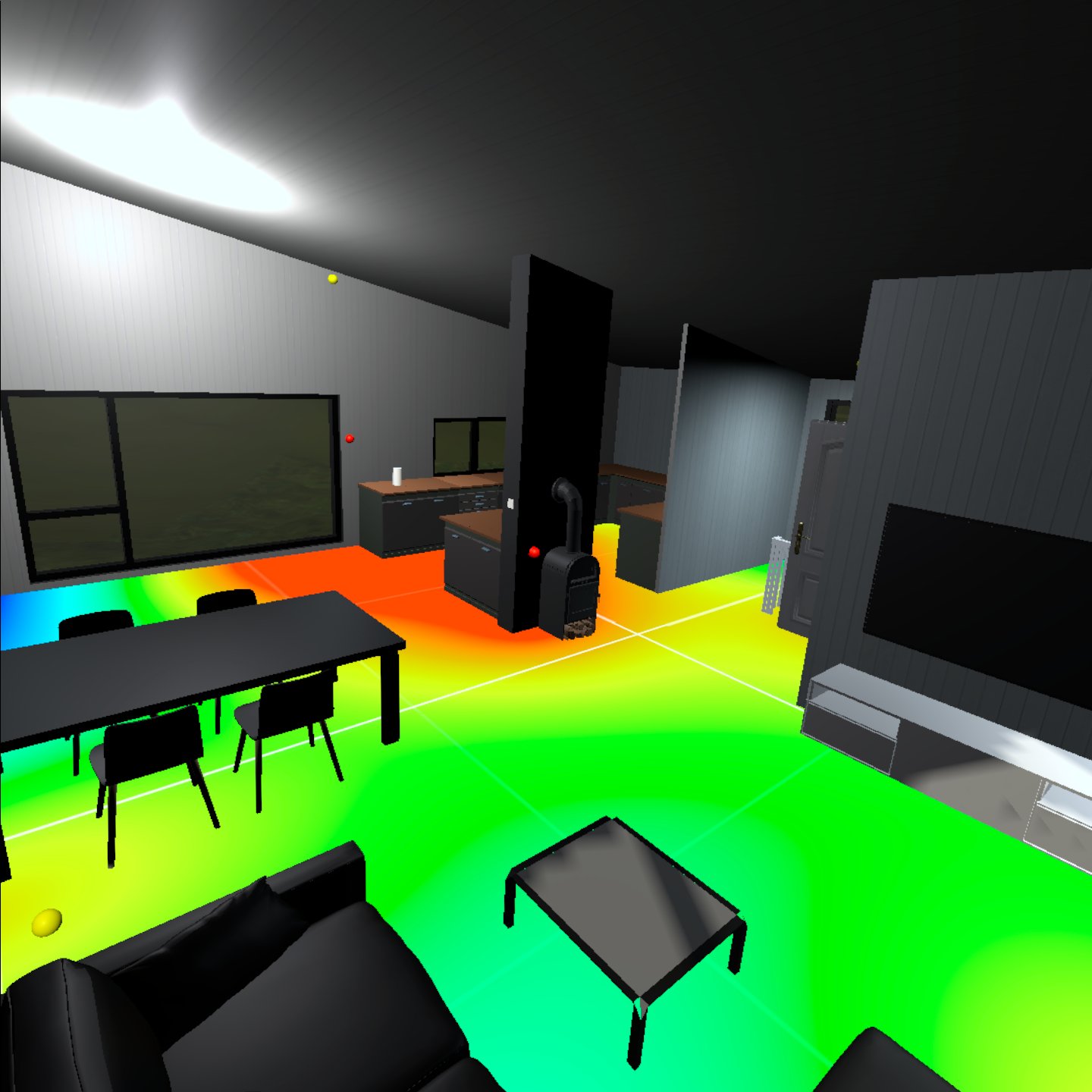}
     \caption{10:00.}
   \end{subfigure}
   \begin{subfigure}{0.475\linewidth}
     \centering
     \includegraphics[width=\linewidth]{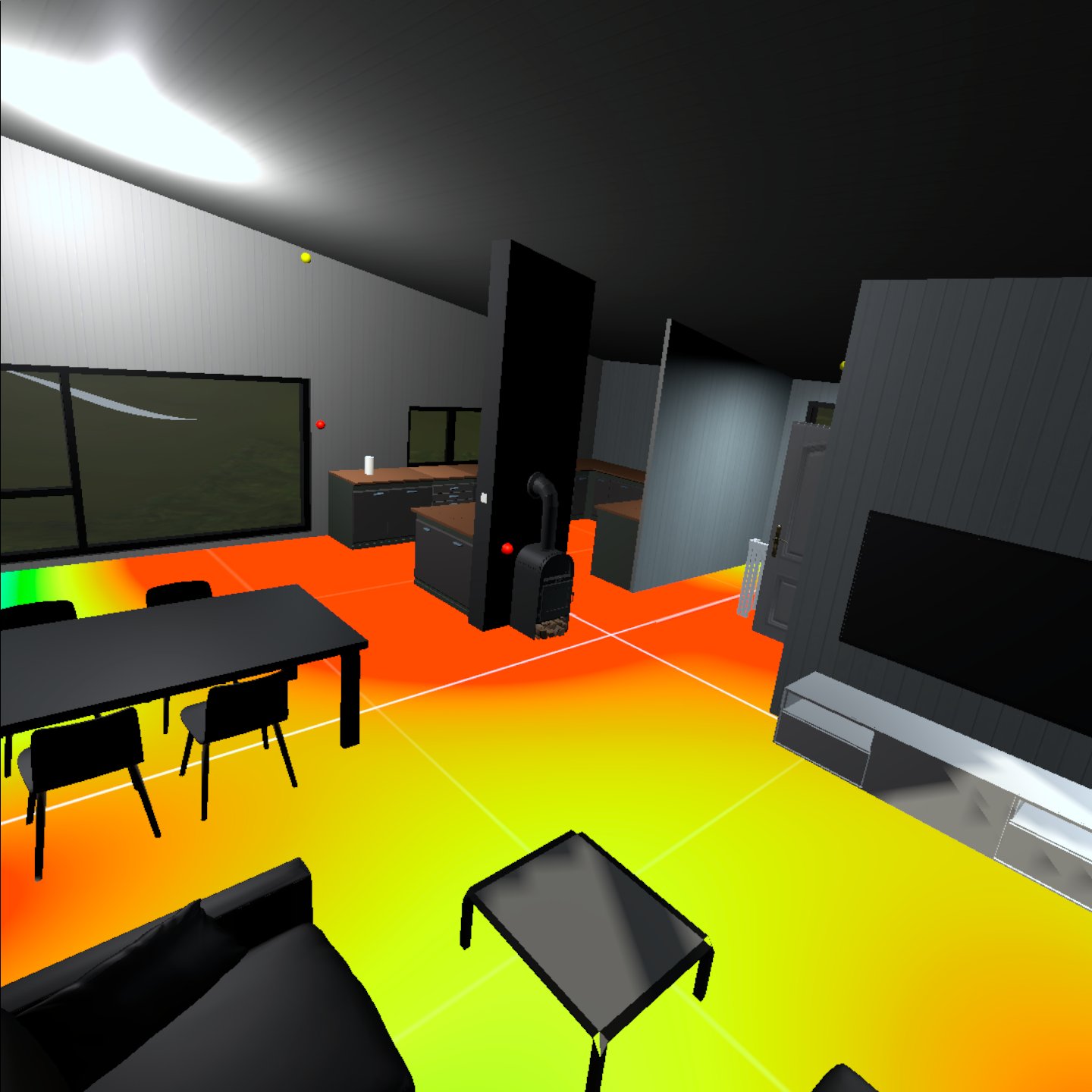}
     \caption{12:00.}
   \end{subfigure}\\
   \begin{subfigure}{0.475\linewidth}
     \centering
     \includegraphics[width=\linewidth]{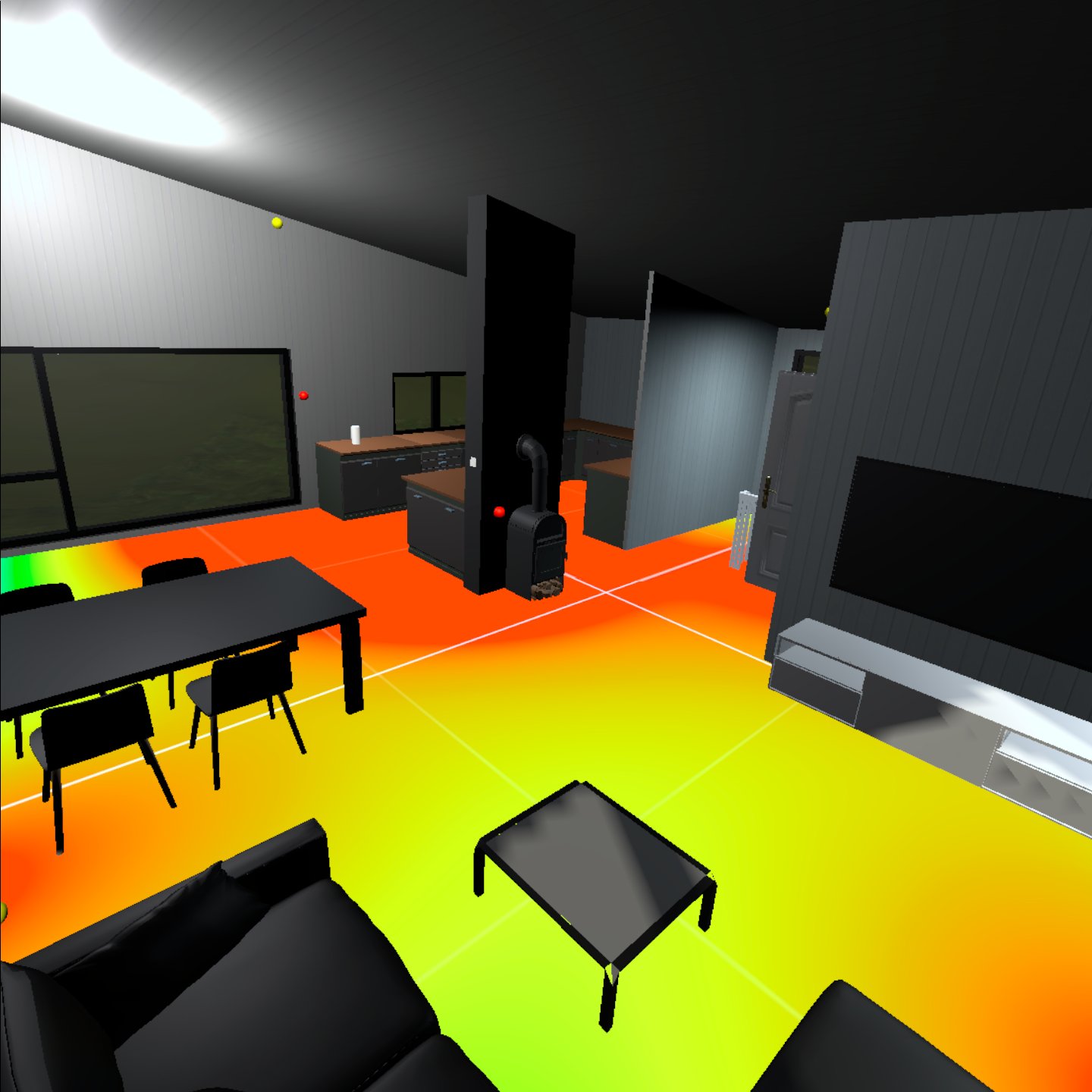}
     \caption{16:00.}
   \end{subfigure}
   \begin{subfigure}{0.475\linewidth}
     \centering
     \includegraphics[width=\linewidth]{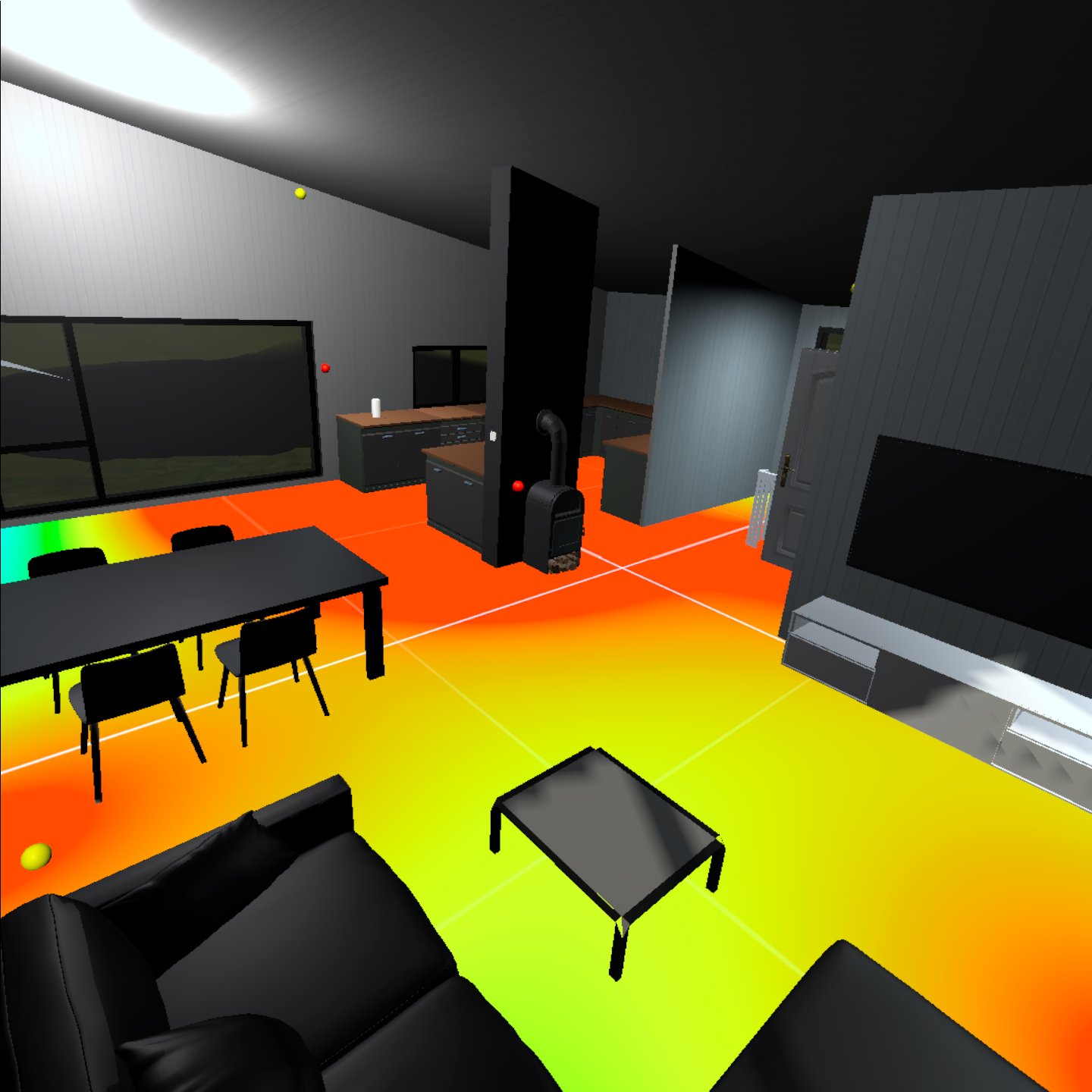}
     \caption{20:00.}
   \end{subfigure}
        \caption{Using the forecasting model to forecast the next 24 hours in the future (29.03.2022), visualized in the temperature heat map made for the diagnostic DT. The visualization shows how the model believes that the temperature on the second floor will develop the next day.}
    \label{fig:temperature-forecast-vr}
\end{figure}

Fig.~\ref{fig:forecasting_fireplace} shows how different timeseries forecasting models predict the future temperature profile at the fireplace (2Fireplace) during a typical day when the fireplace is routinely lit. The black line represents the profile used to train the model for the next 24 hours. The weight averaging ensemble model was found to be the best for making accurate forecasts. Once the future temperature profile is available, it can be used to predict the temperature profiles at other sensor locations. Fig.~\ref{fig:forecasting_restof_secondfloor_usingfireplace} presents the weight averaging ensemble predictions at all the other locations, showing good agreement with what was later observed. Finally, these point measurements are converted into contour plots and projected onto the floor surface for visualization purposes in Fig.~\ref{fig:temperature-forecast-vr}.
 
\subsubsection{Sun Position Prediction Model Performance}\label{predictive:PBM_prediction_model}
\begin{figure}
    \centering
    \includegraphics[width=\linewidth,height=0.7\linewidth]{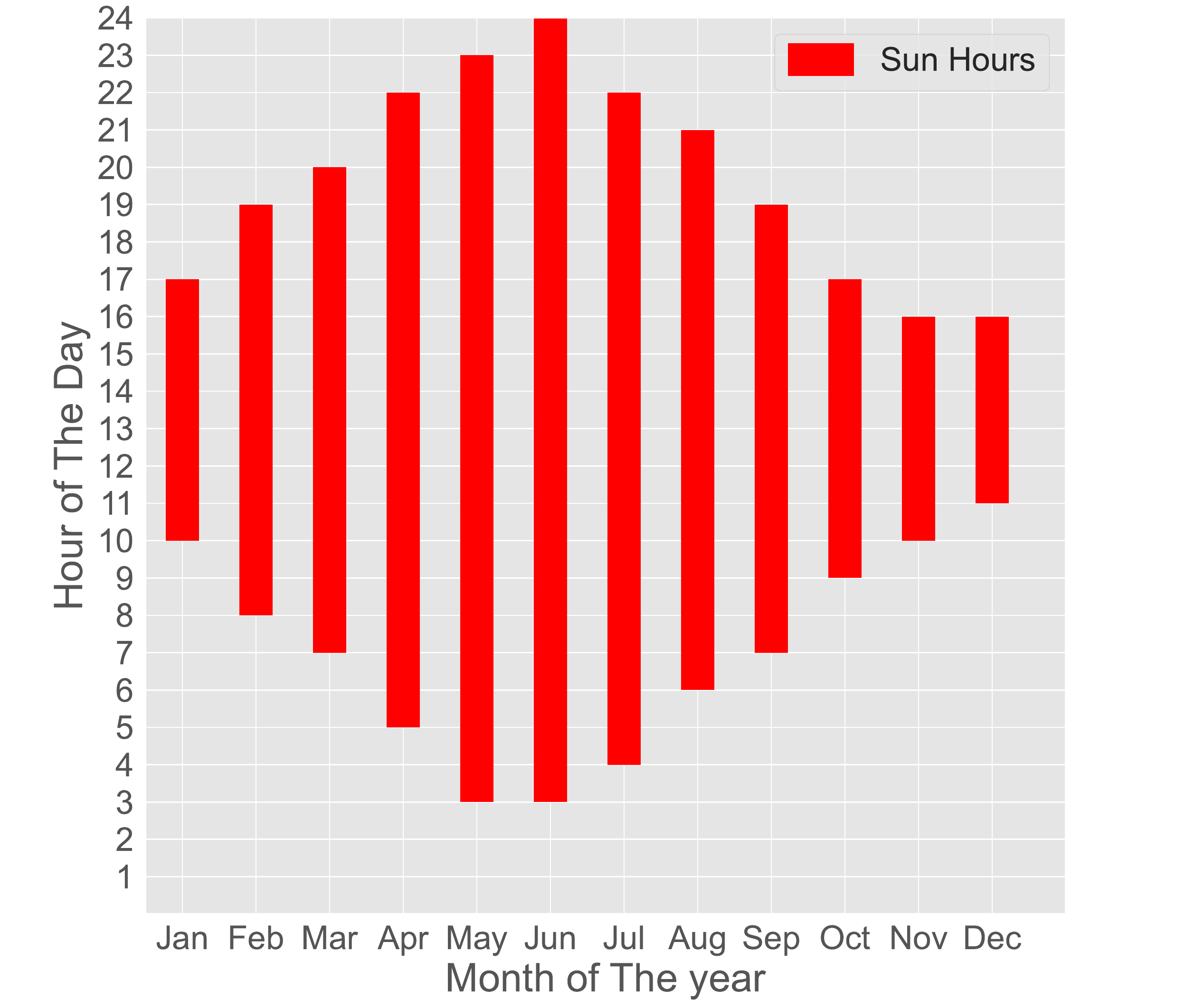}
    \caption{Predictive sun position model used to give a range of sunlight for the last day of each month in the year 2022 from the location of the house.}
    \label{fig:sun_hours_range}
\end{figure}
\begin{figure}
  \begin{subfigure}{0.475\linewidth}
     \centering
     \includegraphics[width=\linewidth]{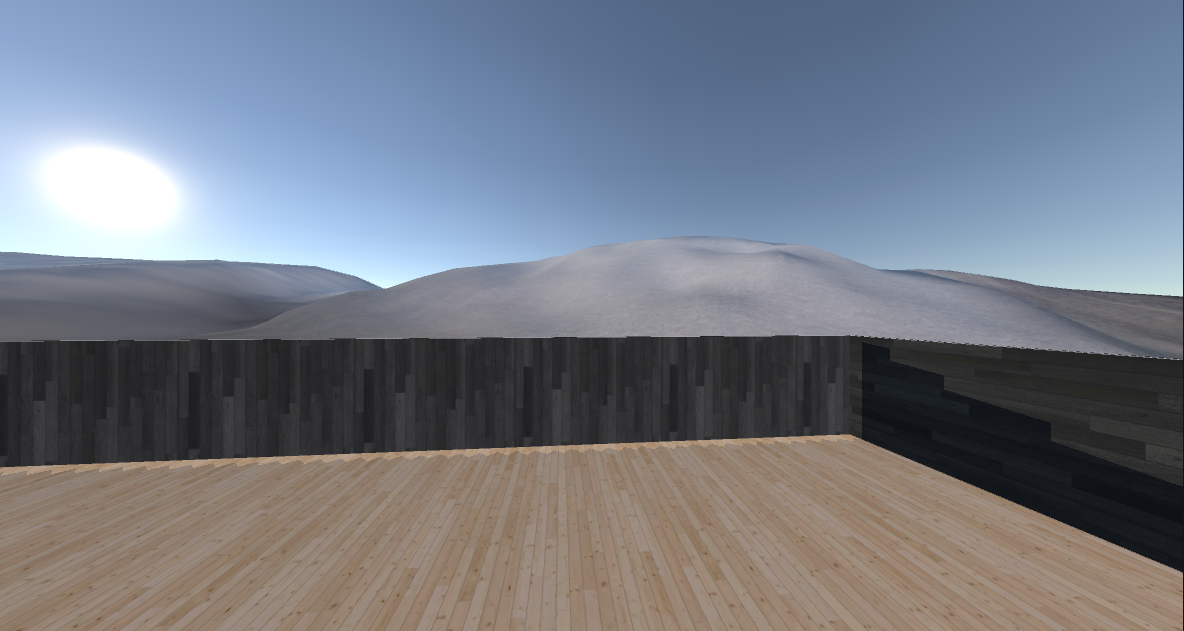}
     \caption{09:15.}
   \end{subfigure}
   \begin{subfigure}{0.475\linewidth}
     \centering
     \includegraphics[width=\linewidth]{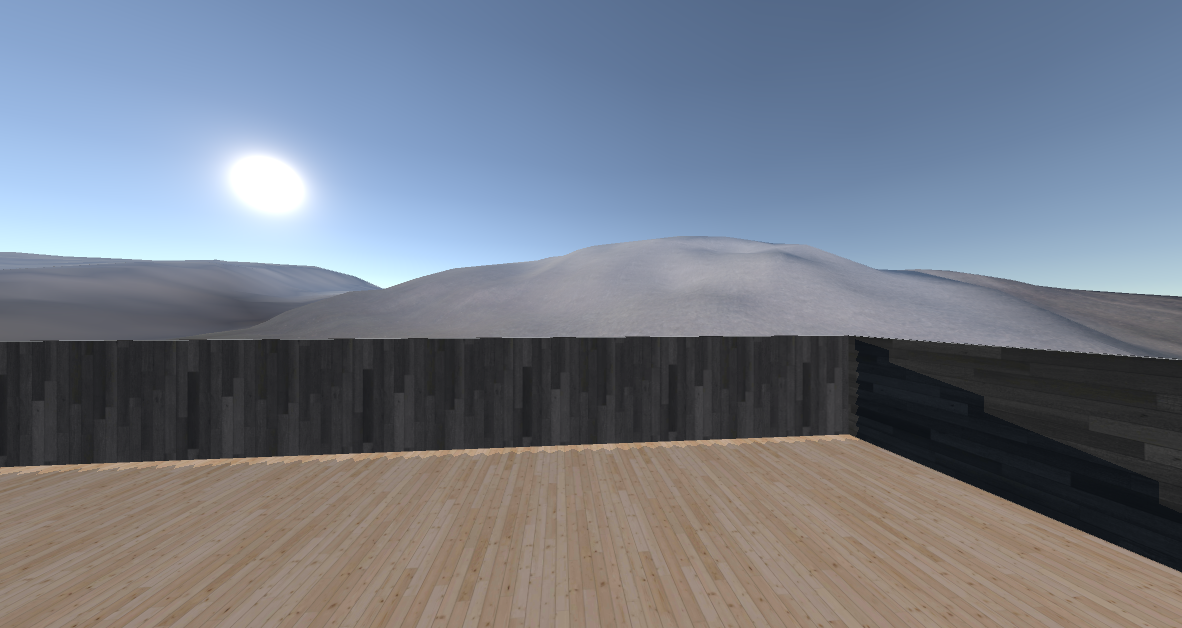}
     \caption{10:00.}
   \end{subfigure}\\
      \begin{subfigure}{0.475\linewidth}
     \centering
     \includegraphics[width=\linewidth]{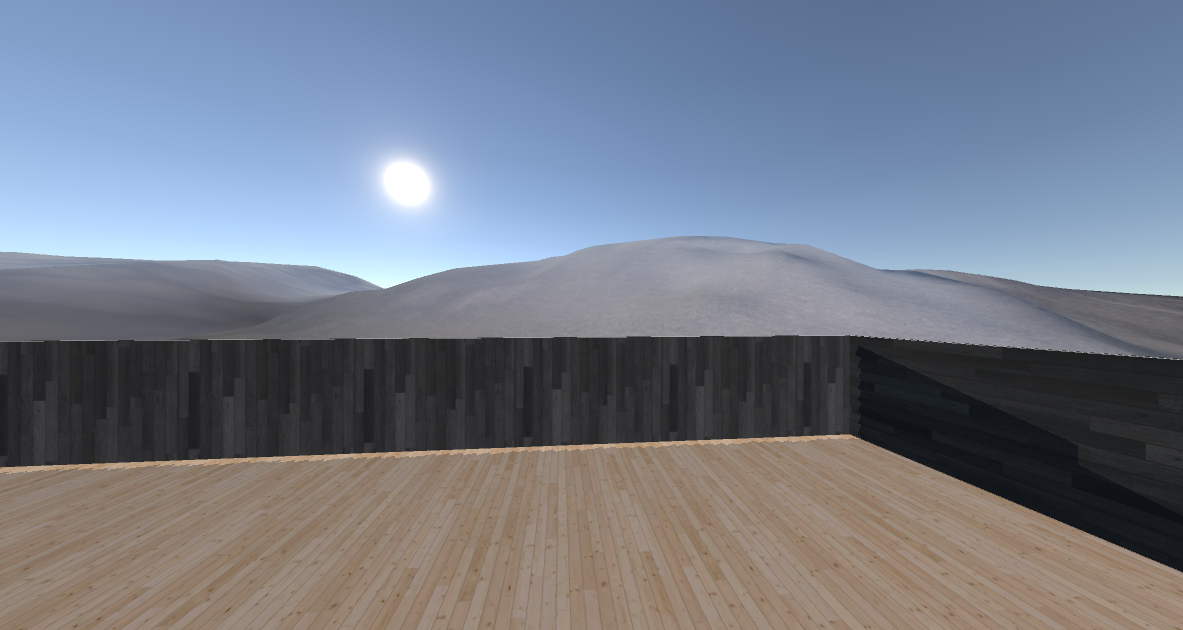}
     \caption{12:00.}
   \end{subfigure}
   \begin{subfigure}{0.475\linewidth}
     \centering
     \includegraphics[width=\linewidth]{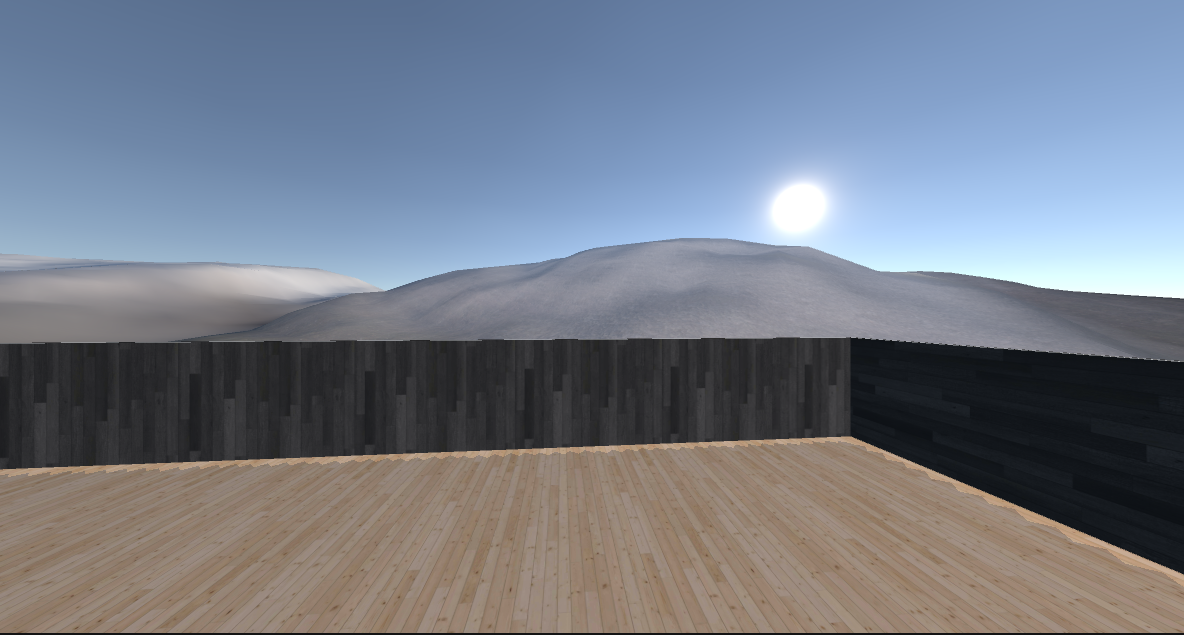}
     \caption{15:00}
   \end{subfigure}\\
   \begin{subfigure}{0.475\linewidth}
     \centering
     \includegraphics[width=\linewidth]{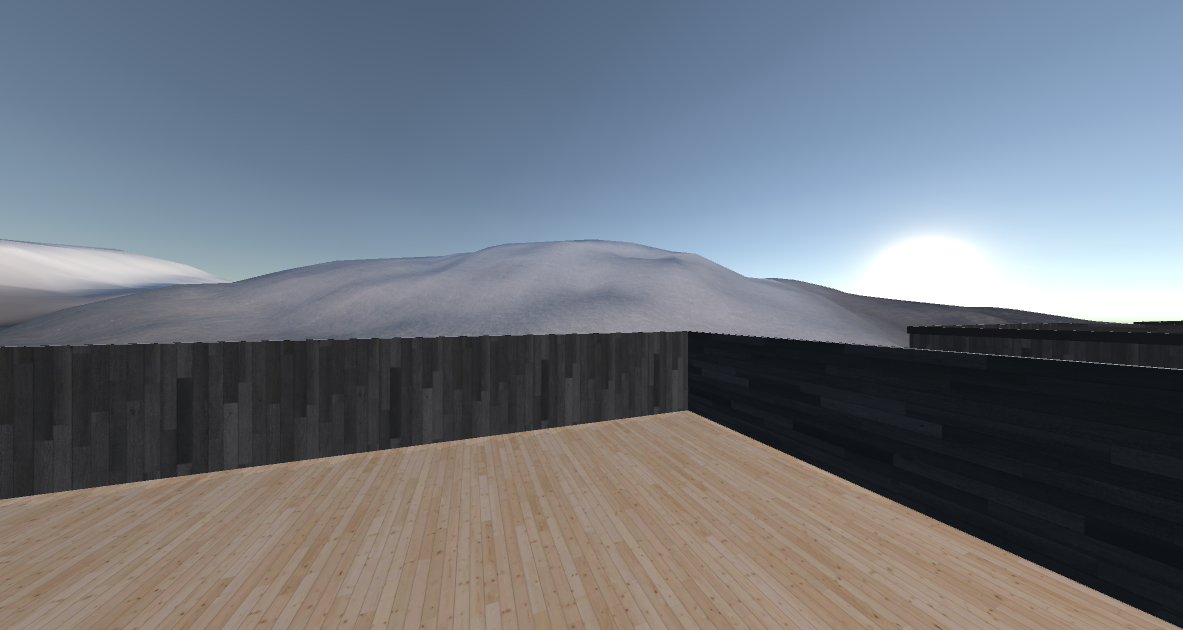}
     \caption{17:00.}
   \end{subfigure}
   \begin{subfigure}{0.475\linewidth}
     \centering
     \includegraphics[width=\linewidth]{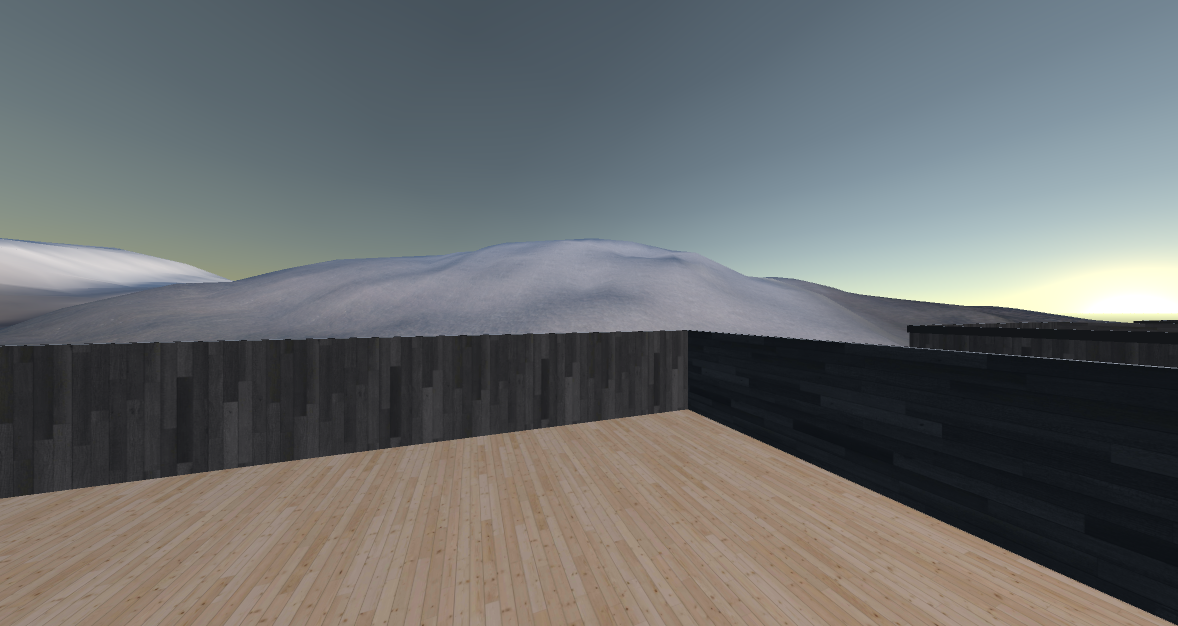}
     \caption{18:00.}
   \end{subfigure}
    \caption{Sunlight simulation in Unity using the sun position algorithm. This observation was made at 07.03.2022. Note that the view is from the balcony of the virtual house placed in the correct altitude, rotation and geographic location on a terrain generated based on a Trondheim height map from Kartverket.}
    \label{fig:Sun_in_Unity_with_terrain}
\end{figure}

Fig.~\ref{fig:sun_hours_range} shows the predictive sun model for each month, for informing decision-making around sun hours for the entire year. Furthermore, Fig.~\ref{fig:Sun_in_Unity_with_terrain} shows the same algorithm used in VR. The significance of knowing about how many sun-hours, a house gets in the span of a day is a key factor for any potential homeowner. Therefore having a PBM that accurately displays that future prediction both as a bar plot of the entire year as well as viewing a specific day visually within the VR setup, allows for better decision-making around the biggest investment that the majority of people go through. While the bar plot does show you the sun hours, the value of the visual demonstration is that it fills in the gaps by showing exactly which part of the house will be exposed to the sun. For instance, if the terrace or balcony does not get a lot of sun exposure, some buyers would become disinterested in the property. 

\subsection{Prescriptive DT}
\begin{figure}[!htb]
    \centering
    \includegraphics[width=\linewidth]{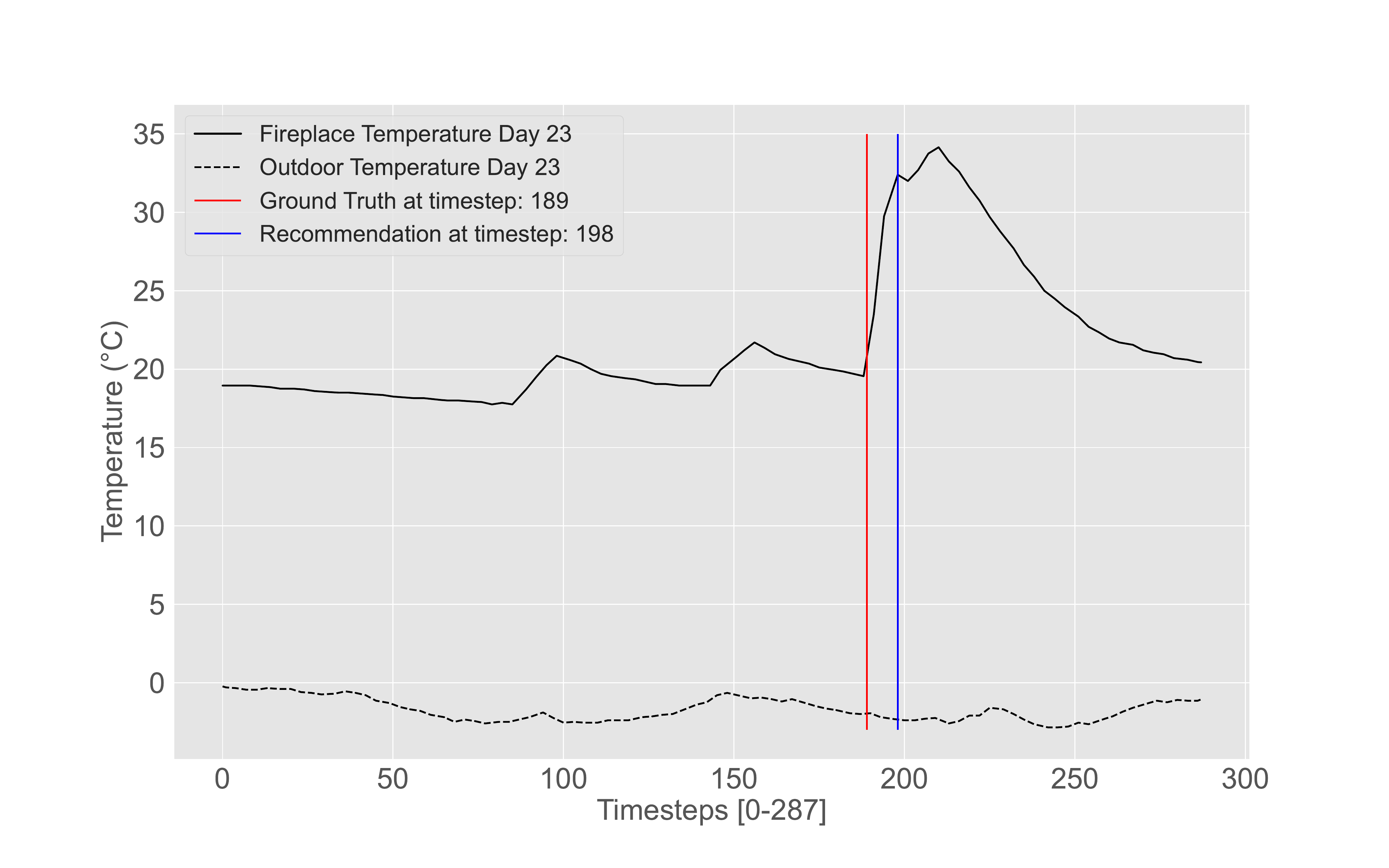}
    \caption{Recommendation to turn on the fireplace. Performance for day 23 of the dataset. 0 corresponds to midnight and the timestep corresponds to 5 minutes resulting in 288 data points, which corresponds to the 24th hour}
    \label{fig:fireplace_recommendation_performance}
\end{figure}
    \textit{Scenario:} Now that the homeowner has a predictive DT, he can foresee possible future scenarios based on past data or physics simulations. However, he is not only interested in the forecasts but specific recommendations in the house based on his past behavior. Such as what time of day is it recommended for the user to turn on the fireplace, given the temperature profile the next day. Perhaps there is also a cluster of neighbors with a DT setup, and the recommendations can be supported by all of them if they share a similar behavioral pattern. 
    
In Fig.~\ref{fig:fireplace_recommendation_performance} we are trying to predict day 23 of the 102 registered days in the dataset, using the seven other existing scenarios where the fireplace is turned on. The RMSE threshold is set to 1.5, and the days with RMSE under 1.5 are then days one, two, and 25, which resemble the outside weather of day 23 the most out of the existing samples. There isn't any other user data, such that the final recommendation is only based on the user's weighted average of other similar days, and not the Pearson correlation. Therefore the recommendation uses only the RMSE part of the UBCF pipeline in Fig.~\ref{fig:UBCF_pipeline}. We can see the recommendation for turning on the fireplace based on previous observations is very close.  

\subsection{Autonomous DT}
    \textit{Scenario:} Now imagine that the homeowner has an accurate and computationally efficient model for predicting the future state of the house and its surroundings under the influence of input changes that the homeowner can affect. These models in combination with advanced control algorithms can be used to get humans completely out of the loop. The asset can continuously update the DT while the latter can control the asset to push it towards a preset optimal operating condition.
    
Despite the ability to develop control algorithms and the availability of remotely controllable equipment like the balance ventilation system and heat pump, no effort was made to demonstrate this capability. There are several reasons for that, which we briefly mention here. Due to a lack of time, all the algorithms and modeling tools presented could not be rigorously tested. Furthermore, using black-box neural network-based methods for predictions complicates the matter. Unless the model's working is humanly interpretable, it is not wise to use them in a controlled setting. Doing so could risk the safety of the inhabitants or at the least make the manufacturer's guarantee on the equipment null and void. 

\section{Conclusion and Future Work}
\label{sec:conclusionsandfuturework}
In this work, we exploited the power of artificial intelligence, advanced sensor technologies, and virtual reality to develop a fully functional and high capability level digital twin (DT) of a modern house. The work involved creating a realistic 3D model of the house that is not only good for visualization but also for conducting engineering simulations. This corresponded to a standalone DT. The physical house was then equipped with a diverse class of sensors, and the corresponding digital representation of the house was updated accordingly. A real-time data acquisition pipeline was established to update the state of the DT with any changes in the state of the physical house, resulting in a descriptive DT. Analytics tools were applied to the incoming data to detect critical changes, resulting in a diagnostic DT. These three levels of DT were not capable of giving any information about the future state as they all relied on the incoming data. At the next level, i.e., the predictive DT,  an ensemble of pure data-driven timeseries forecasting models was built and trained to predict the future state of the house accurately. In addition, a physics-based modeling approach to predict the sun's movement and its obstruction by the local terrain was also implemented to predict the solar potential for any time in the future. At the prescriptive DT level, it was argued how data/insight from similar houses in the neighborhood could be utilized using collaborative filtering. It also demonstrated how a prescriptive DT could learn about the user's own behavior to support future recommendations. Finally, the autonomous DT wove all the subsystems together by closing the control loop. The main contributions of the work can be enumerated as follows:

\begin{itemize}
    \item We demonstrated the concept and value of DT \cite{rasheed2020dtv} and its capability levels \cite{San2021haa}. Although the asset chosen for the demonstration was a modern house, the workflow proposed is generic in nature.
    \item We have shown the concept and value of DT \cite{rasheed2020dtv}, as well as its capability levels \cite{San2021haa}. While we used a modern house as an example, the proposed workflow is applicable to various contexts.
    \item We have highlighted the significance of a diverse set of data and two distinct modeling approaches (physics-based and data-driven) to enhance the physical realism of DTs.
    \item We have demonstrated how computer graphics, specifically virtual reality technology developed with game engines, can significantly enhance the capability levels of DTs. 
    \item Through an interactive graphical interface in virtual reality, we have showcased the potential of DTs not only for remote monitoring but also for remote interaction with assets. 
    \item  Our proposed workflow and future research directions outlined in the following section can serve as a guide for developing DTs from scratch.
\end{itemize}

While we have highlighted the strengths of our work, we have also identified areas where improvements can be made. One advantage of the DT framework we developed is its modular nature, which allows for individual capabilities to be extended and improved without compromising the functionality of others. We list these areas for improvement below:

\begin{itemize}
    \item \textit{Standalone DT:} At this capability level, we manually created a 3D model of the house and its furnishings, which can be a bottleneck for scaling DT technology to encompass multiple houses in a neighborhood. However, this issue can be addressed with image-based photogrammetry, as demonstrated in recent works \cite{Yang2022ana, Hu2021c3r}. Additionally, solid models are typically represented by textured tessellated polygon surfaces, whose number can be reduced without an observable degradation in quality to enable the DT to run on low-end, affordable computing devices. 
    
    \item \textit{Descriptive DT:} Geometric change detection \cite{Sundby2021gcd} can be implemented in a descriptive DT to keep track of geometric changes within the house using cost-effective solutions like RGB cameras, limited communication bandwidth, and storage. Furthermore, real-time satellite data can be used to more accurately describe the external environment, such as cloud cover. 

    \item \textit{Diagnostic DT:} Principal component analysis or autoencoder can be used for detecting deviations from the norm in heterogeneous multivariate data to detect anomalies. Other sensor data, like temperature, humidity, noise, and air quality, are only measured in a few discrete locations. A simple interpolation scheme was implemented to create heatmaps, which can be improved by sensitizing inverse distance weighting with door states and wall corner locations \cite{DisruptiveImprovements}. Additionally, optimal sensor placement strategies \cite{Manohar2018dds} can be evaluated for more efficient use of sensors and reconstruction.

    \item \textit{Predictive DT:} In the current project, we used either a purely physics-based model or a data-driven model to predict the external and internal state of the house, but both approaches have inherent weaknesses, as discussed in \cite{San2021haa}. Recent works \cite{Blakseth2022dnn, Blakseth2022cpb} have shown how a hybrid modeling approach can address these weaknesses and make accurate and more certain predictions, making it ideal for modeling partially understood physics and addressing the issues of input parameter uncertainties. For instance, \cite{Robinson2022anc} has already shown the applicability of accurately modeling heat transfer in an aluminum extraction process, which is similar to the building energy modeling considered in our work.

    \item \textit{Prescriptive DT:} At this level, we used the DT to provide recommendations based on learning from the behavior of the same house. However, we faced challenges due to the lack of available data for training ML algorithms, as the house was newly constructed. Collaborative filtering or self-organizing maps can be useful to learn from the performance of older houses for which data exists. 
    
    \item \textit{Autonomous DT:} We could not practically demonstrate the full potential of a fully autonomous DT due to concerns of voiding equipment guarantees, such as the balance ventilation system and heat pumps. While the smart lights could be controlled remotely, no data was recorded to develop a control strategy for lighting. Therefore, recording data regarding the lighting preferences of occupants and developing a controller to satisfy those preferences would be interesting. Alternatively, research on the psychological effects of lighting on occupants can be integrated into the autonomous DT. Additionally, making the models on which decisions are made humanly interpretable is a challenge worth addressing before realizing a fully autonomous DT.
\end{itemize}

The concept of DT is rapidly advancing and this work, along with the future research directions proposed, represents only a small piece of the larger puzzle. However, this work has produced an extensible DT framework that can be valuable for educational purposes and for testing new techniques that can help make DT indistinguishable from its physical counterpart.

\section*{Acknowledgments} The project owes a great deal to the generous support of Disruptive Technologies, who provided advanced sensors for temperature, proximity, humidity, and water detection. These sensors played a critical role in demonstrating the potential of next-generation digital twins for the built environment. We are also grateful for the prize awarded by Tekna, which was invested in the development and demonstration of the project's VR component using the Oculus Quest 2 virtual reality headset.

\makenomenclature
\printnomenclature
\nomenclature{$ARIMA$}{AutoRegressive Integrated Moving Average}\nomenclature{$RMSE$}{Root Mean Squared Error}
\nomenclature{AI}{Artificial Intelligence}
\nomenclature{AR}{AutoRegressive}
\nomenclature{MA}{Moving Average}
\nomenclature{API}{Application Program Interface}
\nomenclature{I}{Integration}
\nomenclature{RGB}{Red Green Blue}
\nomenclature{HSV}{Hue Saturation Value}
\nomenclature{UI}{User Interface}
\nomenclature{IoT}{Internet of Things}
\nomenclature{VR}{Virtual Reality}
\nomenclature{FEM}{Finite Element Method}
\nomenclature{RL}{Reinforcement Learning}
\nomenclature{MPC}{Model Predictive Control}
\nomenclature{CFD}{Computational Fluid Dynamics}
\nomenclature{UBCF}{User-Based Collaborative Filtering}
\nomenclature{MSE}{Mean Squared Error}
\nomenclature{ADAM}{Adaptive Moment Estimation}
\nomenclature{CAD}{Computer Aided Design}
\nomenclature{URP}{Universal Render Pipeline}
\nomenclature{HDRP}{High Definition Render Pipeline}
\nomenclature{RNN}{Recurrent Neural Network}
\nomenclature{HAM}{Hybrid Analysis and Modeling}
\nomenclature{GBM}{Gradient Boosting Machine}
\nomenclature{LSTM}{Long Short Time Memory}
\nomenclature{ML}{Machine Learning}
\nomenclature{BIM}{Building Information Model}
\nomenclature{PBM}{Physics-Based Modeling}
\nomenclature{DDM}{Data-Driven Modeling}
\nomenclature{DT}{Digital Twin}
\nomenclature{$C$}{Sun's Center}
\nomenclature{$JD$}{Julian Date}
\nomenclature{$JC$}{Julian Century}
\nomenclature{$L_0$}{Sun's Mean Longitude}
\nomenclature{$M_0$}{Sun's Mean Anomaly}
\nomenclature{$\lambda$}{Sun's Ecliptic Longitude}
\nomenclature{$\beta$}{Sun's Ecliptic Latitude}
\nomenclature{$\Omega$}{The Earth's Obliquity of the Ecliptic}
\nomenclature{$\alpha$}{Right Ascension}
\nomenclature{$\delta$}{Declination}
\nomenclature{$t_{UT}$}{Universal Time}
\nomenclature{$t_{SR}$}{Sidereal Time Greenwich}
\nomenclature{$t_{SRUT}$}{Sidereal Time Greenwich for Universal time}
\nomenclature{$t_{LSR}$}{Local Sidereal Time}
\nomenclature{$HA$}{Hour Angle of Object}
\nomenclature{$L$}{Input Longitude}
\nomenclature{$B$}{Input Latitude}
\nomenclature{$\phi$}{Azimuth from North}
\nomenclature{$\theta$}{Altitude}
\nomenclature{$D/M/Y$}{Input Day/Month/Year}

\bibliographystyle{elsarticle-harv}
\bibliography{zoterorefsDT}
\end{document}